\documentclass{article} 
\usepackage{iclr2025_conference,times}

\usepackage{amsmath,amsfonts,bm}

\def\eqref#1{equation~\ref{#1}}

\def\1{\bm{1}}

\DeclareMathAlphabet{\mathsfit}{\encodingdefault}{\sfdefault}{m}{sl}
\SetMathAlphabet{\mathsfit}{bold}{\encodingdefault}{\sfdefault}{bx}{n}

\usepackage[hidelinks]{hyperref, colortbl}
\usepackage{url, booktabs, multirow, diagbox}
\usepackage{svg}
\usepackage{xcolor, soul}
\usepackage{tikz}  
\usepackage{caption}
\usepackage{subcaption}
\usepackage{wrapfig}
\usepackage{amsthm, dsfont}
\theoremstyle{definition}

\usepackage{soul}
\usepackage{amssymb}
\usepackage{pifont}  
\usepackage{makecell}
\usepackage{tablefootnote}
\usepackage{bbm}

\newcommand{\xmark}{\text{\ding{55}}}
\newcommand{\xmarkpartial}{\text{\ding{55}}*}

\newcommand{\smat}[1]{{\scriptsize #1}}
\newcommand{\model}{\textsc{PaPaGei}}
\title{\protect\includegraphics[height=1.2em]{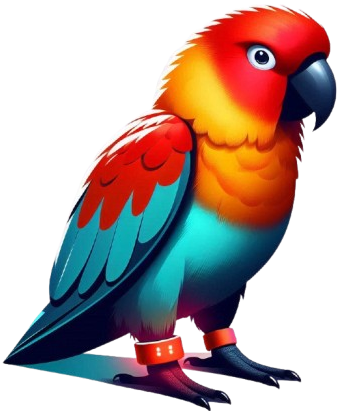}PaPaGei: Open Foundation Models for \newline Optical Physiological Signals}
\author{Arvind Pillai$^2$\thanks{Work has been done during the author’s internship at Nokia Bell Labs.},  Dimitris Spathis$^{1,3}$, Fahim Kawsar$^{1,4}$, Mohammad Malekzadeh$^1$ \\ 
$^1$Nokia Bell Labs, Cambridge, UK,  $^2$Dartmouth College, NH, USA, \\ $^3$University of Cambridge, UK, $^4$University of Glasgow, Scotland, UK\\}
\iclrfinalcopy
\begin{document}
\maketitle
\begin{abstract}
Photoplethysmography (PPG) is the leading non-invasive technique for monitoring biosignals and cardiovascular health, with widespread adoption in both clinical settings and consumer wearable devices. While machine learning models trained on PPG signals have shown promise, they tend to be task-specific and struggle with generalization. Current research is limited by the use of single-device datasets, insufficient exploration of out-of-domain generalization, and a lack of publicly available models, which hampers reproducibility. To address these limitations, we present \model{}, the first open foundation model for PPG signals. The model is pre-trained on over 57,000 hours of data, comprising 20 million unlabeled PPG segments from publicly available datasets. We introduce a novel representation learning approach that leverages domain knowledge of PPG signal morphology across individuals, enabling the capture of richer representations compared to traditional contrastive learning methods.
 We evaluate \model{} against state-of-the-art time-series foundation models and self-supervised learning benchmarks across 20 tasks from 10 diverse datasets, spanning cardiovascular health, sleep disorders, pregnancy monitoring, and wellbeing assessment. Our model demonstrates superior performance, improving classification and regression metrics by 6.3\% and 2.9\% respectively in at least 14 tasks. Notably, \model{} achieves these results while being more data- and parameter-efficient, outperforming models that are 70$\times$ larger. Beyond accuracy, we examine model robustness across different skin tones, establishing a benchmark for bias evaluation in future models. \model{} can serve as both a feature extractor and an encoder for multimodal models, opening up new opportunities for multimodal health monitoring\footnote{Models, data, and code are available at: \textcolor{blue}{\href{https://github.com/nokia-bell-labs/papagei-foundation-model}{github.com/nokia-bell-labs/papagei-foundation-model}}}.  
\end{abstract}

\section{Introduction}
\label{sec:introduction}
Photoplethysmography (PPG), a non-invasive optical sensing technique, is widely used to monitor cardiovascular health and physiological signals in both clinical and consumer health applications \citep{charlton20232023}. From hospital pulse oximeters to smartwatches, PPG enables continuous health monitoring in various settings, bridging acute medical care and long-term health management. PPG signals help in tracking a diverse range of health indicators, including cardiovascular health, blood pressure, mood, and sleep disorders \citep{sadad2022detection, ave2015early, reiss2019deep, liang2018new, haddad2021continuous, schrumpf2021assessment}. Despite its widespread adoption, PPG poses substantial challenges for machine learning applications. A primary obstacle is the high cost of data annotation, which requires specialized domain expertise. This challenge is particularly pronounced in consumer health applications, where varying sensing conditions and diverse user populations create additional complexity. PPG signals are susceptible to noise and motion artifacts \citep{afandizadeh2023accurate}, as well as inherent variability due to factors like skin tone and body composition \citep{bent2020investigating}. These complicate the development of generalizable ML models for PPG. Consequently, existing PPG datasets are often small, task-specific, and limited in their generalizability, posing a major obstacle to the development of robust and widely applicable models that could fully leverage the potential of PPG technology.

\begin{figure}
  \centering
  \includegraphics[width=.95\linewidth]{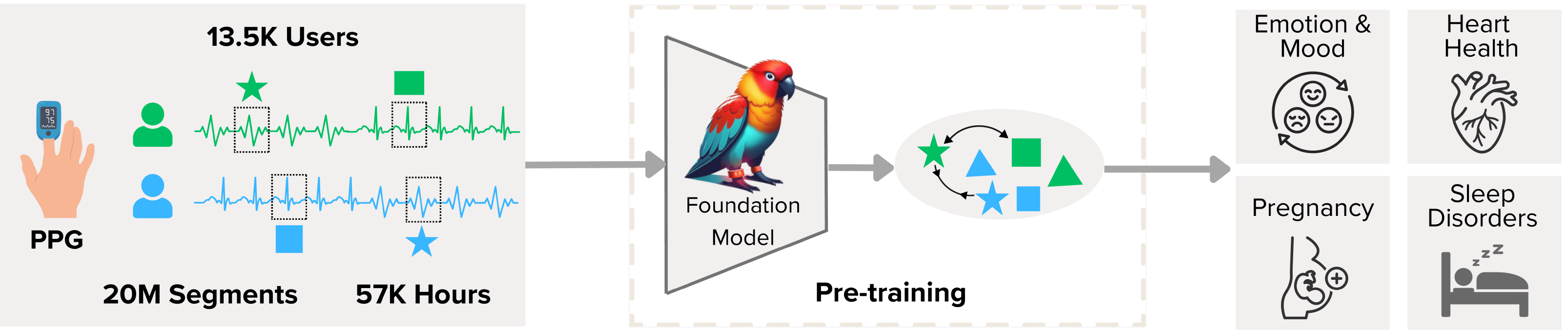}
  \caption{
  \model{} Overview. We curate public datasets of diverse PPG signals, and train a foundation model leveraging a novel morphology-aware contrastive learning approach. To evaluate its effectiveness, we apply the embeddings generated by \model{} to 20 tasks from 10 different datasets.}
  \label{fig:papagei_overview}
  \vspace{-0.5cm}
\end{figure}

The PPG domain, unlike language or vision domains, lacks general-purpose foundation models (FMs), with most current works focused on single-dataset task-specific models. Although PPG can detect vital signs like heart rate variability and blood oxygen saturation, the absence of generalizable pre-trained models limits progress \citep{abbaspourazad2023large}. Despite the ongoing challenges of acquiring large-scale, high-quality data, recent expansions in diverse PPG datasets have created new opportunities \citep{johnson2016mimic, zhang2018national, lee2022vitaldb}. To address these challenges, we introduce \textbf{\model{}}, a set of robust, pre-trained models capable of serving as a backbone for various PPG-related tasks, capturing rich PPG representations through large-scale pre-training.

The key \textbf{contributions} of \model{} are: 

\textbf{(1) Large-scale pre-training for PPG signals}: To our knowledge, \model{} is the first open foundation model pre-trained on PPG signals, using 57,000 hours of data from 20 million signals sourced entirely from public datasets. This establishes a new benchmark for large-scale model development in wearable and clinical health monitoring. 

\textbf{(2) PPG-aware self-supervised learning (SSL) framework}: We introduce a novel SSL framework with a unique PPG signal morphology augmentation module. Our approach optimizes agreement between PPG signals with similar blood volume changes while pointing the model to pay attention to the changes around the systolic peak and dicrotic notch (key PPG markers). 

\textbf{(3) Comprehensive evaluation across diverse out-of-domain health tasks}: We evaluate \model{} across 20 tasks, including cardiovascular health, sleep disorders, pregnancy monitoring, and overall well-being. Our results show that the model embeddings contain rich and predictive information applicable to various health conditions, outperforming existing benchmarks. 

\textbf{(4) Extensive robustness studies}: We conduct ablation studies to assess the impact of key components, including signal morphology augmentation, comparisons with established contrastive learning approaches, model size, data efficiency, and the effect of skin tone.

\vspace{-0.3cm}
\section{Related work}
\label{sec:related}
Self-supervised learning has become a prominent paradigm for learning general representations from unlabeled datasets, with applications in physiological signal analysis including health, fitness, and brain signals \citep{ tonekaboni2021unsupervised, zhang2022self, chen2021forecasting, yeche2021neighborhood, spathis2021self, cheng2020subject, kiyasseh2021clocs, sarkar2020self}. Despite its popularity, there are no widely used models for PPG signals pre-trained through SSL. Recently, \cite{abbaspourazad2023large} demonstrated that embeddings derived from PPG signals can predict over 45 diverse downstream health-related tasks using proprietary Apple Watch data. Their approach uses an SSL framework based on patient-level positive pair contrastive learning. Similarly, \citep{yun2024unsupervised} showed that embedding PPG signals can improve genetic discovery and risk prediction outcomes using the UK Biobank dataset. Other works \citep{weng2024predicting, ding2024siamquality, zhou2024utilizing} explored PPG embeddings for various applications. However, these studies often used proprietary datasets, did not explore out-of-domain generalization, or did not release their models, highlighting the need for openly available, pre-trained PPG FMs (Table \ref{tab:studies}). For example, in contrast to \citep{abbaspourazad2023large}, our work exclusively uses public datasets for large-scale PPG training and introduces a novel SSL framework to incorporate PPG morphology. While \cite{abbaspourazad2023large} evaluate a single proprietary dataset, we validate on 10 diverse downstream datasets, showcasing greater generalizability and robustness across varied real-world scenarios.

Generic time-series FMs, like Chronos \citep{ansari2024chronos} and Moment \citep{goswami2024moment}, lack physiological data representation. There is growing interest in modality-specific FMs tailored to physiological signals \citep{song2024foundation, lai2023practical} and human activity \citep{yuan2024self}. Knowledge transfer from time-series FMs might benefit PPG tasks, but their performance is limited compared to PPG-specific FMs. Adapting other domain-specific models, like ECG \citep{mckeen2024ecg,song2024foundation} or EEG \citep{yuan2024brant}, is challenging due to distinct signal characteristics. We specifically design FMs for PPG signals, contributing to the growing movement toward foundation models tailored to individual modalities. See Appendix~\S\ref{appendix:extended_related_work} for an extended discussion.

\vspace{-0.25cm}
\section{Methods}
\vspace{-0.25cm}
\label{sec:methods}
\begin{figure}
    \centering
    \includegraphics[width=\linewidth]{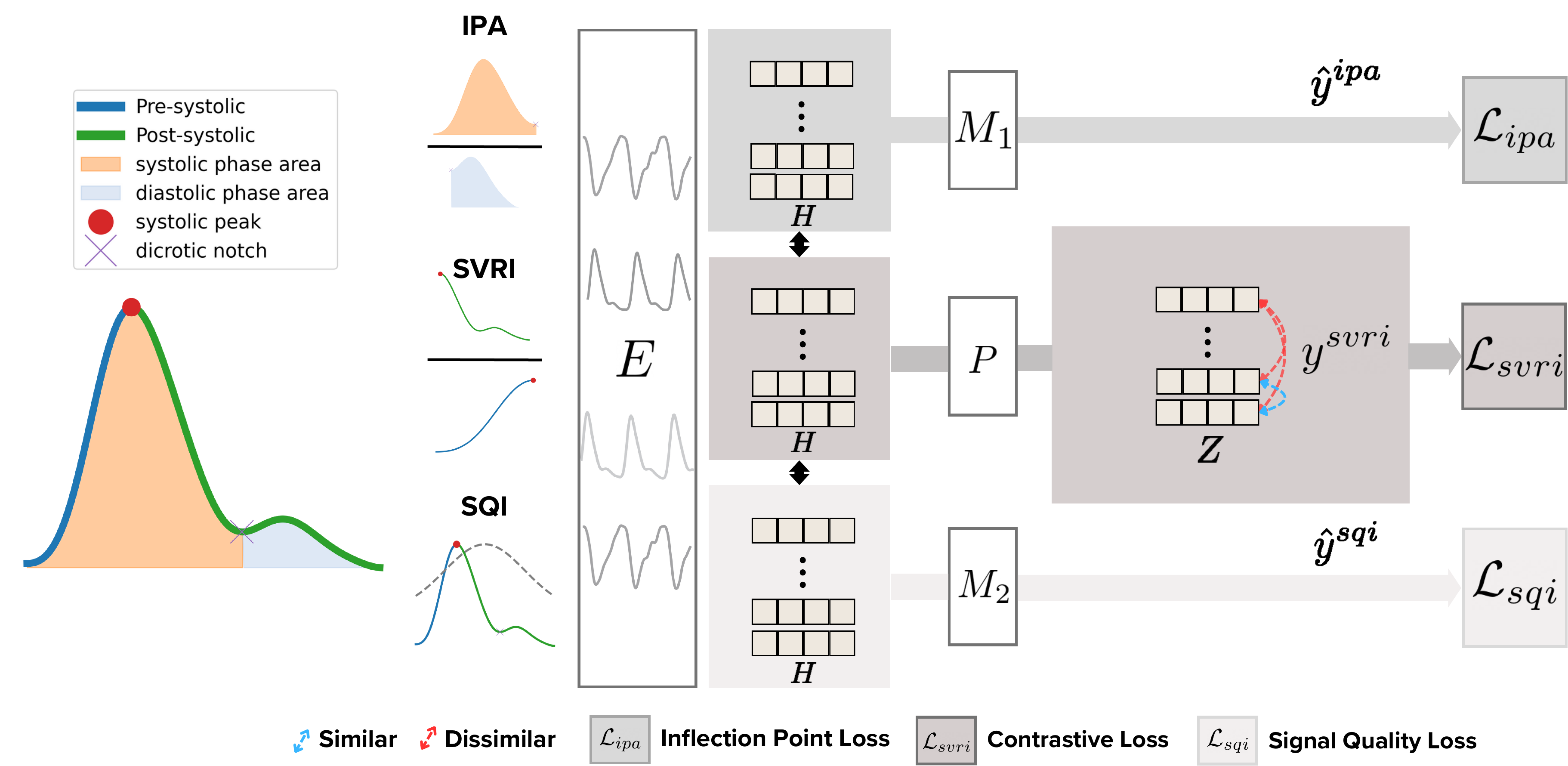}
    \caption{Overview of \model{}-S. The process begins by computing three morphology metrics (IPA, SVRI, and SQI) for each PPG segment. The raw PPG signals are then processed through an encoder ($E$) to generate embeddings ($H$). These same embeddings feed into three specialized heads: a projection head ($P$) that contrasts PPG signals based on sVRI values, and two mixture-of-expert heads ($M_1$ and $M_2$) that refine the embeddings by predicting IPA and SQI values.
} 
    \label{fig:papagei_s}
    \vspace{-0.5cm}
\end{figure}

Given a dataset $\mathcal{D} = \{\mathbf{p}^1, \mathbf{p}^2, \cdots, \mathbf{p}^S\}$ representing diverse PPG signals from $S$ participants, a PPG signal $\mathbf{p}^s \in \mathbb{R}^n$ is defined as a time-series that captures variations in light intensity caused by arterial blood flow. To model granular changes in PPG signal of each subject $s$, we segment $\mathbf{p}^s$ without overlap to obtain $X^s = \{\mathbf{x}_1^s, \mathbf{x}_2^s, \cdots \mathbf{x}^s_{N}\}$. Here, the number of segments $N$ depends on the sampling frequency ($f$) and the desired length of time window. To train our foundation models, \model{}-P employs a {\em patient contrastive} SSL approach that maximizes agreement between signals from the same subject. Importantly, we propose \model{}-S, a {\em morphology-aware} self-supervised approach that maximizes agreement between PPG segments with similar morphology.
\vspace{-0.25cm}
\subsection{Participant-aware objective: \model{}-P}
\vspace{-0.2cm}
In \model{}-P, we train an SSL model to maximize agreement between the embeddings of PPG signals from the same subject. While previous studies have demonstrated the effectiveness of this strategy for physiological signals~\citep{kiyasseh2021clocs, abbaspourazad2023large}, our work represents the first attempt to train and evaluate a foundation model using publicly available PPG datasets.

\noindent\textbf{Training.} We define a {\em positive pair} as any two distinct segments of PPG signals from the same subject, denoted as $\{(\mathbf{x}^s_i, \mathbf{x}^s_j) | i \neq j\}$. Next, we apply a series of time-series augmentations such as random cropping, adding Gaussian noise, time flipping, negation, and magnitude scaling \citep{tang2020exploring}, each applied with a predefined probability during training. Each augmentation includes hyper-parameters that control the intensity of the data transformation. During training, the augmented version of a randomly sampled positive pair $(\mathbf{x}^s_i, \mathbf{x}^s_j)$ is passed through the encoder $E$, and subsequently projection $P$, to obtain an {\em embeddings} pair denoted by $(\mathbf{z}_i^s, \mathbf{z}_j^s)$. Given a batch of embeddings from $N$ positive pairs of the form $(\mathbf{z}_i, \mathbf{z}_j)$, the model optimizes the normalized temperature-scaled cross entropy (NT-Xent) loss \citep{sohn2016improved, oord2018representation, chen2020simple} given by: $\mathcal{L}_p = \frac{1}{2}(\ell_p(i, j) + \ell_p(j, i))$, where 
$\ell_p(i, j) = - \frac{1}{N} \sum_{u=1}^N \log \frac{\exp(sim(\mathbf{z}_i^u, \mathbf{z}_j^u)/\tau)}{\sum_{v=1}^{2N} \mathbbm{1}[v \neq u] \exp(sim(\mathbf{z}_i^u, \mathbf{z}_j^v)/\tau)}$ and $sim(\cdot, \cdot)$ is the cosine similarity. In contrast, vanilla SimCLR \citep{chen2020simple} would use positive pairs as augmented versions of randomly sampled PPG segments.

\vspace{-0.1cm}
\subsection{Morphology-aware objective: \model{}-S} \label{sec:segment_aware}
\vspace{-0.1cm}

In \model{}-S, we leverage the PPG signal morphology to train a SSL model that maximizes agreement between similar physiological features of PPG signals across participants.

\noindent\textbf{PPG Morphology.} Total peripheral resistance (TPR)---the force exerted by the body's blood vessels on circulating blood---varies under certain medical conditions, such as hypertension and diabetes \citep{trammel2020physiology}. Variations in TPR are reflected in PPG signals, presenting as distinct regions within the waveform. 
To capture these variations, we introduce a morphology augmentation module before training, which computes three key PPG metrics (Figure \ref{fig:papagei_s}, left): (1) \textbf{stress-induced Vascular Response Index (sVRI)} \citep{lyu2015measuring, zhang2019photoplethysmogram}: the ratio of mean PPG signal between post- to pre-systolic phases, (2) \textbf{Inflection Point Area ratio (IPA)} \citep{wang2009noninvasive}: the ratio of systolic to diastolic areas defined by the dicrotic notch, and (3) \textbf{Signal Quality Index (SQI)}: skewness of the signal as an indicator of quality \citep{elgendi2016optimal}. Prior studies have shown that incorporating the PPG signal quality during training yields positive results \citep{ding2024siamquality}. We selected these metrics for their complementary nature: sVRI captures variations in amplitude, while IPA measures signal width. To address scenarios where computing IPA is challenging because of noisy signals or different morphology, we incorporate SQI. In particular, we empirically find that SQI is significantly larger ($p < 0.05$) in signals with a dicrotic notch (Appendix~\S \ref{appendix:statistics_ipa_sqi}).
\begin{align*}
    sVRI(\mathbf{x}) = \frac{sys\sum_{i=sys}^{n} x_i}{(n-sys)\sum_{i=1}^{sys}x_i},
    \quad
    IPA(\mathbf{x}) = \frac{\int_0^{\hat{n}} \mathbf{x} \, dn}{\int_{\hat{n}}^{n} \mathbf{x} \, dn}, \text{ and}
    \quad
    SQI(\textbf{x}) = \frac{1}{W}\sum_{w}\frac{m_3}{m_2^{3/2}}, \tag{1} \label{eqn:svri}
\end{align*}
where $\mathbf{x} \in \mathbb{R}^N$ is the PPG segment, $sys$ is the systolic peak, $n$ is the length of time series, and $\hat{n}$ is the dicrotic notch. For $SQI$, we divide $\textbf{x}$ into 5 second windows ($w$; total windows $W$) and compute the skewness $m_i = \frac{1}{5\times f}\sum_{j=1}^{5 \times f}(x[j] - \mu_x[j])^i$, which gives the best signal quality discrimination. 

\noindent\textbf{Training.} Before training, the morphology augmentation module takes an augmented input, by applying Gaussian noise and cropping to time series $\mathbf{x}$, and outputs $y = \{y^{svri}, y^{ipa}, y^{sqi}\} \in \mathbb{R}^3$ (Figure \ref{fig:papagei_s} middle). Next, we discretize $y^{svri}$ into a predefined set of $b = 8$ bins to denote positive pairs, where $y^{svri} \in \{1, \dots, b\}$. We define positive pairs based on the sVRI labels as $\{(\mathbf{x}_i, \mathbf{x}_j) | y^{svri}_i = y^{svri}_j, i \neq j \}$. Note that positive pairs are not defined based on participants.
\begin{align}
    \ell_s(i, j) = - \log \frac{\exp\left(sim(\mathbf{z}_i, \mathbf{z}_j)/\tau\right)}{\sum_{k=1}^{2N} \mathbbm{1}[k \neq i] \exp\left(sim(\mathbf{z}_i, \mathbf{z}_k)/\tau\right)} \tag{2}  \\
    \mathcal{L}_{svri} = \frac{1}{2N} \sum_{k=1}^{N}[\ell(2k-1, 2k) + \ell(2k, 2k-1)] \quad \tag{3} \label{eqn:contrastive_svri}\\
    \mathcal{L}_{ipa} = \frac{1}{N}\sum_{i=1}^N \left| y^{ipa}_i - \hat{y}^{ipa}_i \right| \quad 
    \mathcal{L}_{sqi} = \frac{1}{N}\sum_{i=1}^N \left| y^{sqi}_i - \hat{y}^{sqi}_i \right| \tag{4} \label{eqn:multi_loss}\\
    \mathcal{L}_s = \alpha \mathcal{L}_{svri} + (1 - \alpha)\left( \mathcal{L}_{ipa} + \mathcal{L}_{sqi} \right), \text{where} \;\alpha \in [0, 1] \tag{5} \label{eqn:papagei_s}
\end{align}

Given a batch of $N$ PPG signals and their morphology, we optimize three heads using the encoder ($E$) embeddings $H = \{\mathbf{h}_1, \mathbf{h}_2, \cdots, \mathbf{h}_N\}$. First, we extract the embeddings $Z = \{\textbf{z}_1, \textbf{z}_2, \cdots, \textbf{z}_N\}$ from the projection ($P$), and compute the contrastive loss for sVRI (equation \ref{eqn:contrastive_svri}). Next, we use the embeddings $H$ to predict the IPA ($\mathbf{\hat{y}}^{ipa} \in \mathbb{R}^{N}$) and SQI ($\mathbf{\hat{y}}^{sqi} \in \mathbb{R}^N$) using the mixture of expert (MoE) heads $M_1$ and $M_2$. Each MoE head is composed of three fully connected neural networks (FCNNs), with the head's output calculated as a weighted sum of the FCNNs, using softmax to determine the weights. These heads are optimized using the mean absolute error (equation \ref{eqn:multi_loss}). The morphology indices encapsulate various PPG characteristics. Our rationale for utilizing MoE is that each expert can specialize in learning distinct properties that contribute to the overall index. Finally, the overall \model{}-S training objective is given in equation \ref{eqn:papagei_s}.

\vspace{-0.4cm}
\section{Experiments}
\vspace{-0.2cm}
\label{sec:experiments}
\subsection{Pre-training} \label{sec:pretraining}
\vspace{-0.25cm}
\noindent\textbf{Datasets.} We pre-train \model{} on three datasets: (1) VitalDB \citep{lee2022vitaldb}, which includes PPG signals collected during surgery from the patient's finger ($f$=500Hz), (2) the MIMIC-III waveform database matched subset \citep{johnson2016mimic}, where finger-tip PPG data is collected from an ICU monitor ($f$= 125Hz), and (3) the Multi-Ethnic Study of Atherosclerosis (MESA) sleep sub-study \citep{zhang2018national, chen2015racial}, which provides PPG data obtained through finger-tip polysomnography ($f$= 256Hz). In total, we have 13.5K participants with 20M segments (Table \ref{tab:pretraining_dataset}).

\begin{wraptable}{r}{0.42\textwidth}
\centering
\caption{\model{}'s pre-training datasets.}
\label{tab:pretraining_dataset}
\scalebox{0.8}{
\begin{tabular}{@{}llll@{}}
\toprule
\textbf{Dataset} & \textbf{\#Participants} & \textbf{\#Segments} & \textbf{Hours}  \\ \midrule
VitalDB   &5,866          &6,248,100               &17,355        \\
MIMIC-III        &5,596          &7,196,401                &19,990      \\
MESA        &2,055          &7,306,705                &20,296        \\
 \midrule
Total & 13,517  &20,751,206 &57,641 \\
\bottomrule
\end{tabular}
}
\end{wraptable} 

\noindent\textbf{Pre-processing.} To curate {single-channel} PPG signals across all datasets, we perform the following steps: (1) Apply a 4th-order Chebyshev bandpass filter with low and high pass cut-offs set at 0.5Hz and 12Hz, respectively \citep{lapitan2024estimation, liang2018optimal}; (2) {Segment the signal into 10-second windows (\citep{orphanidou2018quality, koteska2022deep} use 10s windows whereas larger studies use 30s \citep{ding2024siamquality} and 60s \cite{abbaspourazad2023large})}; (3) Detect flatline segments and remove any segment where more than 25\% of the data is flat \citep{biobss_documentation}; (4) Normalize the segments using Z-score \citep{temko2017accurate, zhou2017indexing}; and (5) Resample the segments to 125Hz (the lowest sampling rate of our pre-training datasets, MIMIC-III). 
\begin{wraptable}{}{0.65\textwidth}
\centering
\caption{\model{}'s evaluation datasets. Gray lines are unseen during training (out-of-domain). For those used for pre-training, we keep a held-out test-sets and use labels. Task Types are: B=binary, R=regression, M-\#classes= muticlass classification.}
\label{tab:downstream_tasks_main}
\scalebox{0.68}{
\begin{tabular}{@{}llll@{}}
\toprule
\textbf{\#ID} &\textbf{Dataset} & \textbf{Task (Task Type)} &\textbf{\#Subj.(\#Samp.)} \\ \midrule
T1 & VitalDB \citep{lee2022vitaldb} &ICU admission (B) &5866 \\ 
T2 & &Operation Type (M-9) &5866 \\
T3 &MIMIC-III \citep{moody2020mimic} &Mortality (B) &5596 \\
T4 &MESA \citep{zhang2018national} &Smoker (B) &2055 \\
T5 & &AHI $>$ 3\% Oxygen Desat. (R) &2055 \\
T6 & &AHI $>$ 4\% Oxygen Desat. (R) &2055 \\
T7 &\cellcolor{gray!10}nuMom2B \citep{facco2015numom2b} &Pregnancy stage (B) &3163 (5337) \\ 
T8 &\cellcolor{gray!10} &Gestation Age (R) &3163 (5337) \\
T9 &\cellcolor{gray!10}VV (Skin Tone) \citep{toye2023vital} &Systolic BP* (R) &231 \\ 
T10 &\cellcolor{gray!10} &Diastolic BP*  (R)&231 \\

T11 &\cellcolor{gray!10}PPG-BP \citep{liang2018new}& Systolic BP (R) &219\\
T12 &\cellcolor{gray!10} &Diastolic BP (R) &219 \\
T13 &\cellcolor{gray!10}&Average Heart Rate (R)&219 \\
T14 &\cellcolor{gray!10} &Hypertension (B) &219 \\
T15 &\cellcolor{gray!10}SDB \citep{garde2014development} &Sleep Disordered Breathing (B) &146 \\
T16 &\cellcolor{gray!10}ECSMP \citep{gao2021ecsmp} &Mood Disturbance (B) &89 \\
T17 &\cellcolor{gray!10}WESAD \citep{schmidt2018introducing} &Valence (B) &15 (4497) \\
T18 &\cellcolor{gray!10} &Arousal (B) &15 (4497) \\
T19 &\cellcolor{gray!10}PPG-DaLiA \citep{reiss2019deep} &Heart Rate (R)&15 (64697) \\
T20 &\cellcolor{gray!10} &Activity (M-9) &15 (64697) \\
\bottomrule
\end{tabular}
}
\end{wraptable}

\noindent\textbf{Implementation.} We adopt a ResNet-style CNN encoder, following \citep{ding2024siamquality}. \cite{abbaspourazad2023large} also utilize an EfficientNet-style CNN. Our model has 18 convolutional blocks, starting with a filter size of 32, which doubles every 4 blocks. The projection layer is a single FC layer, generating a 512-d embedding. In the \model{}-S variant, the expert block ($M_1$ \& $M_2$) uses three parallel FCNNs, each with two FC layers, resulting in a 128-d embedding. For augmentations, \model{}-P uses cropping (0.50), negation (0.20), flipping (0.20), and scaling (0.40). \model{}-S uses cropping (0.25) and Gaussian noise (0.25). \model{}-S avoids augmentations that alter PPG's morphology. We set $\alpha=0.6$ and train on eight V100 GPUs for 15,000 steps (lr$=10^{-4}$), with \model{}-P and \model{}-S having 5M and 5.7M parameters, respectively, {while previous works use model sizes of 3.3M \citep{abbaspourazad2023large} (we study scaling in Section \ref{sec:ablation_study})}.

\vspace{-0.2cm}
\subsection{Downstream Tasks}
\vspace{-0.2cm}
To evaluate the effectiveness of \model{}, we benchmark it against a diverse set of datasets, tasks, and baselines, chosen for their large size and clinical relevance (where applicable)\footnote{\url{https://peterhcharlton.github.io/post/ppg\_datasets/}}. A description of the tasks with their corresponding \#ID is provided in Table \ref{tab:downstream_tasks_main}, with further details in Appendix~\S \ref{appendix:datasets_tasks}. As a motivation, identifying patient risk factors is crucial for hospitals to allocate resources effectively. To address this, we evaluate several indicators, including ICU admission (T1), type of operation (T2), mortality (T3), and smoking status (T4). For sleep apnea diagnosis, the American Academy of Sleep Medicine recommends using the Apnea/Hypopnea Index (AHI) with at least 3\% or 4\% oxygen desaturation as a key metric \citep{ruehland2009new}. Thus, we predict AHI at 3\% and 4\% desaturation thresholds (T5 \& T6) and classify sleep-disordered breathing (T15). For pregnancy outcomes, changes in gestational age and pregnancy stage are linked to risks like hypertensive disorders and small-for-gestational-age delivery \citep{bouariu2022first, wu2020gestational, crump2023adverse}, enabling us to classify pregnancy stage (T7) and predict gestational age (T8). In cardiovascular health, we estimate systolic (T9 \& T11) and diastolic (T10 \& T12) blood pressure (BP) using two datasets. While PPG-BP (T11 \& T12) provides high-frequency, short PPG signals, the VV dataset helps explore skin tone's influence on BP estimation. We also assess hypertension classification (T14), average seated heart rate (T13), and continuous heart rate during activities (T19), along with activity classification (T20). In the emotion domain, we classify PPG signals into mood disturbance levels (T16), valence (T17), and arousal (T18).

\begin{figure}
     \centering
     \begin{subfigure}[b]{0.24\textwidth}
         \centering
         \includegraphics[width=\textwidth]{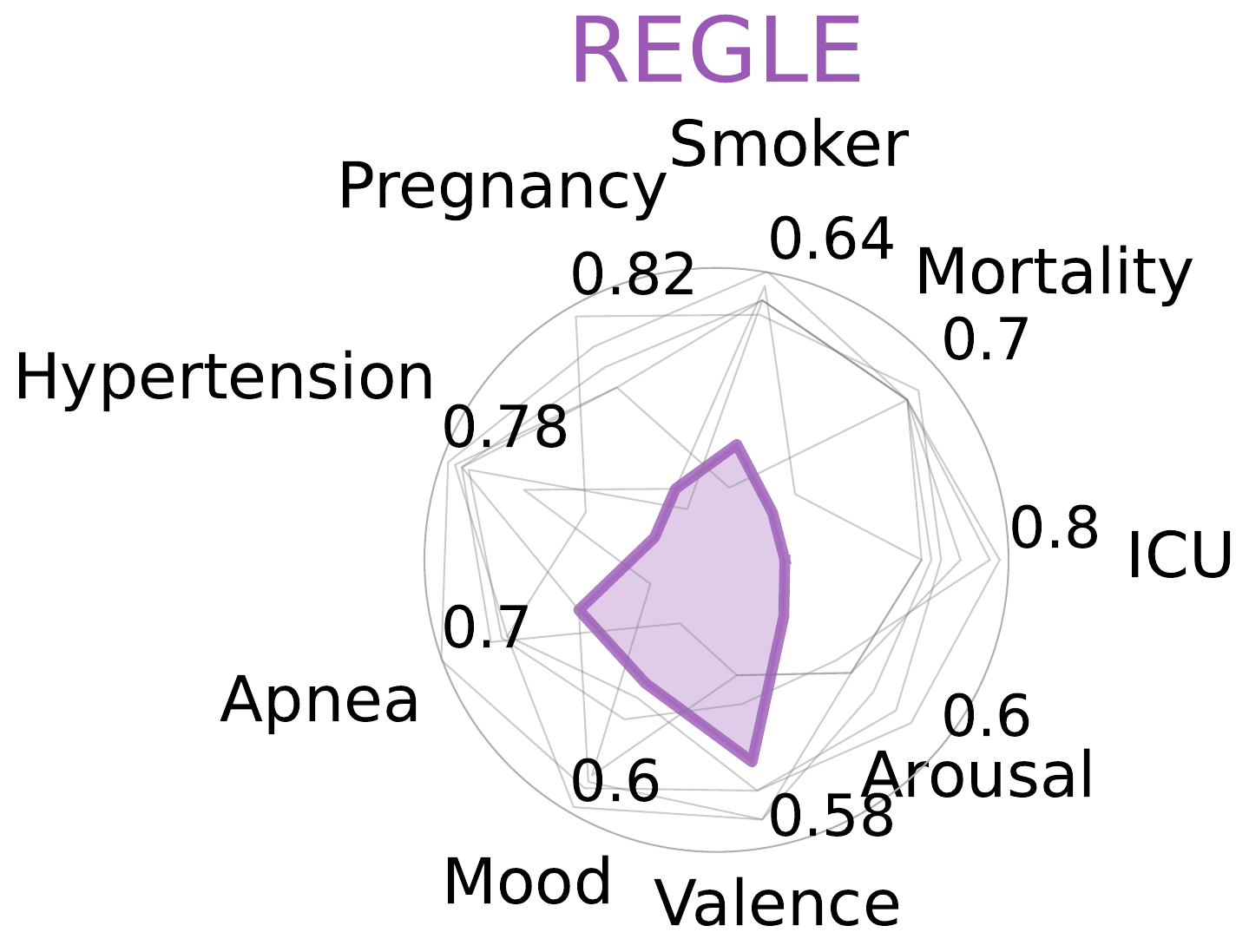}     \end{subfigure}
     \hfill
     \begin{subfigure}[b]{0.24\textwidth}
         \centering
         \includegraphics[width=\textwidth]{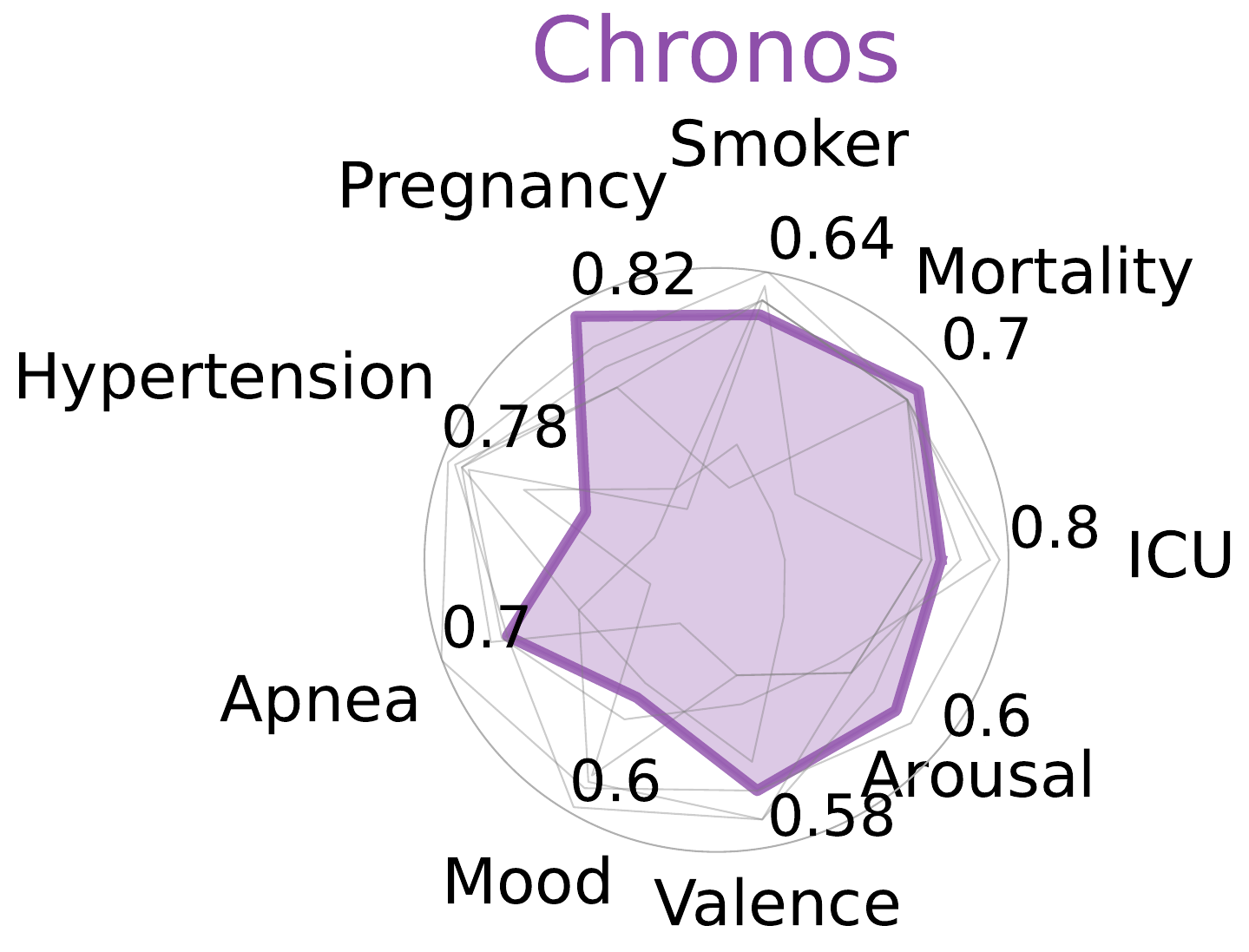}
     \end{subfigure}
     \hfill
     \begin{subfigure}[b]{0.24\textwidth}
         \centering
         \includegraphics[width=\textwidth]{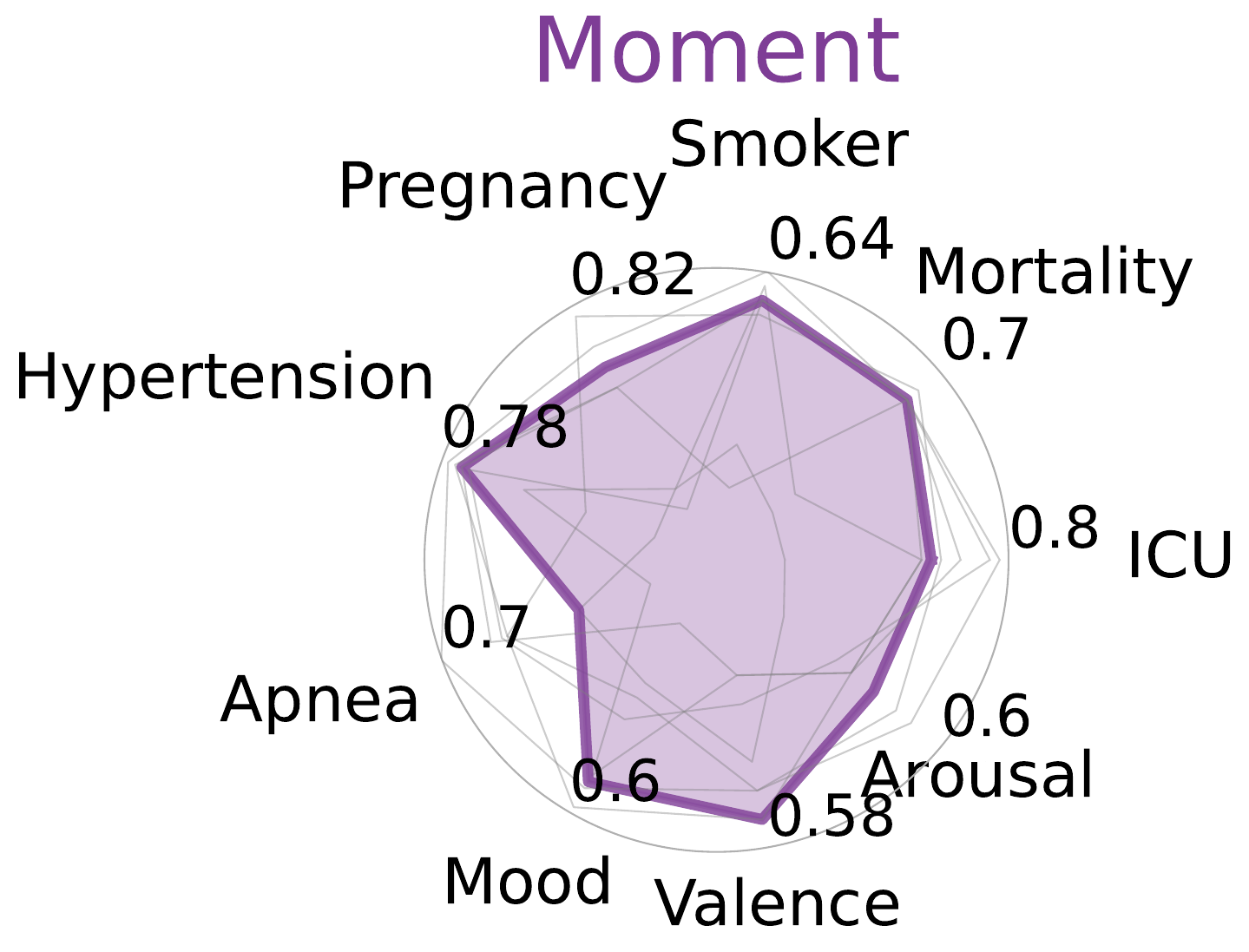}
     \end{subfigure} 
    \begin{subfigure}[b]{0.24\textwidth}
         \centering
         \includegraphics[width=\textwidth]{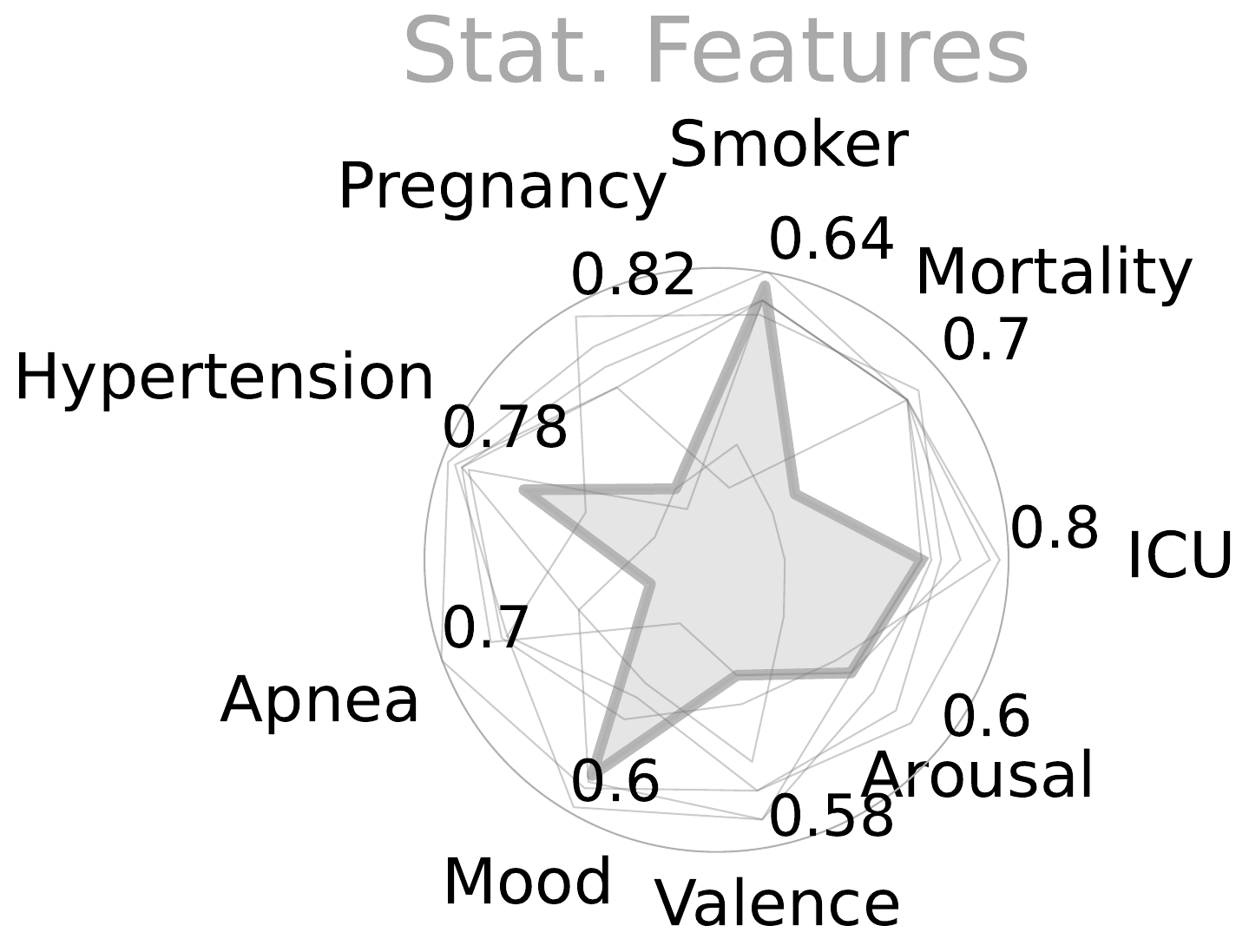}
     \end{subfigure}
     \hfill
    \begin{subfigure}[b]{0.24\textwidth}
         \centering
         \includegraphics[width=\textwidth]{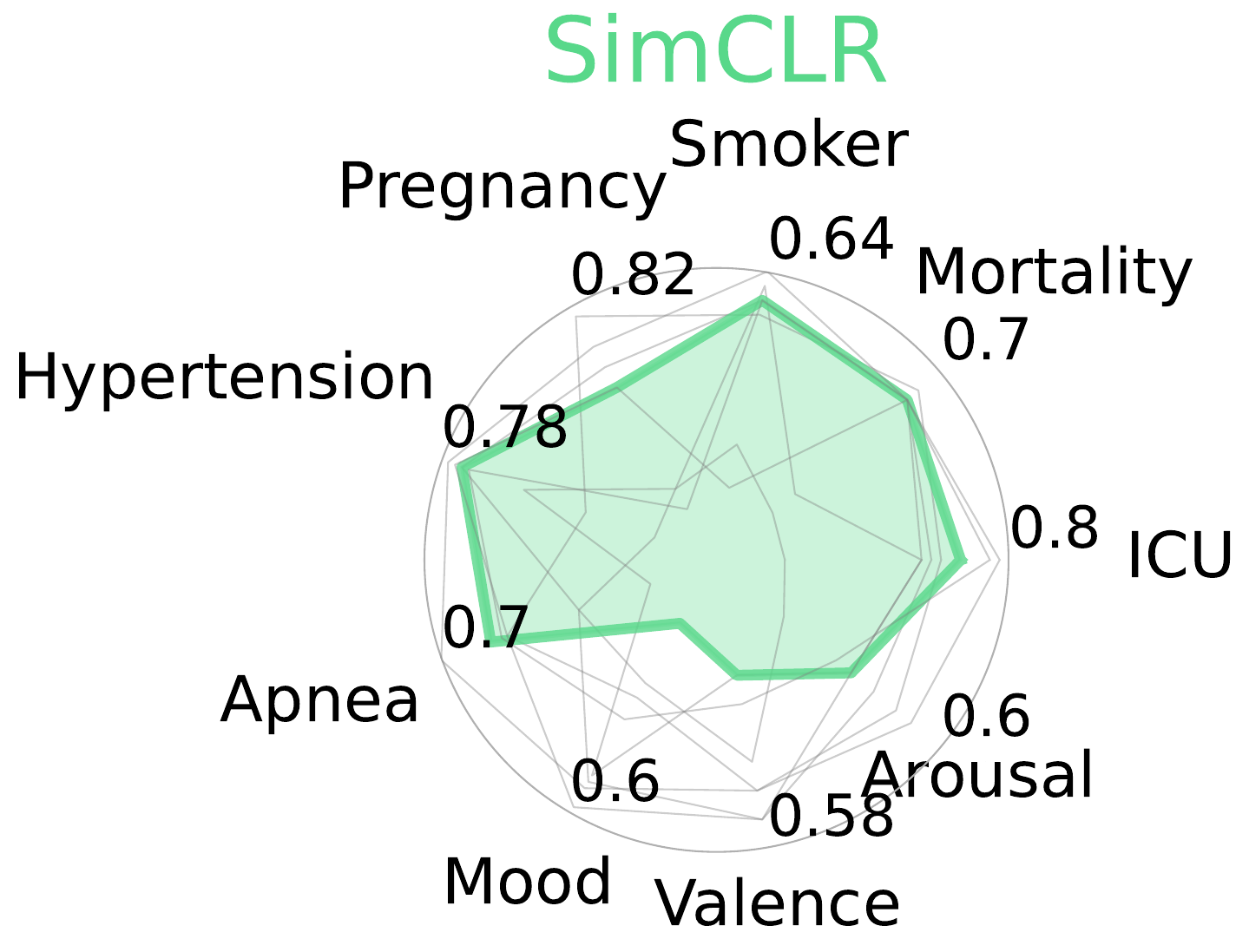}
     \end{subfigure}
     \hfill
     \begin{subfigure}[b]{0.24\textwidth}
         \centering
         \includegraphics[width=\textwidth]{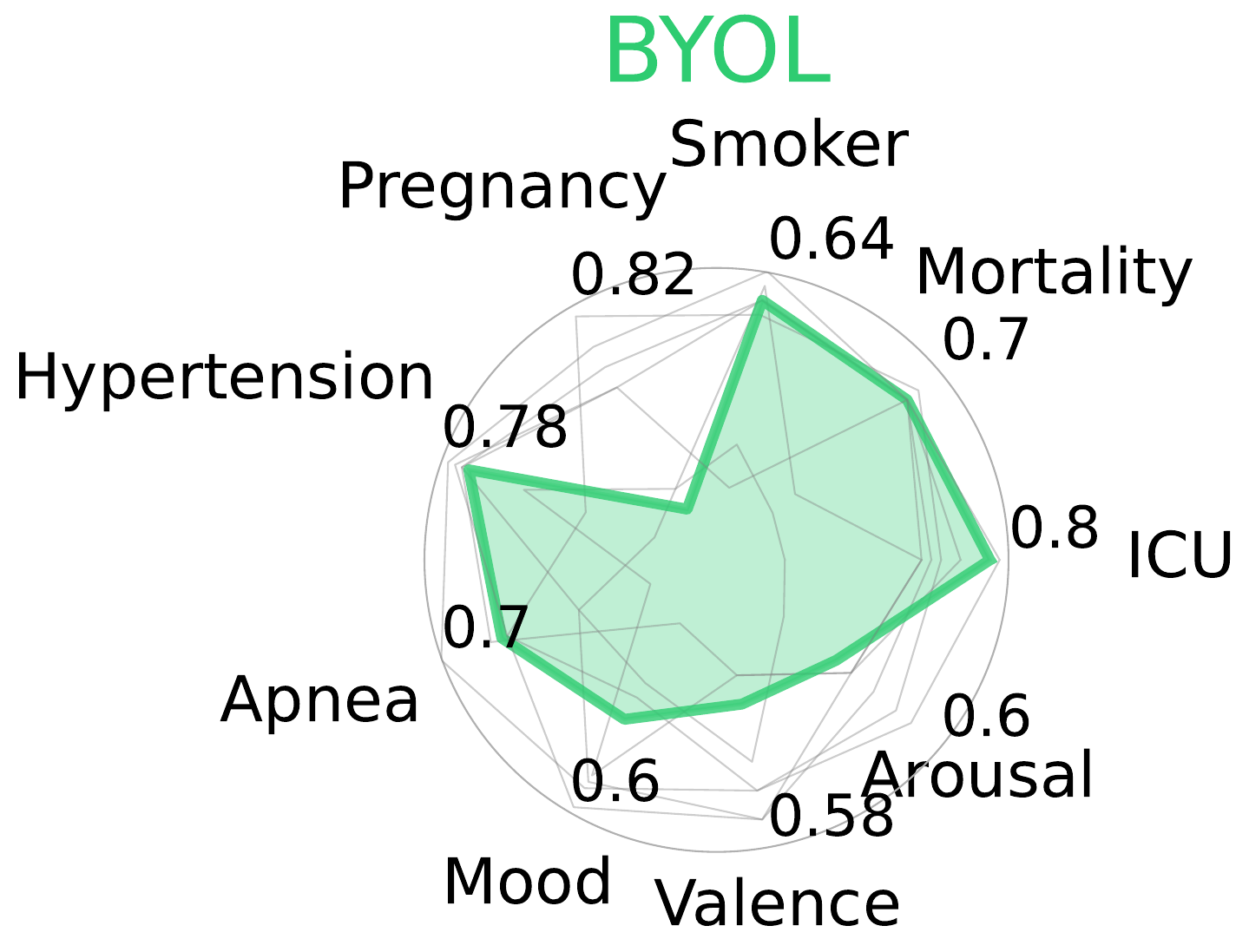}
     \end{subfigure}
     \hfill
     \begin{subfigure}[b]{0.24\textwidth}
         \centering
         \includegraphics[width=\textwidth]{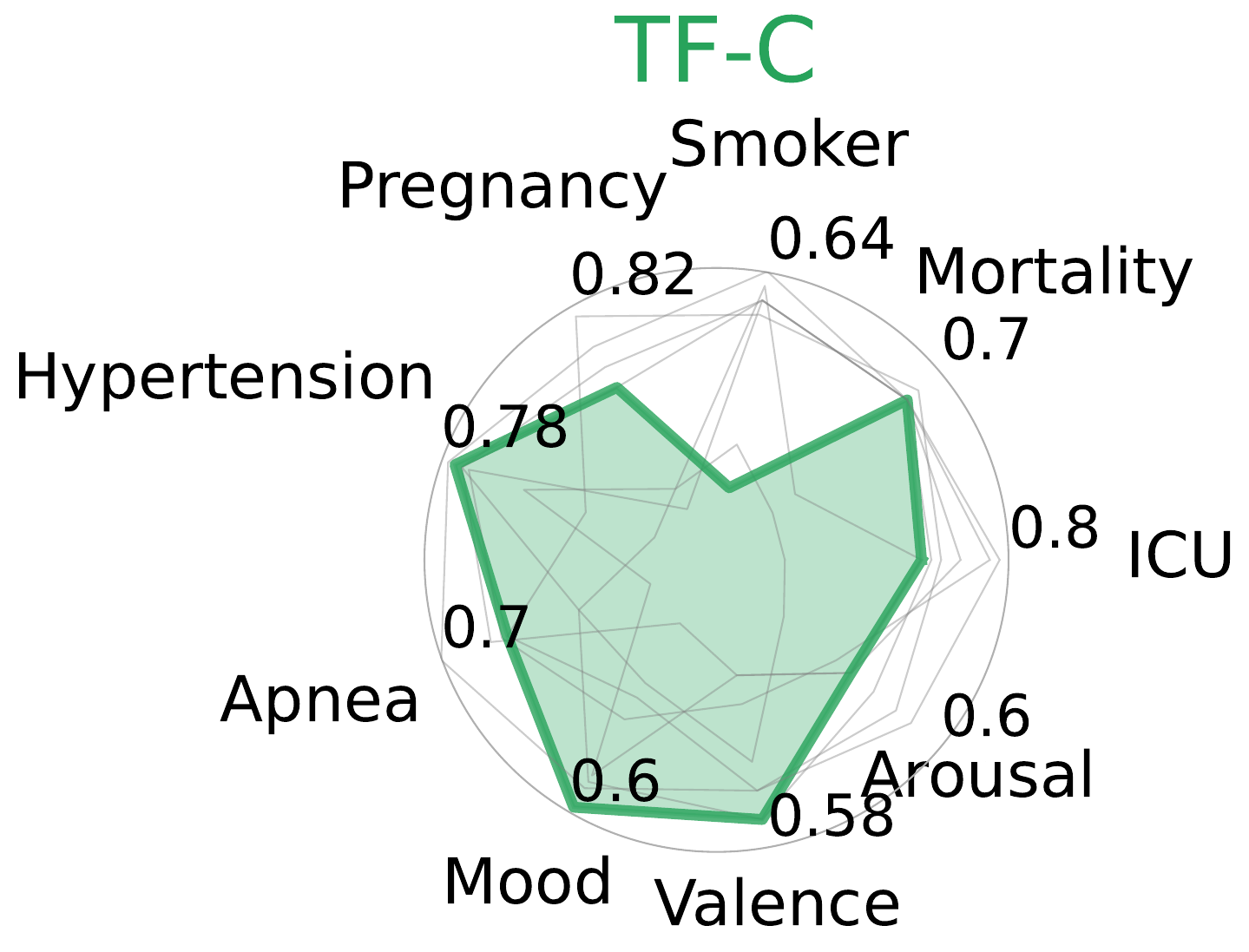}
     \end{subfigure}
     \hfill
     \begin{subfigure}[b]{0.24\textwidth}
         \centering
         \includegraphics[width=\textwidth]{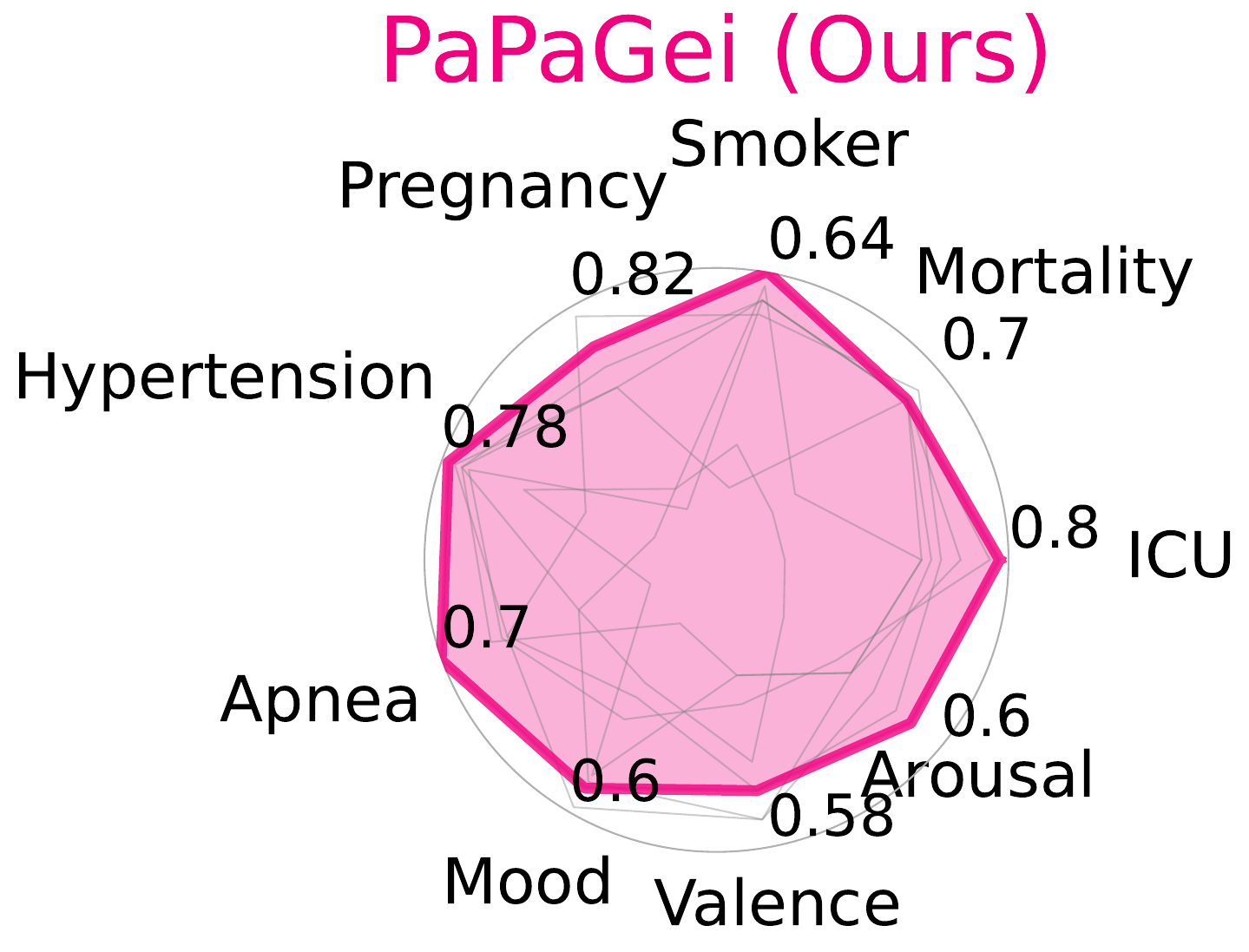}
     \end{subfigure}
     \hfill
     \vspace{0.25cm}
     \hrule
     \vspace{0.25cm}
     \begin{subfigure}[b]{0.24\textwidth}
         \centering
         \includegraphics[width=\textwidth]{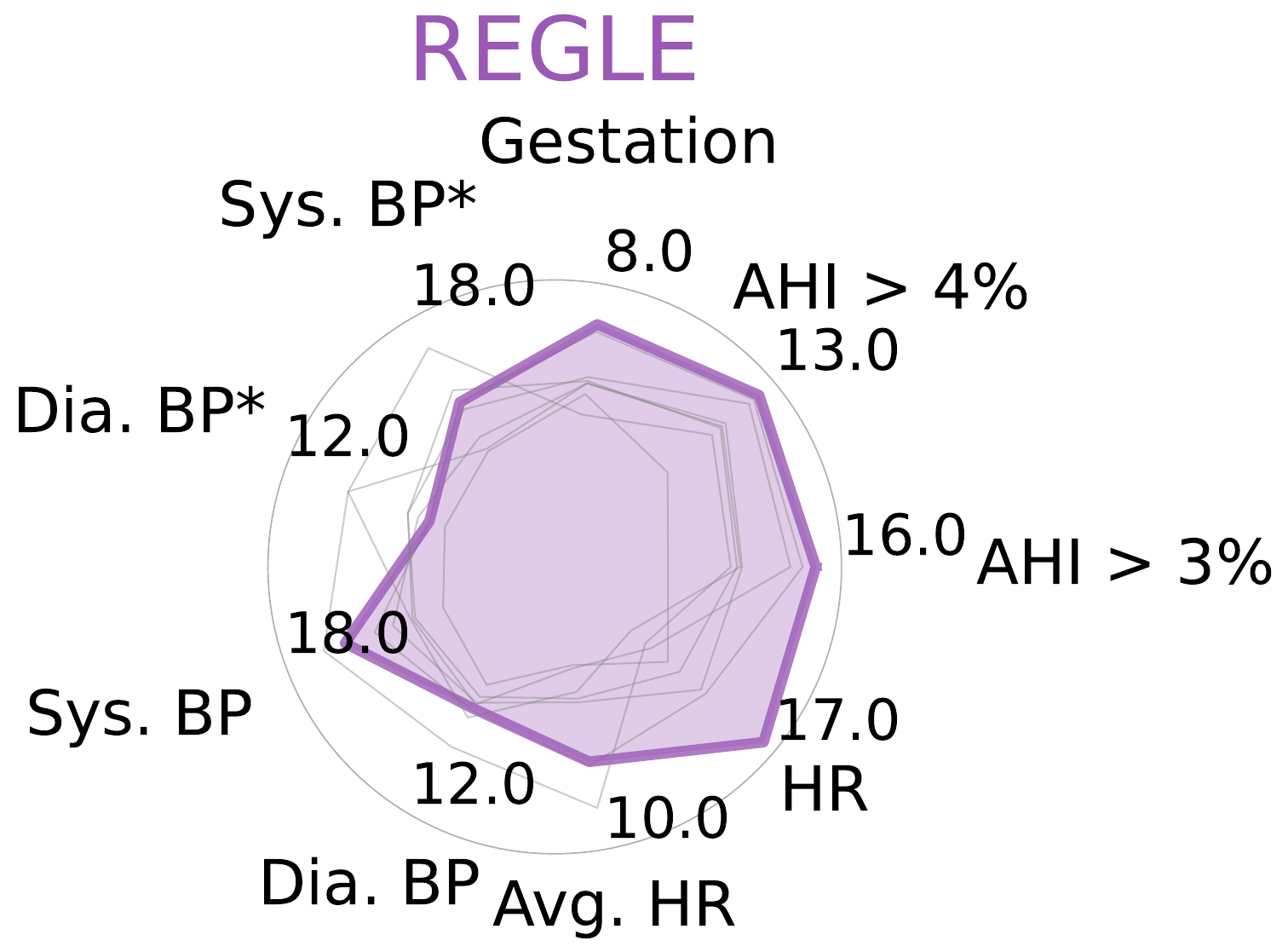}
     \end{subfigure}
     \hfill
     \begin{subfigure}[b]{0.24\textwidth}
         \centering
         \includegraphics[width=\textwidth]{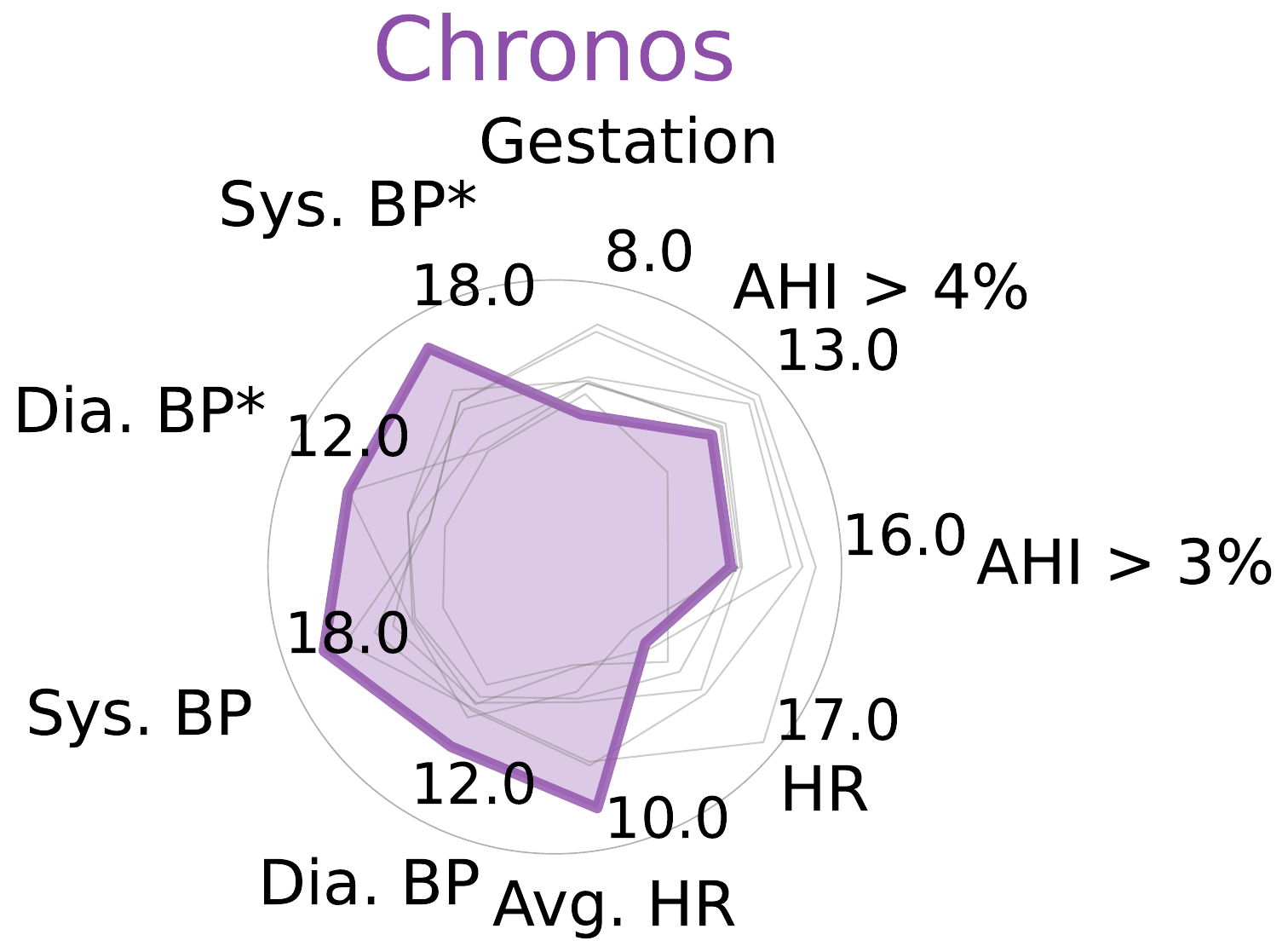}
     \end{subfigure}
     \hfill
     \begin{subfigure}[b]{0.24\textwidth}
         \centering
         \includegraphics[width=\textwidth]{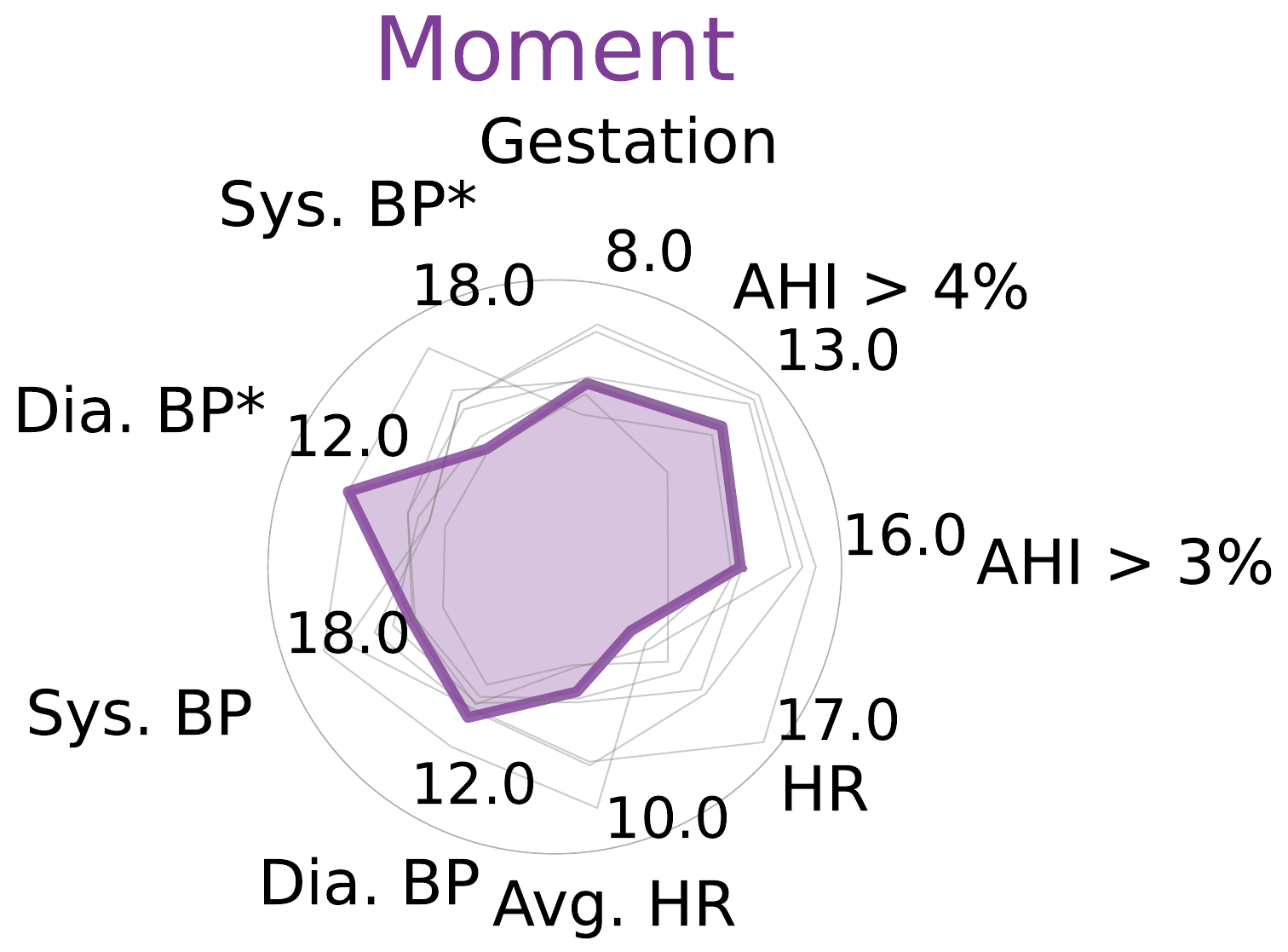}
     \end{subfigure} 
     \begin{subfigure}[b]{0.24\textwidth}
         \centering
         \includegraphics[width=\textwidth]{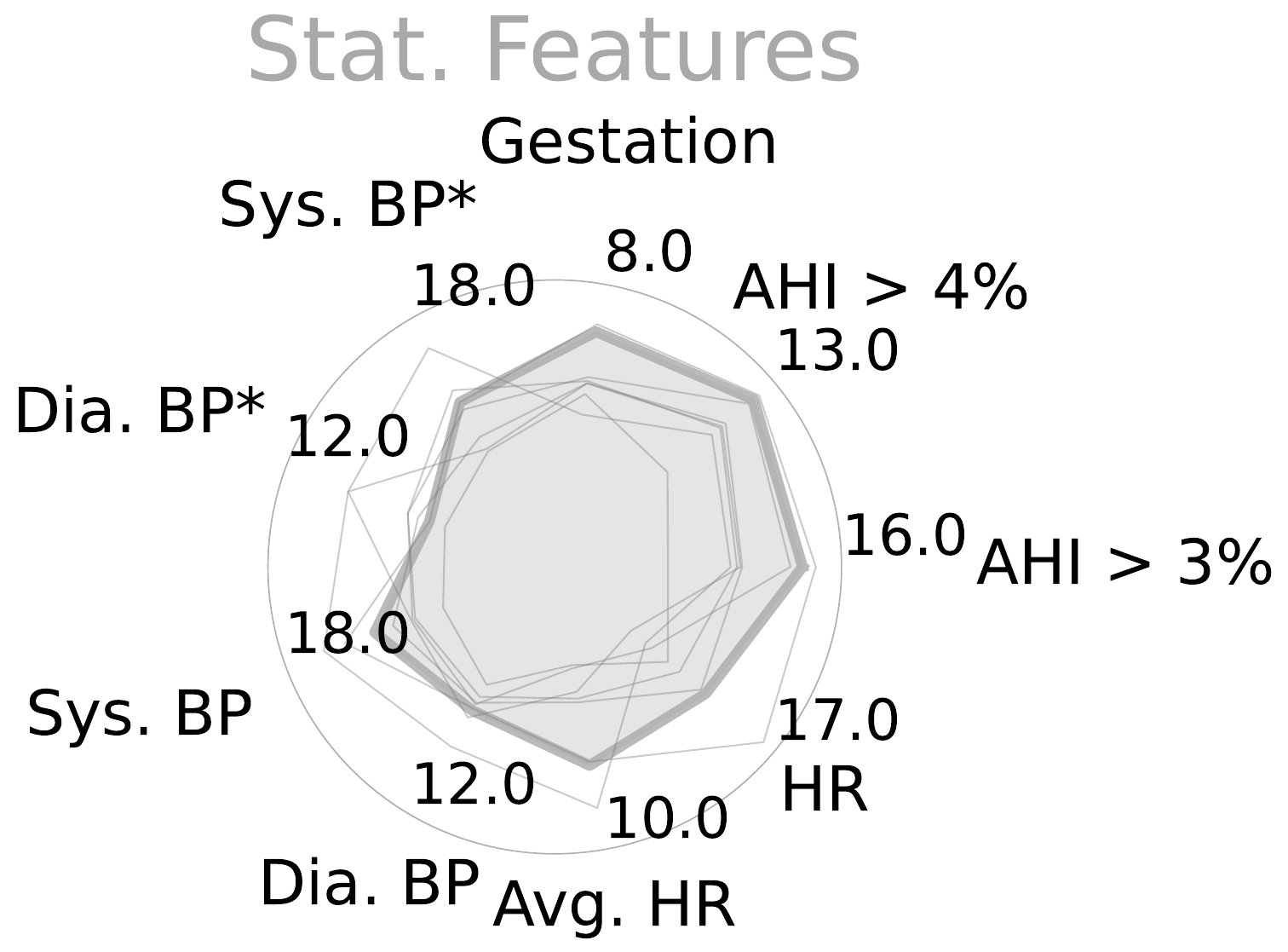}
     \end{subfigure}
     \hfill
     \begin{subfigure}[b]{0.24\textwidth}
         \centering
         \includegraphics[width=\textwidth]{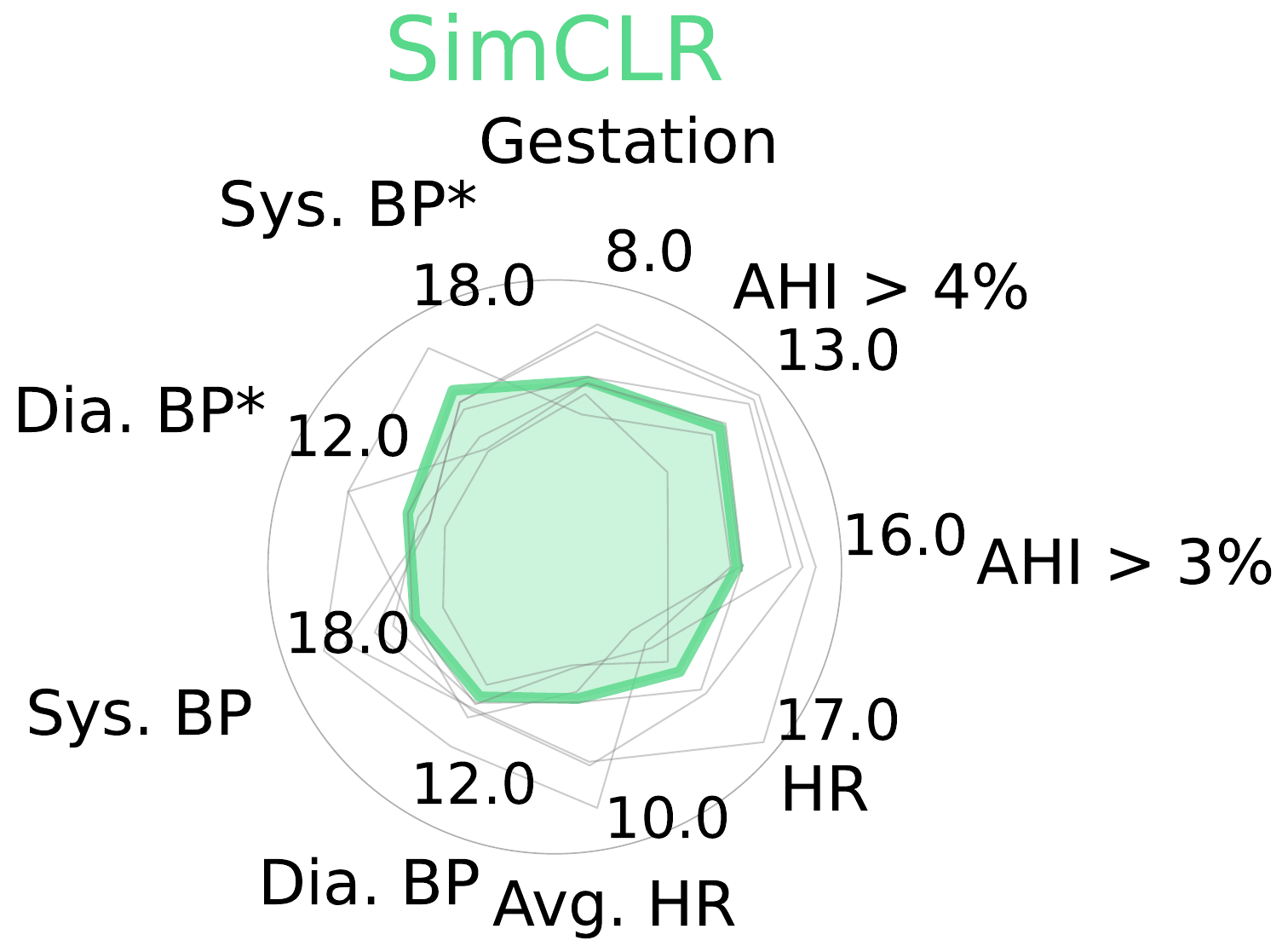}
     \end{subfigure}
     \hfill
     \begin{subfigure}[b]{0.24\textwidth}
         \centering
         \includegraphics[width=\textwidth]{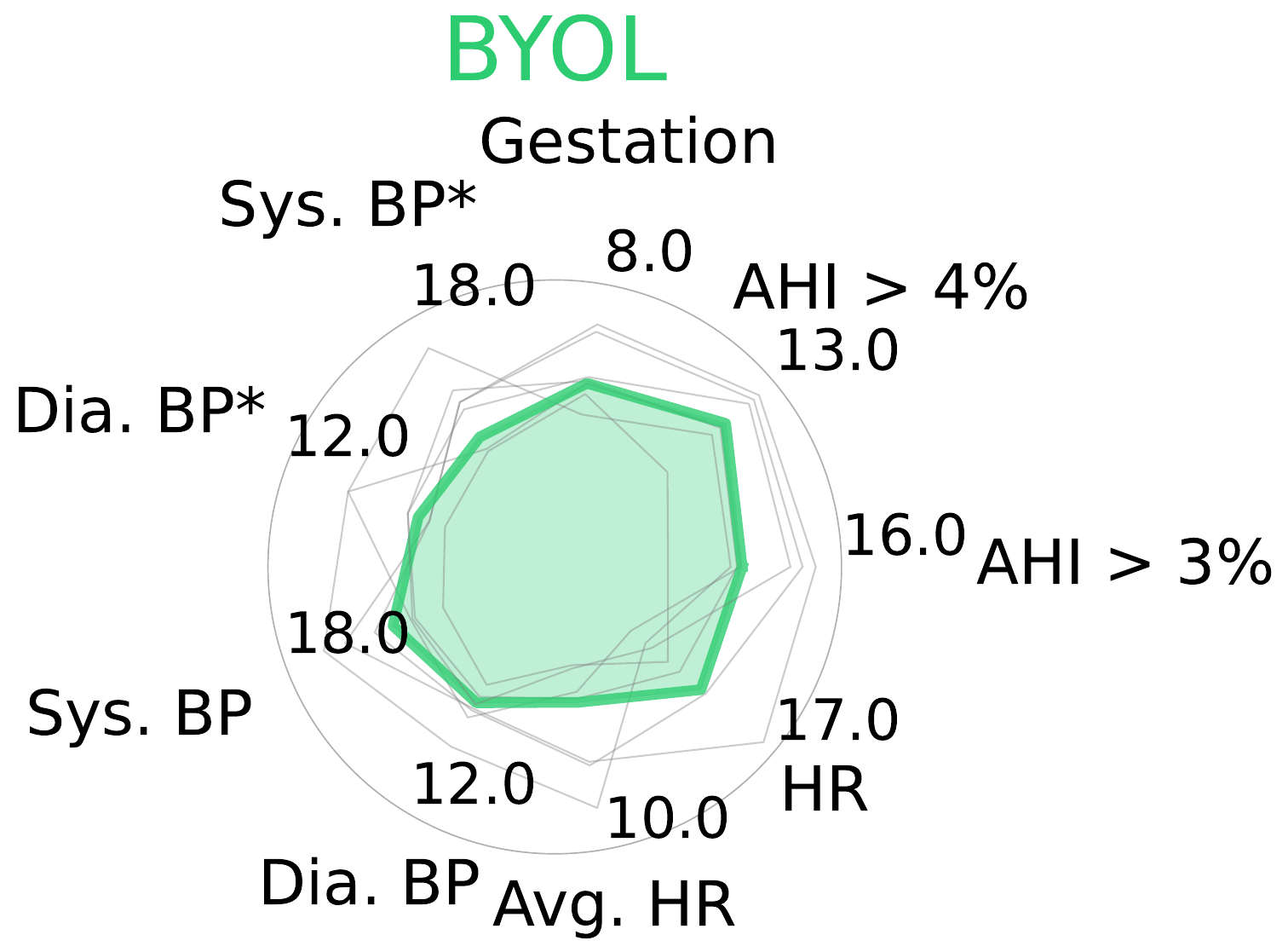}
     \end{subfigure}
     \hfill
     \begin{subfigure}[b]{0.24\textwidth}
         \centering
         \includegraphics[width=\textwidth]{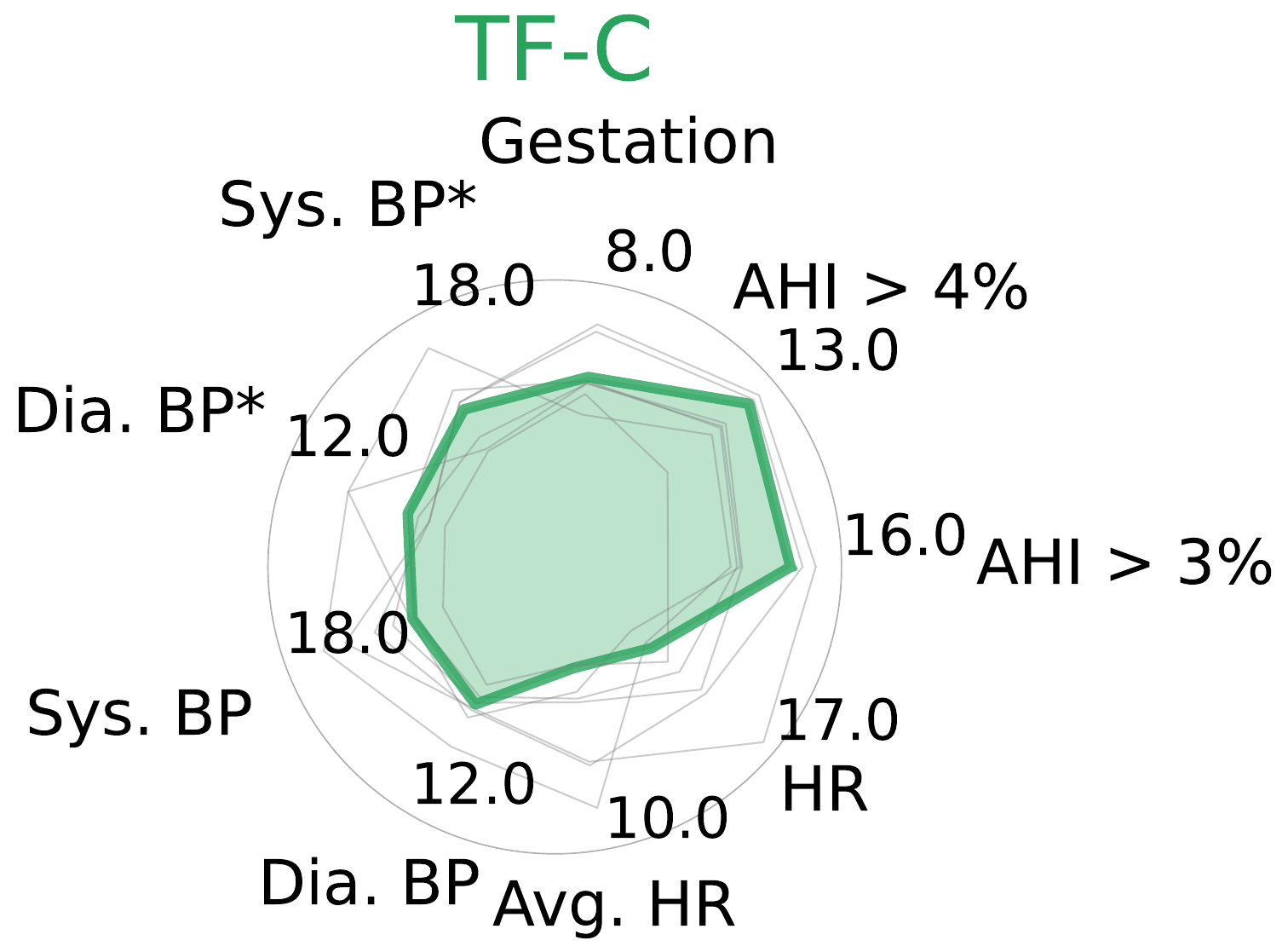}
     \end{subfigure}
     \hfill
     \begin{subfigure}[b]{0.24\textwidth}
         \centering
         \includegraphics[width=\textwidth]{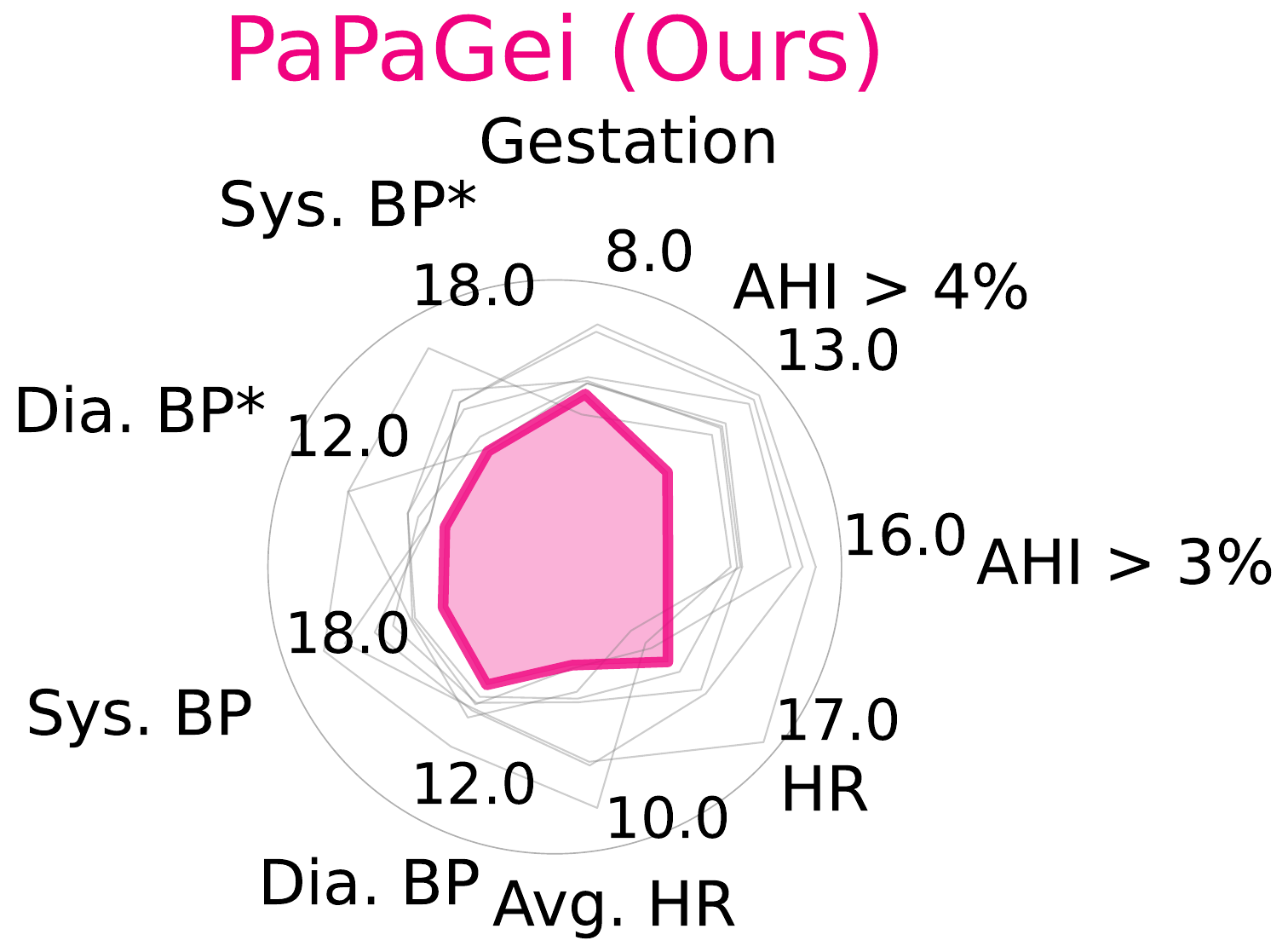}
     \end{subfigure}
    \caption{Radar charts of downstream tasks. (Top) \textit{Classification} performance in \textbf{AUROC (larger area is better)}. (Bottom) \textit{Regression} performance in \textbf{MAE (smaller area is better)}. Pre-trained models in purple: REGLE, Chronos, \& Moment. Statistical feature baseline in gray. SSL methods in green: SimCLR, BYOL, \& TF-C. \model{} (ours), in pink. Details are in Tables~\ref{tab:binary_classification_fm}~\&~\ref{tab:binary_classification_ssl}.}
    \label{fig:radar_regression}
    \vspace{-0.5cm}
\end{figure}
\vspace{-0.3cm}
\subsection{Baselines}
\vspace{-0.25cm}
We benchmark \model{}'s performance against competitive baselines. As open-source foundation models designed for physiological signals, \model{} is compared to recent time-series FMs: \textbf{Chronos} \citep{ansari2024chronos} and \textbf{MOMENT} \citep{goswami2024moment}. To evaluate the merits of our SSL framework, we also compare \model{} with common SSL methods (trained from scratch) such as \textbf{SimCLR} \citep{chen2020simple}, \textbf{BYOL} \citep{grill2020bootstrap}, and \textbf{TF-C} \citep{zhang2022self}. In addition, to assess model generalizability on PPG signals, we compare against \textbf{REGLE}, a model pre-trained on UK Biobank's PPG signals \citep{yun2024unsupervised}. As a simple baseline, we employ a random forest trained on \textbf{statistical features} extracted from the PPG signal, including mean, median, maximum, minimum, and the 25th, 50th, and 75th percentiles ("Stat. Features"). This task-specific approach serves as a benchmark for comparison with more advanced techniques.
 
\subsection{Linear Evaluation}
\vspace{-0.25cm}
Initially, we split the in-domain and out-of-domain datasets into training, validation, and test sets at 80/10/10 and 60/20/20 ratios. The splitting is performed at the subject level ensuring no overlap between individuals across the sets. The models are evaluated by extracting feature representations from resampled data (125Hz) and applying linear probing for each task. For binary classification tasks, we employ a logistic regression model, with performance measured by the AUROC score. For regression tasks, ridge regression is used, and performance is evaluated based on the mean absolute error (MAE). {Regression tasks are aggregated using the symmetric mean absolute percentage error (sMAPE)}. Multi-class classification tasks are trained using a random forest model, with accuracy as the evaluation metric. To ensure robustness, we compute 95\% confidence intervals through bootstrapping (500 sampling runs with replacement). More details are provided in the Appendix~\S \ref{appendix:hyperparameters}.

\vspace{-0.25cm}
\section{Results}
\label{sec:results}
\vspace{-0.25cm}
\subsection{Overall Performance} \label{sec:overall_performance}
\begin{table}[]
\centering
\caption{\textbf{Downstream comparison against pre-trained models.} Feature extraction parameters are indicated next to each name. 95\% CIs are reported in square brackets and the best value is \textbf{bolded}.}
\label{tab:binary_classification_fm}
\scalebox{0.68}{
\begin{tabular}{@{}l|lll|ll@{}}
\toprule
 &\cellcolor{violet!30}\textbf{REGLE} \smat{(0.07M)} &\cellcolor{violet!30}\textbf{Chronos} \smat{(200M)} &\cellcolor{violet!30}\textbf{Moment}  \smat{(385M)} &\cellcolor{pink!30}\textbf{\model{}-P} \smat{(5M)} &\cellcolor{pink!30}\textbf{\model{}-S} \smat{(5M)}  \\ 
\textbf{Classification} - AUROC ($\uparrow$) &\cellcolor{violet!30}\smat{\citep{yun2024unsupervised}} &\cellcolor{violet!30}\smat{\citep{ansari2024chronos}} &\cellcolor{violet!30}\smat{\citep{goswami2024moment}} &\cellcolor{pink!30} &\cellcolor{pink!30} \\\midrule
ICU Admission &0.57 \smat{[0.52-0.62]} &0.73 \smat{[0.68-0.80]} &0.72 \smat{[0.70-0.80]} &0.73 \smat{[0.67-0.78]} &\textbf{0.79} \smat{[0.75-0.82]} \\
Mortality &0.55 \smat{[0.52-0.59]} &\textbf{0.68} \smat{[0.65-0.71]} &0.67 \smat{[0.63-0.71]} &0.67 \smat{[0.63-0.71]} &0.67 \smat{[0.63-0.70]} \\
Smoker &0.54 \smat{[0.47-0.59]} &0.62 \smat{[0.57-0.67]} &0.62 \smat{[0.56-0.67]} &\textbf{0.64} \smat{[0.58-0.69]} &0.61 \smat{[0.56-0.66]} \\
Pregnancy stage &0.64 \smat{[0.57-0.63]} &\textbf{0.81} \smat{[0.79-0.82]} &0.76 \smat{[0.74-0.78]} &0.74 \smat{[0.72-0.76]} &0.78 \smat{[0.75-0.80]} \\
Hypertension &0.47 \smat{[0.34-0.58]} &0.57 \smat{[0.43-0.71]} &0.75 \smat{[0.64-0.85]} &0.74 \smat{[0.55-0.90]} &\textbf{0.77} \smat{[0.68-0.87]} \\
Sleep Disordered Breathing &0.45 \smat{[0.30-0.61]} &0.58 \smat{[0.35-0.82]} &0.45 \smat{[0.23-0.66]} &0.54 \smat{[0.23-0.66]} &\textbf{0.70} \smat{[0.57-0.84]} \\
Mood Disturbance &0.41 \smat{[0.16-0.66]} &0.43 \smat{[0.21-0.68]} &0.55 \smat{[0.33-0.78]} &0.53 \smat{[0.27-0.78]} &\textbf{0.56} \smat{[0.33-0.77]} \\
Valence &0.55 \smat{[0.52-0.57]} &0.56 \smat{[0.53-0.59]} &\textbf{0.57} \smat{[0.54-0.59]} &0.53 \smat{[0.51-0.56]} &0.56 \smat{[0.54-0.59]} \\
Arousal &0.51 \smat{[0.52-0.58]} &0.57 \smat{[0.54-0.60]} &0.56 \smat{[0.53-0.58]} &\textbf{0.58} \smat{[0.55-0.61]} &0.55 \smat{[0.52-0.57]} \\
\midrule
Average &0.52 $\pm$ 0.06 &0.62 $\pm$ 0.10 &0.63 $\pm$ 0.09 &0.63 $\pm$ 0.08 &\textbf{0.67 $\pm$ 0.09}\\
\midrule
\textbf{Regression} - MAE ($\downarrow$)& & & & & \\
\midrule
Apnea/Hypopnea Index $>$ 3\% &15.54 \smat{[14.20-16.69]} &14.06 \smat{[13.05-15.16]} &14.23 \smat{[13.04-15.42]} &13.85 \smat{[12.43-15.49]} &\textbf{12.97} \smat{[11.87-14.05]} \\
Apnea/Hypopnea Index $>$ 4\% &12.64 \smat{[11.47-13.78]} &11.57 \smat{[10.51-12.72]} &11.80 \smat{[10.79-12.93]} &11.24 \smat{[9.71-12.87]} &\textbf{10.56} \smat{[9.59-11.62]} \\
Gestation Age &7.28 \smat{[7.16-7.39]} &\textbf{5.69} \smat{[5.54-5.85]} &6.24 \smat{[6.10-6.37]} &6.40 \smat{[6.21-6.59]} &6.05 \smat{[5.91-6.17]} \\
Systolic BP (VV) &15.88 \smat{[13.67-18.36]} &17.24 \smat{[14.57-20.13]} &14.71 \smat{[12.38-17.29]} &19.11 \smat{[16.26-22.23]} &\textbf{14.65} \smat{[12.50-16.78]} \\
Diastolic BP (VV) &8.65 \smat{[7.16-10.27]} &10.53 \smat{[8.91-12.19]} &10.53 \smat{[8.91-12.19]} &10.87 \smat{[9.10-12.98]} &\textbf{8.29} \smat{[6.61-10.22]} \\
Systolic BP (PPG-BP) &16.32 \smat{[13.87-19.13]} &16.91 \smat{[13.31-19.34]} &14.50 \smat{[11.98-17.31]} &\textbf{13.60} \smat{[10.65-16.51]} &14.39 \smat{[12.53-16.45]} \\
Diastolic BP (PPG-BP) &9.30 \smat{[7.94-10.87]} &10.26 \smat{[8.13-12.57]} &9.53 \smat{[8.28-10.96]} & 8.88 \smat{[7.33-10.76]} &\textbf{8.71} \smat{[7.18-10.01]} \\
Average HR &6.88 \smat{[5.81-8.12]} &8.51 \smat{[7.05-10.07]} &4.41 \smat{[3.48-5.48]} &\textbf{3.47} \smat{[2.74-4.32]} & 4.00 \smat{[3.34-4.67]} \\
HR &16.35 \smat{[16.20-16.50]} &9.65 \smat{[9.50-9.79]} &\textbf{8.82} \smat{[8.68-8.96]} &10.92 \smat{[10.80-11.04]} &11.53 \smat{[11.40-11.66]} \\
\midrule
Average MAE \smat{(sMAPE)} &12.09 $\pm$ 3.83 \smat{(15.23\%)} &11.60 $\pm$ 3.60 \smat{(14.20\%)} &10.43 $\pm$ 3.46 \smat{(13.82\%)} &10.92 $\pm$ 4.25 \smat{(14.09\%)} &\textbf{10.12 $\pm$ 3.47} \smat{(13.34\%)} \\
\bottomrule
\end{tabular}}
\vspace{-0.3cm}
\end{table}
\vspace{-0.25cm}
In general, from Figure \ref{fig:radar_regression}, we observe that \model{} is more accurate across many tasks indicated by the larger AUROC area and smaller MAE area. Table \ref{tab:binary_classification_fm} presents a more detailed comparison 
 \begin{wrapfigure}{r}{0.49\textwidth}
    \centering    \includegraphics[width=\linewidth]{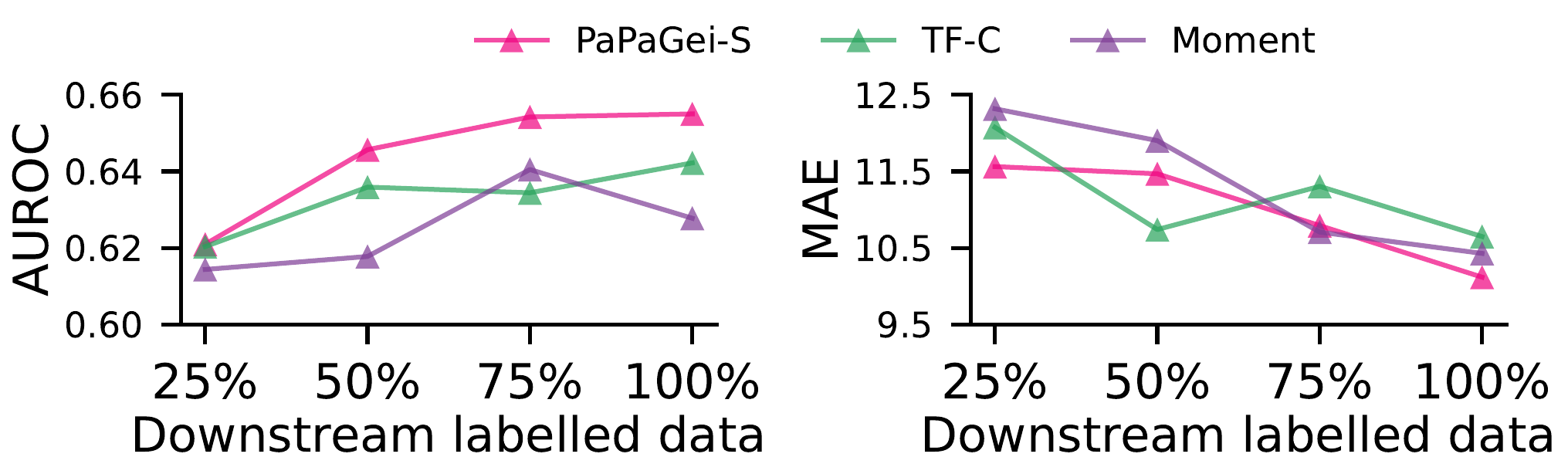}
    \caption{Downstream data-efficiency analysis. Results are averaged over all binary classification (left) and regression tasks (right). \model{}-S performs better with increased label availability.} 
    \label{fig:data_efficiency}
\end{wrapfigure}
between \model{} and other pre-trained models. For classification tasks, PaPaGei-S achieves the highest average AUROC of 0.67, outperforming other models across several tasks, particularly in ICU Admission~(0.79), Hypertension~(0.77), and Sleep Disordered Breathing~(0.70). In regression tasks, PaPaGei-S again demonstrates strong performance, achieving the lowest average MAE~(10.12), particularly in tasks related to Apnea/Hypopnea Index and BP measurements. REGLE, a small model trained on a large PPG dataset, generally underperforms compared to other models, suggesting its compact size may limit learning complex patterns. {Chronos obtains good performance in predicting mortality, pregnancy stage, and smoking, likely due to their slower rate of change and reduced reliance on granular PPG-specific features. General-purpose models suffice for these high-level outcomes. However, tasks requiring finer PPG-specific granularity, such as heart rate prediction, blood pressure estimation, or sleep apnea, benefit from \model{}'s specialized feature extraction.} Notably, \model{}-S consistently outperforms \model{}-P, highlighting the advantages of signal morphology objectives in enhancing predictive accuracy.

 Table \ref{tab:binary_classification_ssl} presents a comparison against three SSL methods and a baseline model trained on statistical features. In classification tasks, PaPaGei-S again shows the highest average AUROC, outperforming all others. SimCLR, BYOL, and TF-C generally outperform the statistical feature baseline but fall short of PaPaGei-S's performance. TF-C shows competitive results in some tasks, achieving the highest AUROC for Mood Disturbance and Valence. For regression tasks, PaPaGei-S again achieves the lowest average MAE. SimCLR, BYOL, and TF-C show mixed results, as each excels in different tasks. SimCLR comes second in estimating Avg HR, while BYOL does so in Systolic BP (VV). The statistical feature baseline generally underperforms compared to the advanced methods across most tasks. PaPaGei-P, while not consistently outperforming PaPaGei-S, shows strong results that are often competitive with or better than other contrastive learning methods. Overall, both PaPaGei variants offer robust performance across a wide range of tasks.
\begin{table}[]
\centering
\caption{\textbf{Downstream comparison against baseline and SSL methods.} Feature extraction parameters are indicated next to each name. 95\% CIs are reported in square brackets and the best value is \textbf{bolded}. Implementation details are in Appendix~\S \ref{appendix:hyperparameters}.}
\label{tab:binary_classification_ssl}
\scalebox{0.58}{
\begin{tabular}{@{}l|l|lll|ll@{}}
\toprule
 &\cellcolor{gray!30}\textbf{Stat. Features} &\cellcolor{green!30}\textbf{SimCLR} \smat{(5M)} &\cellcolor{green!30}\textbf{BYOL} \smat{(5M)} &\cellcolor{green!30}\textbf{TF-C} \smat{(10M)} &\cellcolor{pink!30}\textbf{\model{}-P \smat{(5M)}} &\cellcolor{pink!30}\textbf{\model{}-S \smat{(5M)}} \\ 
 \textbf{Classification} - AUROC ($\uparrow$) &\cellcolor{gray!30} &\cellcolor{green!30}\smat{\citep{chen2020simple}} &\cellcolor{green!30}\smat{\citep{grill2020bootstrap}} &\cellcolor{green!30}\smat{\citep{zhang2022self}} &\cellcolor{pink!20} &\cellcolor{pink!30} \\ \midrule
ICU Admission &0.71 \smat{[0.65-0.78]} &0.75 \smat{[0.72-0.79]} &0.78 \smat{[0.73-0.81]} &0.71 \smat{[0.67-0.75]} &0.73 \smat{[0.67-0.78]} &\textbf{0.79} \smat{[0.75-0.82]} \\
Mortality &0.57 \smat{[0.54-0.61]} &0.67 \smat{[0.63-0.70]} &0.67 \smat{[0.64-0.71]}  &0.67 \smat{[0.63-0.70]} &0.67 \smat{[0.63-0.71]} &0.67 \smat{[0.63-0.70]} \\
Smoker &0.63 \smat{[0.58-0.67]} &0.62 \smat{[0.57-0.68]} &0.62 \smat{[0.57-0.68]} &0.61 \smat{[0.56-0.67]} &\textbf{0.64} \smat{[0.58-0.69]} &0.61 \smat{[0.56-0.66]} \\
Pregnancy stage &0.64 \smat{[0.62-0.67]} &0.74 \smat{[0.72-0.75]} &0.62 \smat{[0.57-0.68]} &0.74 \smat{[0.72-0.76]} &0.74 \smat{[0.72-0.76]} &\textbf{0.78} \smat{[0.75-0.80]} \\
Hypertension &0.66 \smat{[0.47-0.83]} &0.75 \smat{[0.64-0.86]} &0.74 \smat{[0.64-0.84]} &0.76 \smat{[0.63-0.86]} &0.74 \smat{[0.55-0.90]} &\textbf{0.77} \smat{[0.68-0.87]} \\
SDB &0.32 \smat{[0.14-0.55]} &0.61 \smat{[0.46-0.76]} &0.59 \smat{[0.42-0.74]} &0.58 \smat{[0.44-0.73]} &0.54 \smat{[0.23-0.66]} &\textbf{0.70} \smat{[0.57-0.84]} \\
Mood Disturbance &0.54 \smat{[0.31-0.77]} &0.32 \smat{[0.12-0.55]} &0.46 \smat{[0.21-0.71]} &\textbf{0.59} \smat{[0.33-0.84]} &0.53 \smat{[0.27-0.78]} &0.56 \smat{[0.33-0.77]} \\
Valence &0.52 \smat{[0.49-0.55]} &0.52 \smat{[0.49-0.55]} &0.53 \smat{[0.50-0.56]} &\textbf{0.57} \smat{[0.54-0.59]} &0.53 \smat{[0.51-0.56]} &0.56 \smat{[0.54-0.59]} \\
Arousal &0.55 \smat{[0.53-0.58]} &0.55 \smat{[0.52-0.58]} &0.54 \smat{[0.30-0.78]} &0.55 \smat{[0.52-0.58]} &\textbf{0.58} \smat{[0.55-0.61]} &0.55 \smat{[0.52-0.57]} \\
\midrule
Average &0.57 $\pm$ 0.11 &0.61 $\pm$ 0.13 &0.62 $\pm$ 0.10  &0.64 $\pm$ 0.07 &0.63 $\pm$ 0.08 &\textbf{0.67 $\pm$ 0.09}\\
\midrule
\textbf{Regression} - MAE ($\downarrow$) & & & & & & \\
\midrule
Apnea/Hypopnea Index $>$ 3\% &15.31 \smat{[13.63-17.14]} &14.17 \smat{[13.04-15.38]} &14.26 \smat{[13.10-15.57]}  &15.10 \smat{[13.84-16.40]} &13.85 \smat{[12.43-15.49]} &\textbf{12.97} \smat{[11.87-14.05]} \\
Apnea/Hypopnea Index $>$ 4\% &12.52 \smat{[10.92-14.14]} &11.76 \smat{[10.65-12.89]} &11.88 \smat{[10.71-13.05]} &12.41 \smat{[11.33-13.49]} &11.24 \smat{[9.71-12.87]} &\textbf{10.56} \smat{[9.59-11.62]} \\
Gestation Age &7.15 \smat{[6.99-7.34]} &6.28 \smat{[6.21-6.49]} &6.24 \smat{[6.09-6.38]} &6.35 \smat{[6.21-6.49]} &6.40 \smat{[6.21-6.59]} &\textbf{6.05} \smat{[5.91-6.17]} \\
Systolic BP (VV) &15.76 \smat{[13.67-18.36]} &16.18 \smat{[13.73-18.85]} &15.01 \smat{[12.32-17.80]} &15.70 \smat{[13.23-18.13]} &19.11 \smat{[16.26-22.23]} &\textbf{14.65} \smat{[12.50-16.78]} \\
Diastolic BP (VV) &9.75 \smat{[7.16-11.27]} &9.15 \smat{[7.65-10.65]} &8.91 \smat{[7.48-10.43]} &9.15 \smat{[7.65-10.65]} &10.87 \smat{[9.10-12.98]} &\textbf{8.29} \smat{[6.61-10.22]} \\
Systolic BP (PPG-BP) &15.50 \smat{[11.68-20.25]} &14.38 \smat{[11.80-16.88]} &14.99 \smat{[13.03-17.38]} &14.45 \smat{[12.20-17.00]} &\textbf{13.60} \smat{[10.65-16.51]} &14.39 \smat{[12.53-16.45]} \\
Diastolic BP (PPG-BP) &9.35 \smat{[7.44-11.66]}  &9.01 \smat{[7.90-10.60]} &9.16 \smat{[8.00-10.50]} &9.20 \smat{[7.90-10.60]} & 8.88 \smat{[7.33-10.76]} &\textbf{8.71} \smat{[7.18-10.01]} \\
Average HR &7.01 \smat{[5.48-8.89]} &4.65 \smat{[3.99-5.39]} &4.78 \smat{[3.88-5.93]} &3.58 \smat{[2.90-4.21]} &\textbf{3.47} \smat{[2.74-4.32]} & 4.00 \smat{[3.34-4.67]} \\
HR &13.07 \smat{[12.90-13.23]} &11.59 \smat{[11.46-11.72]} &12.80 \smat{[12.66-12.94]} &\textbf{9.99} \smat{[9.86-10.12]} &10.92 \smat{[10.80-11.04]} &11.53 \smat{[11.40-11.66]} \\
\midrule
Average MAE \smat{(sMAPE)} &11.60 $\pm$ 3.41 \smat{(15.12\%)} &10.79 $\pm$ 3.63 \smat{(13.91\%)} &10.89 $\pm$ 3.58 \smat{(14.05\%)} & 10.65 $\pm$ 3.88 \smat{(14.07\%)} &10.92 $\pm$ 4.25 \smat{(14.09\%)} &\textbf{10.12 $\pm$ 3.47} \smat{(13.34\%)} \\ 
     \bottomrule
\end{tabular}
}
\vspace{-0.5cm}
\end{table}

\vspace{-0.25cm}
\subsection{Ablation Studies} \label{sec:ablation_study}
\vspace{-0.2cm}
\noindent\textbf{Pre-training data ablation.} We evaluate \model{}-S using different pre-training data 
\begin{wrapfigure}{r}{0.60\textwidth}
    \centering
    \includegraphics[width=\linewidth]{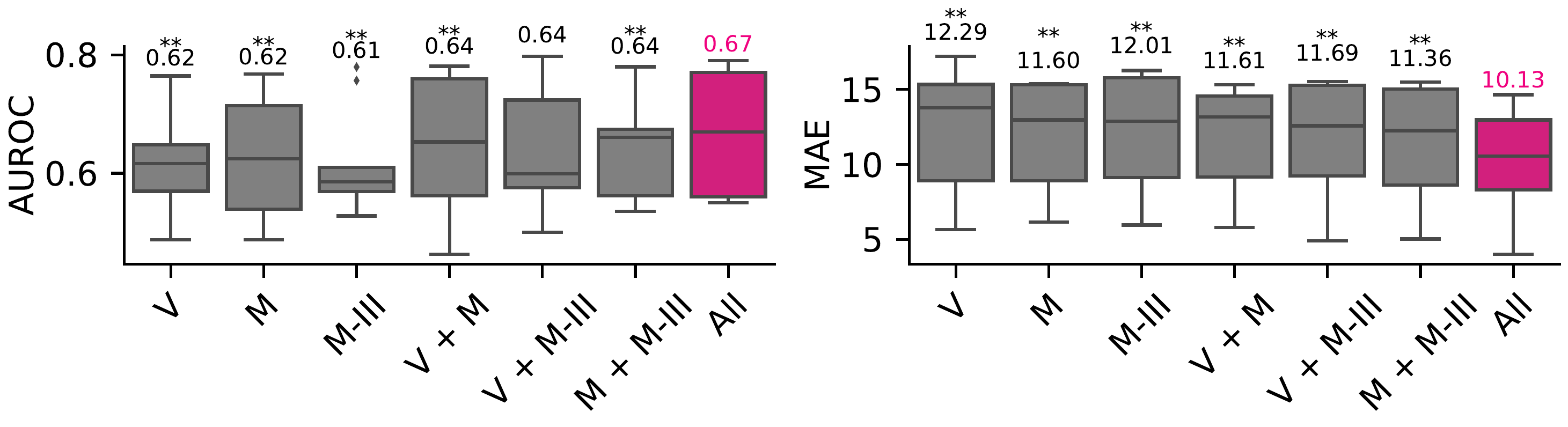}
    \caption{Ablation on pre-training data. Average performance across tasks for models trained on: V (VitalDB), M (MESA), and M-III (MIMIC-III). The mean value is displayed above the plots. {The Wilcoxon signed rank test is applied to evaluate significance between the All dataset and the rest ($**:p < 0.05 \;\text{and}\; *:0.05\leq p < 0.10$).}}
    \label{fig:pretraining_data_ablation}
     \vspace{-0.2cm}
\end{wrapfigure} 
combinations. As shown in Figure  \ref{fig:pretraining_data_ablation}, performance on downstream tasks improves with more upstream data, with the best results achieved when using all three datasets. Notably, MESA outperforms the others despite having the fewest participants but the highest number of segments. This supports findings from language models \citep{dubey2024llama} and wearable sensing research \citep{narayanswamy2024scaling}, indicating that the volume of segments or hours contributes more to performance than the number of users.

\begin{figure}[htbp!]
    \centering
    \begin{subfigure}[b]{0.25\textwidth}
         \includegraphics[width=\textwidth]{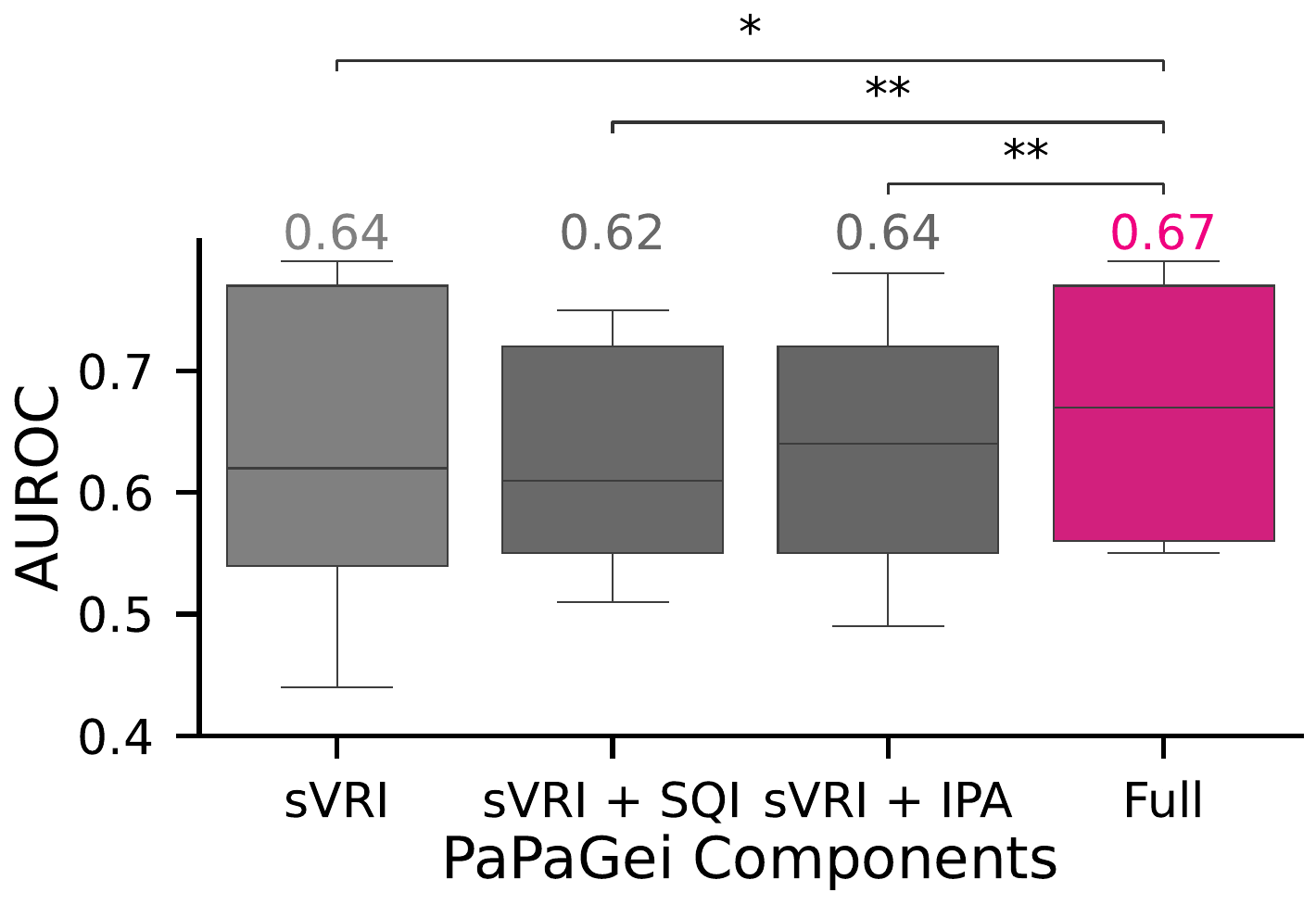}
         \caption{}
         \label{fig:ablation_component_a}
     \end{subfigure}
     \begin{subfigure}[b]{0.23\textwidth}
         \includegraphics[width=\textwidth]{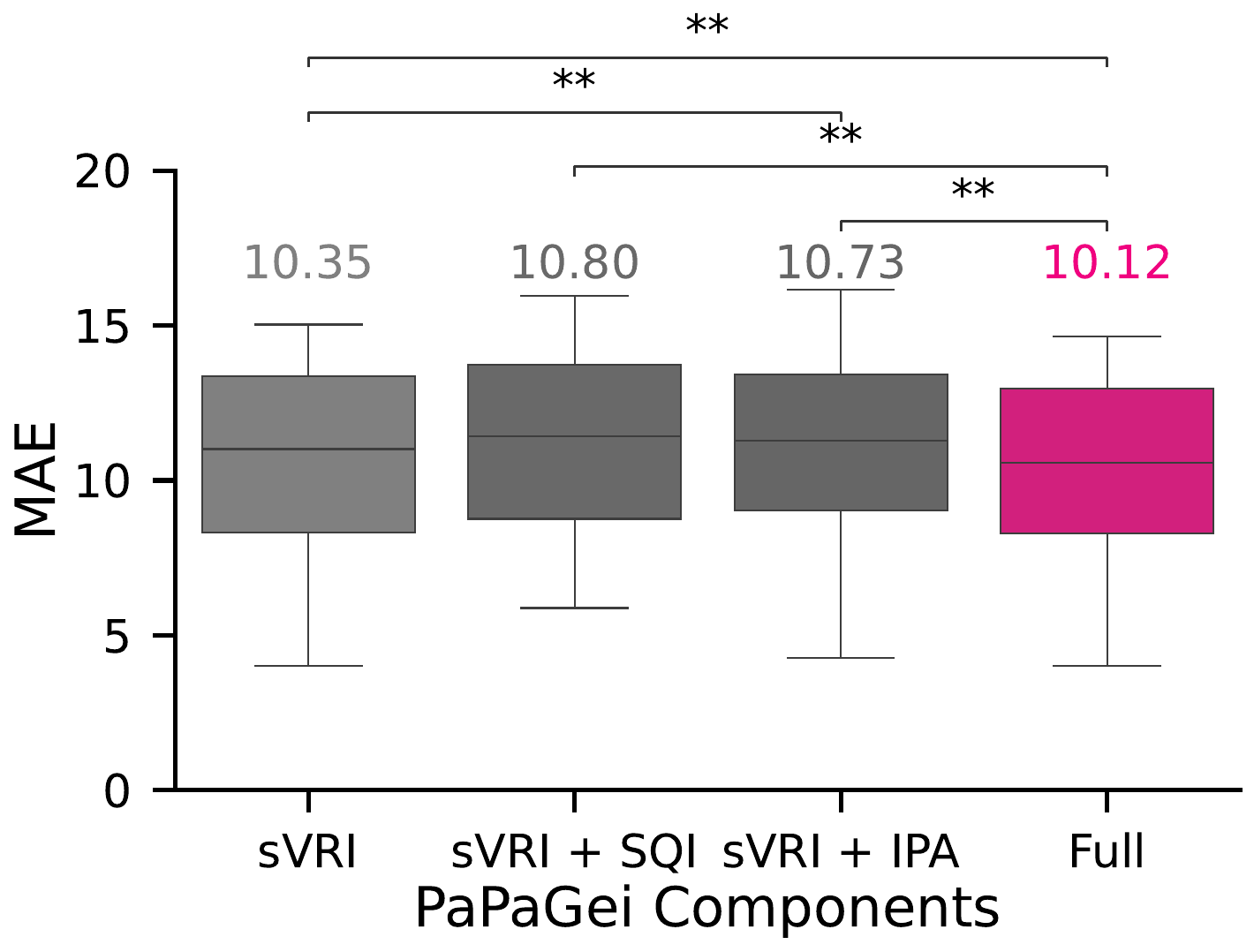}
         \caption{}
         \label{fig:ablation_component_b}
     \end{subfigure}
     \hfill
    \begin{subfigure}[b]{0.49\textwidth}
         \includegraphics[width=\textwidth]{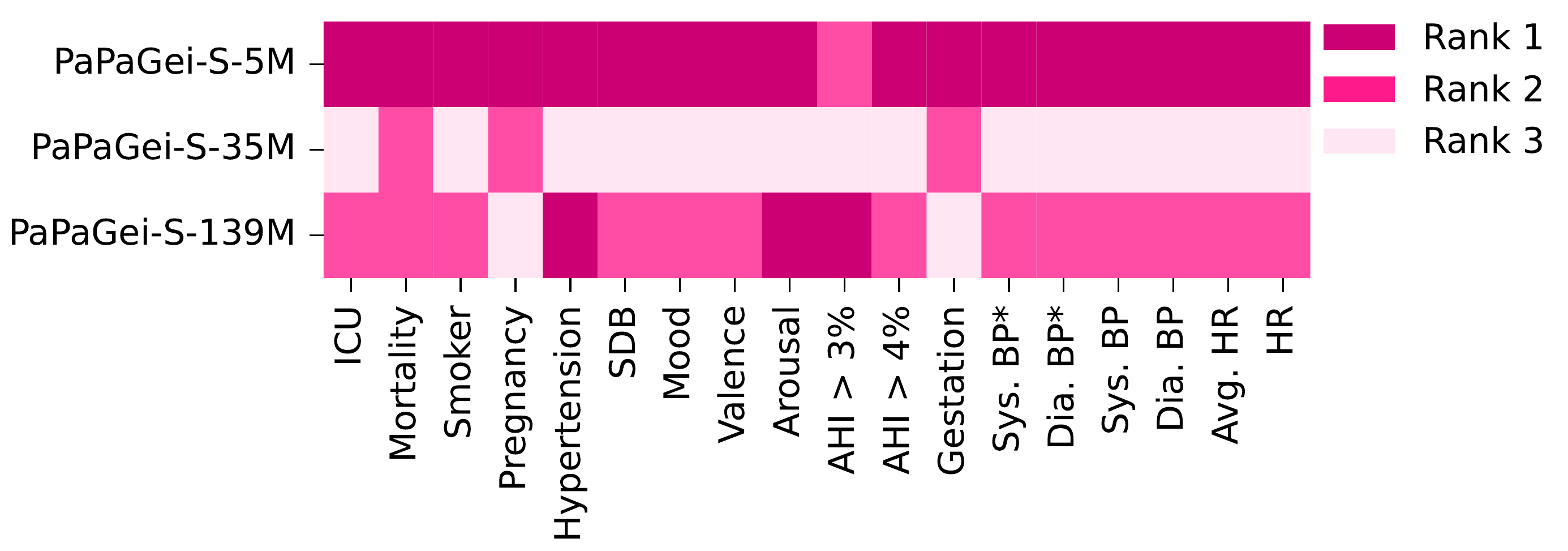}
         \caption{}
         \label{fig:heatmap}
     \end{subfigure}
     \hfill
    \caption{\model{}-S component ablation study (a, b) and scaling analysis (c). (Left) The boxplot shows the performance of \model{}-S components across all tasks. {The Wilcoxon signed rank test is applied to evaluate pair-wise significance ($**:p < 0.05 \;\text{and}\; *:0.05\leq p<0.10$)}. (Right) Heatmap ranks of \model{}-S models with 5M, 35M, and 139M parameters (rank 1 denotes the best performance). Detailed results in Table~\ref{tab:scaling_analysis}.}
    \label{fig:component_scaling}
\end{figure}

\noindent\textbf{\model{}-S component ablation.} We assess the impact of \model{}-S components. Figure \ref{fig:component_scaling} shows that the full model (0.67, 10.12) consistently outperforms individual components in both mean and median metrics. On average, sVRI (0.64, 10.35) outperforms the combinations of sVRI + SQI (0.62, 10.80) and sVRI + IPA (0.64, 10.73). {Our results indicate that combining SQI and IPA yields greater benefits compared to their individual contributions.}

\noindent\textbf{Downstream data-efficiency analysis.} For limited-data scenarios, we assess the performance of downstream linear probing across varying levels of labeled data availability. We compare to the second best-performing baselines from Tables \ref{tab:binary_classification_fm} \& \ref{tab:binary_classification_ssl}, namely TF-C and Moment. As shown in Figure \ref{fig:data_efficiency}, the classification performance of \model{}-S steadily improves as more labeled data becomes available. While TF-C and Moment also show performance gains between 25\% and 100\% labeled data, their improvements are less consistent and smaller than \model{}-S. In regression tasks, \model{}-S achieves the lowest MAE at both 25\% and 100\% data availability, consistently reducing errors. At the middle breakpoints, the results are mixed with TF-C and Moment being competitive. 

\vspace{-0.1cm}
\noindent\textbf{Model size and scaling analysis.} We investigated the impact of model size on performance by training \model{}-S-35M and \model{}-S-139M with 35M and 139M parameters, respectively. 
\begin{wrapfigure}{r}{0.44\textwidth}
    \centering
    \includegraphics[width=0.7\linewidth]{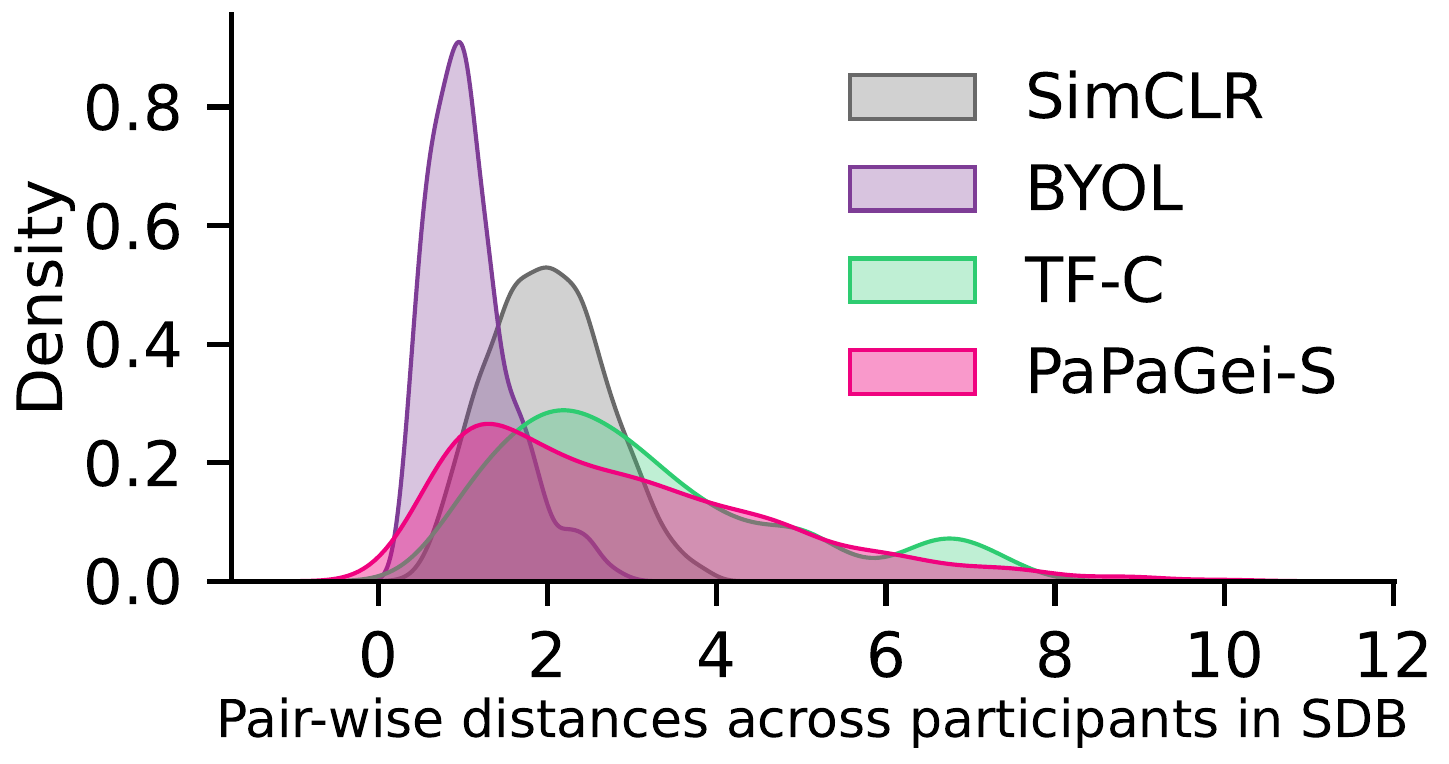}
    \caption{Pair-wise inter-participant embedding distances for SDB.}
    \label{fig:inter-participant}
\end{wrapfigure}
Both models share the same number of layers, but the 35M model uses a 32-filter size while the 139M model uses 64. As shown in \ref{fig:heatmap}, the smallest model (5M parameters) consistently outperformed larger models on all but one task. {This suggests the 5M model is better suited for our pre-training datasets, aligning with prior findings on the proportionality between data and model size \citep{narayanswamy2024scaling}.} While the 139M model surpassed the 35M, it still lagged behind the 5M, indicating that wider models may improve performance in classification tasks, likely due to the contrastive learning objective. {Nevertheless, our scaling analysis shows a non-monotonic trend, indicating other factors strongly influence performance.}

{\noindent\textbf{Effect of Demographics.} We evaluate the effect of demographics (age, sex) and PPG-specific features (sVRI, IPA, SQI) in Appendix~\S \ref{appendix:demographics}. In demographics prediction (Table \ref{tab:demographic_targets}), \model{}-S achieves 7.78 MAE in age regression, 0.85 accuracy in age classification, and 0.79 accuracy in sex classification. While our results trail larger closed studies \citep{abbaspourazad2023large} by 2.18, 0.05, and 0.13 for segment-level SSL, and by 5.59, 0.12, and 0.25 for patient-level SSL, they mark an advancement in open-source efforts. These findings indicate that patient-level positive pair selection in SSL better captures demographic-related features for downstream prediction. Moreover, the reduced performance of \model{}-S can be attributed to evaluations conducted on diverse device setups, as opposed to a single device configuration. Our ablation study (Table \ref{tab:demo_analysis}) shows that \model{}-S outperforms the demo + PPG in 14 out of 18 tasks, particularly in tasks with real-time dependence such as heart rate estimation. Importantly, including demographics in addition to \model{}-S creates a stronger model. These findings emphasize that \textbf{demographic features complement rather than compete with \model{}-S}, showcasing the potential of integrating \model{}'s advanced feature extraction capabilities with demographic context to improve task outcomes.}
\vspace{-0.1cm}
\subsection{Case Studies} \label{sec:case_studies}
\vspace{-0.15cm}

\noindent\textbf{Inter-participant embeddings.} Figure \ref{fig:inter-participant} shows the distribution of pair-wise embedding distances across participants in the SDB dataset~\citep{kiyasseh2021clocs}. SimCLR and BYOL exhibit sharper peaks at lower distances, indicating that participants are more closely clustered within the embedding space. This could be interpreted as a mild form of mode collapse, where the model does not fully capture the individual differences between participants. TF-C demonstrates a more balanced distribution, with both large and small peaks, suggesting it captures both similarities and some variation between participants. In contrast, \model{}-S provides the widest dispersion of embeddings, highlighting its ability to capture a broader range of features that may be valuable for distinguishing between participants' medical conditions. 

\begin{figure}
  \centering
    {\includegraphics[width=0.20\textwidth]{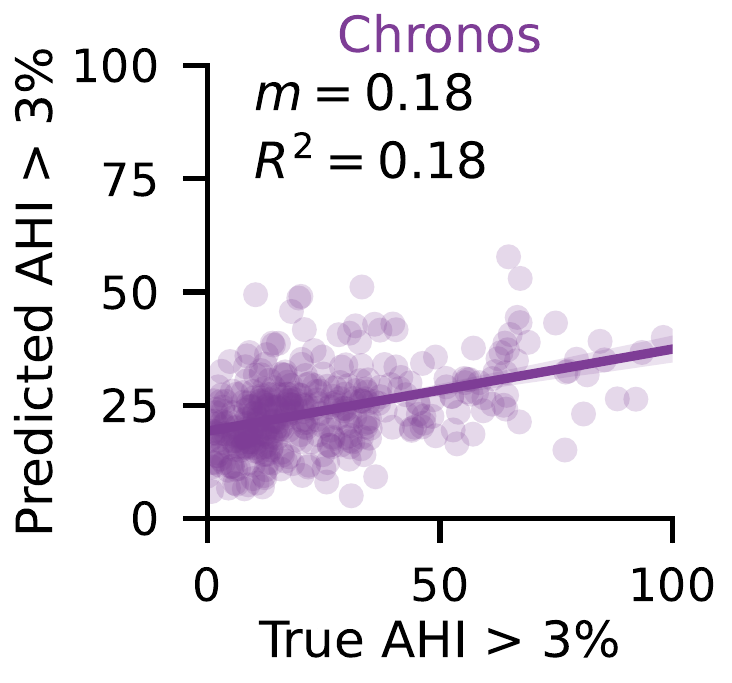}\includegraphics[width=0.18\textwidth]{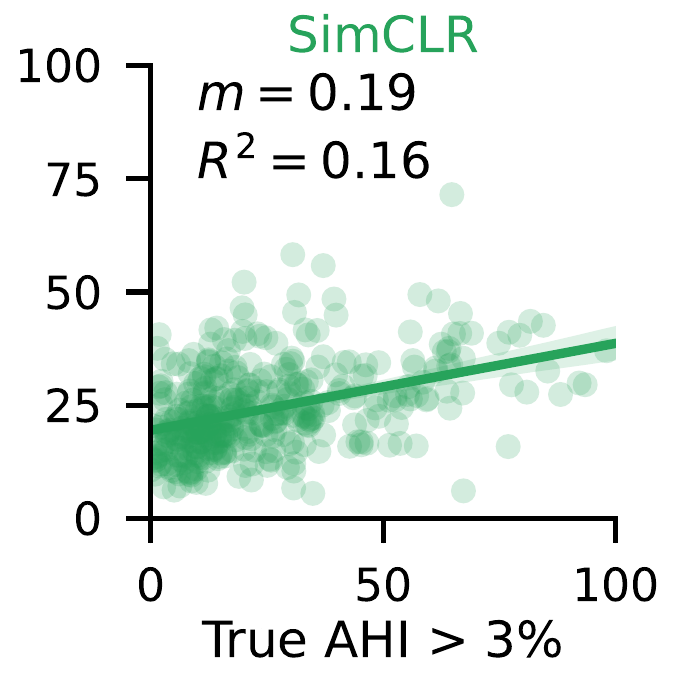} \includegraphics[width=0.18\textwidth]{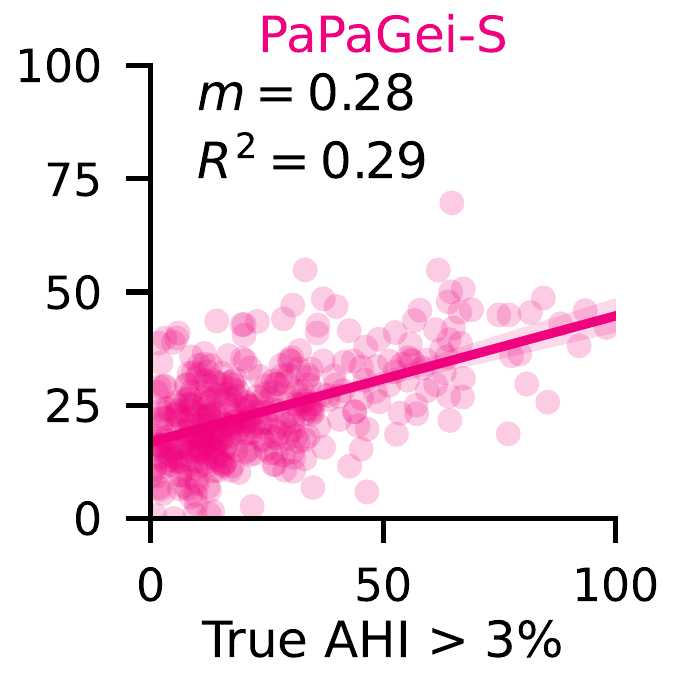}
    \includegraphics[width=0.40\textwidth]{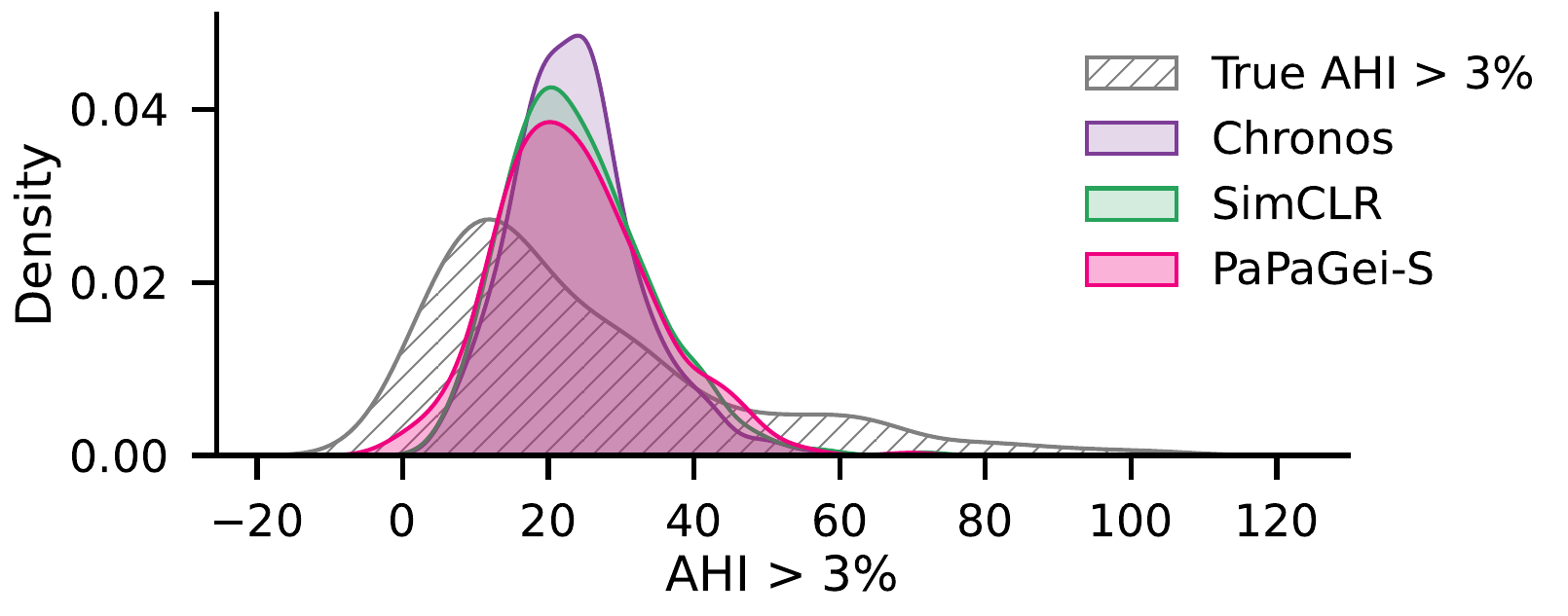}} \label{fig:predictions_a}%
  \caption{Regression plots and prediction distribution of different models compared to ground truth for AHI $>$ 3\%. $R^2$ is the coefficient of determination and $m$ is the correlation slope.}
  \label{fig:predictions}
  \vspace{-0.4cm}
\end{figure}

\noindent\textbf{Regression predictions.} {From Figure \ref{fig:predictions}, compared to the pre-trained and SSL baseline, \model{}-S demonstrates steeper slopes ($m$) and higher $R^2$ values, reflecting a stronger alignment between predictions and true values. Additionally, the prediction distribution for AHI indicates that SimCLR and Chronos tend to regress more toward the mean, while \model{}-S achieves a wider distribution base, highlighting its capacity to capture left tail better. Additional plots are shown in Appendix~\S \ref{appendix:prediction_plots}.}

\begin{wrapfigure}{r}{0.41\textwidth}
    \centering
    \includegraphics[width=0.8\linewidth]{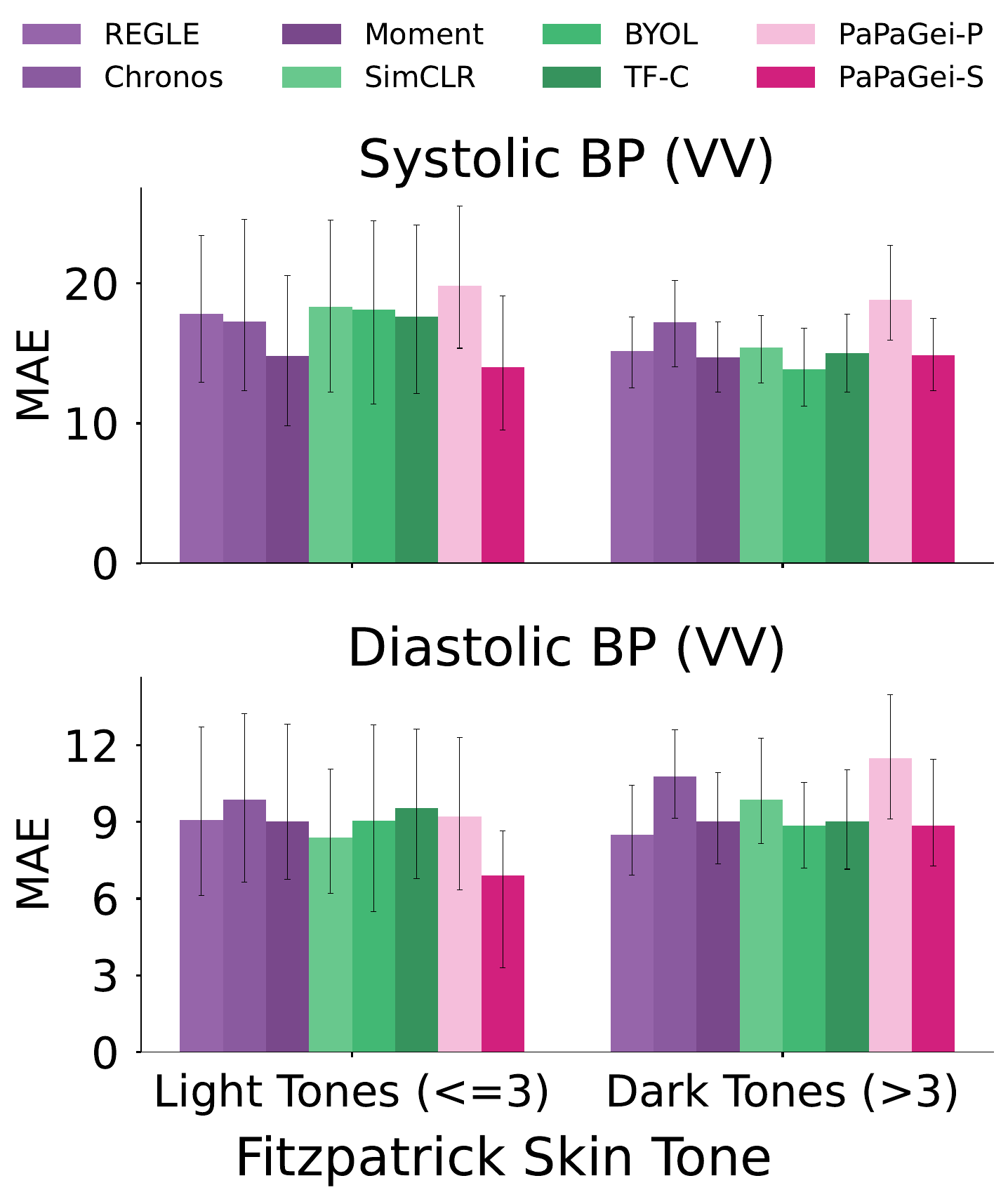}
    \caption{Skin tone analysis for Blood Pressure estimation (VV dataset)}.
    \label{fig:fitzpatrick_bnw}
\end{wrapfigure}

\noindent\textbf{Skin tone analysis.} We examine BP estimation performance across skin tones because it is crucial for practical use \citep{bent2020investigating}. As shown in Figure \ref{fig:fitzpatrick_bnw} (More details in Figure \ref{fig:fitzpatrick_detailed}), \model{}-S achieves the best BP estimation across light tones. Across dark tones, we notice that BYOL and REGLE obtain the lowest MAE for Systolic BP and Diastolic BP. {However, identifying a single model that performs best across all skin tones remains challenging.} While \model{}-S obtains the best overall performance, additional work is necessary to improve robustness on darker skin tones.

\vspace{-0.2cm}
\section{Discussion \& Conclusion}
\vspace{-0.2cm}
Our results show that \model{} outperforms baselines in at least 14 tasks, with classification and regression improvements of 4.7\%-6.3\% and 2.9\%-4.9\%, respectively. \model{}-S excelled in cardiovascular tasks like BP, Hypertension, and HR, which can be attributed to the sVRI and IPA objectives, and \model{}-P outperformed baselines like Moment, excelling in tasks such as Smoking and Arousal. Ablation studies confirmed that the model with all three SSL objectives performs best, with sVRI highlighted as a key component and IPA and SQI providing positive knowledge transfer in multi-task setups. {To assess performance under class imbalance, we examined the F1-score. PaPaGei achieves the highest F1 in 6 out of 9 classification tasks, demonstrating its effectiveness in handling data imbalance. For regression, PaPaGei-S achieves the highest $R^2$ in 7 tasks (Appendix~\S \ref{appendix:additional_results}), reflecting better alignment with the true distribution. These results highlight the robustness and versatility of PaPaGei-S across classification and regression tasks.} \model{} is both data- and size-efficient (5M), making it ideal for medical applications where large models (200M+) are impractical due to on-device limitations or data privacy concerns with cloud model inference. While combining \model{}-P and \model{}-S objectives into one model might seem intuitive, it is impractical because it would constrain positive pairs on both sVRI and the number of participants, resulting in too many unique labels with limited samples per label. Our case studies also showed that \model{}-S captured personal medical information due to well-dispersed embeddings, compared to baselines. Future work should focus on diversifying training data, investigating sampling rate effects, and exploring multi-modal approaches or alternative architectures. Additionally, as extracting PPG features for different morphologies is non-trivial, future work benefit from systematic evaluation of PPG features and modeling. In conclusion, \model{} represents a significant advancement in foundation models for analyzing PPG signals in resource-constrained medical environments, with its open-source nature encouraging further research and development in healthcare applications.

\newpage
\section*{Reproducibility Statement} 
Models, data, and code are publicly available for reproducibility and future research.We exclusively utilize publicly accessible datasets, which can be requested or downloaded from the respective study group websites, allowing others to easily obtain the data for their own analyses. In  \S\ref{sec:experiments} and Appendix~\S\ref{appendix:datasets_tasks}, we provide comprehensive descriptions of the datasets, ground-truth annotations, and data pre-processing methods used in our experiments, ensuring transparency in our data handling procedures. The code to run our model is published with user-friendly examples. We have provided a detailed overview of the model architecture and its hyperparameters in \S\ref{sec:methods}, \S\ref{sec:pretraining}, and Appendix~\S \ref{appendix:hyperparameters}. Thus, our work is designed to be reproducible, enabling future research to build upon our findings.

\section*{Ethics Statement} 

Our research on \model{}, utilizing publicly available PPG datasets, adheres to data privacy regulations and promotes transparency through open-source releases. We acknowledge potential biases in the training data and have evaluated performance across diverse datasets, particularly regarding skin tone variations. While PaPaGei offers significant potential for improving non-invasive health monitoring, we recognize the need to address potential misuse~\citep{perez2021digital}. Examples of misuse could include unauthorized health monitoring, discriminatory practices in insurance or employment, unfair credit scoring, or exploiting personal health data for targeted marketing. We strongly advocate responsible use solely for beneficial healthcare applications. Our study followed established research ethics guidelines, and we declare no conflicts of interest. We encourage ongoing interdisciplinary dialogue to address potential risks and ensure responsible development and deployment of such technologies, recognizing the broader societal impacts of AI in healthcare. We remain committed to ethical AI advancement and welcome further discussion on the critical issues, including the development of governance frameworks to prevent misuse and protect data privacy.

\bibliography{iclr2025_conference}

\begin{thebibliography}{79}
\providecommand{\natexlab}[1]{#1}
\providecommand{\url}[1]{\texttt{#1}}
\expandafter\ifx\csname urlstyle\endcsname\relax
  \providecommand{\doi}[1]{doi: #1}\else
  \providecommand{\doi}{doi: \begingroup \urlstyle{rm}\Url}\fi

\bibitem[Abbaspourazad et~al.(2023)Abbaspourazad, Elachqar, Miller, Emrani, Nallasamy, and Shapiro]{abbaspourazad2023large}
Salar Abbaspourazad, Oussama Elachqar, Andrew~C Miller, Saba Emrani, Udhyakumar Nallasamy, and Ian Shapiro.
\newblock Large-scale training of foundation models for wearable biosignals.
\newblock \emph{arXiv preprint arXiv:2312.05409}, 2023.

\bibitem[Afandizadeh~Zargari et~al.(2023)Afandizadeh~Zargari, Aqajari, Khodabandeh, Rahmani, and Kurdahi]{afandizadeh2023accurate}
Amir~Hosein Afandizadeh~Zargari, Seyed Amir~Hossein Aqajari, Hadi Khodabandeh, Amir Rahmani, and Fadi Kurdahi.
\newblock An accurate non-accelerometer-based ppg motion artifact removal technique using cyclegan.
\newblock \emph{ACM Transactions on Computing for Healthcare}, 4\penalty0 (1):\penalty0 1--14, 2023.

\bibitem[Ansari et~al.(2024)Ansari, Stella, Turkmen, Zhang, Mercado, Shen, Shchur, Rangapuram, Arango, Kapoor, et~al.]{ansari2024chronos}
Abdul~Fatir Ansari, Lorenzo Stella, Caner Turkmen, Xiyuan Zhang, Pedro Mercado, Huibin Shen, Oleksandr Shchur, Syama~Sundar Rangapuram, Sebastian~Pineda Arango, Shubham Kapoor, et~al.
\newblock Chronos: Learning the language of time series.
\newblock \emph{arXiv preprint arXiv:2403.07815}, 2024.

\bibitem[Ave et~al.(2015)Ave, Fauzan, Adhitya, and Zakaria]{ave2015early}
Arrozaq Ave, Hamdan Fauzan, S~Rhandy Adhitya, and Hasballah Zakaria.
\newblock Early detection of cardiovascular disease with photoplethysmogram (ppg) sensor.
\newblock In \emph{2015 international conference on electrical engineering and informatics (ICEEI)}, pp.\  676--681. IEEE, 2015.

\bibitem[Belyaeva et~al.(2023)Belyaeva, Cosentino, Hormozdiari, Eswaran, Shetty, Corrado, Carroll, McLean, and Furlotte]{belyaeva2023multimodal}
Anastasiya Belyaeva, Justin Cosentino, Farhad Hormozdiari, Krish Eswaran, Shravya Shetty, Greg Corrado, Andrew Carroll, Cory~Y McLean, and Nicholas~A Furlotte.
\newblock Multimodal llms for health grounded in individual-specific data.
\newblock In \emph{Workshop on Machine Learning for Multimodal Healthcare Data}, pp.\  86--102. Springer, 2023.

\bibitem[Bent et~al.(2020)Bent, Goldstein, Kibbe, and Dunn]{bent2020investigating}
Brinnae Bent, Benjamin~A Goldstein, Warren~A Kibbe, and Jessilyn~P Dunn.
\newblock Investigating sources of inaccuracy in wearable optical heart rate sensors.
\newblock \emph{NPJ digital medicine}, 3\penalty0 (1):\penalty0 18, 2020.

\bibitem[{BioBSS Documentation}(2023)]{biobss_documentation}
{BioBSS Documentation}.
\newblock Biobss: Biosignal processing toolbox.
\newblock \url{https://biobss.readthedocs.io/en/latest/}, 2023.
\newblock Accessed: 2024-09-10.

\bibitem[Bouariu et~al.(2022)Bouariu, Panaitescu, and Nicolaides]{bouariu2022first}
Alexandra Bouariu, Anca~Maria Panaitescu, and Kypros~H Nicolaides.
\newblock First trimester prediction of adverse pregnancy outcomes—identifying pregnancies at risk from as early as 11--13 weeks.
\newblock \emph{Medicina}, 58\penalty0 (3):\penalty0 332, 2022.

\bibitem[Bradley \& Lang(1994)Bradley and Lang]{bradley1994measuring}
Margaret~M Bradley and Peter~J Lang.
\newblock Measuring emotion: the self-assessment manikin and the semantic differential.
\newblock \emph{Journal of behavior therapy and experimental psychiatry}, 25\penalty0 (1):\penalty0 49--59, 1994.

\bibitem[Charlton et~al.(2023)Charlton, Allen, Bail{\'o}n, Baker, Behar, Chen, Clifford, Clifton, Davies, Ding, et~al.]{charlton20232023}
Peter~H Charlton, John Allen, Raquel Bail{\'o}n, Stephanie Baker, Joachim~A Behar, Fei Chen, Gari~D Clifford, David~A Clifton, Harry~J Davies, Cheng Ding, et~al.
\newblock The 2023 wearable photoplethysmography roadmap.
\newblock \emph{Physiological measurement}, 44\penalty0 (11):\penalty0 111001, 2023.

\bibitem[Chen et~al.(2021)Chen, Lundberg, Erion, Kim, and Lee]{chen2021forecasting}
Hugh Chen, Scott~M Lundberg, Gabriel Erion, Jerry~H Kim, and Su-In Lee.
\newblock Forecasting adverse surgical events using self-supervised transfer learning for physiological signals.
\newblock \emph{NPJ Digital Medicine}, 4\penalty0 (1):\penalty0 167, 2021.

\bibitem[Chen et~al.(2020)Chen, Kornblith, Norouzi, and Hinton]{chen2020simple}
Ting Chen, Simon Kornblith, Mohammad Norouzi, and Geoffrey Hinton.
\newblock A simple framework for contrastive learning of visual representations.
\newblock \emph{International conference on machine learning}, pp.\  1597--1607, 2020.

\bibitem[Chen et~al.(2015)Chen, Wang, Zee, Lutsey, Javaheri, Alc{\'a}ntara, Jackson, Williams, and Redline]{chen2015racial}
Xiaoli Chen, Rui Wang, Phyllis Zee, Pamela~L Lutsey, Sogol Javaheri, Carmela Alc{\'a}ntara, Chandra~L Jackson, Michelle~A Williams, and Susan Redline.
\newblock Racial/ethnic differences in sleep disturbances: the multi-ethnic study of atherosclerosis (mesa).
\newblock \emph{Sleep}, 38\penalty0 (6):\penalty0 877--888, 2015.

\bibitem[Cheng et~al.(2020)Cheng, Goh, Dogrusoz, Tuzel, and Azemi]{cheng2020subject}
Joseph~Y Cheng, Hanlin Goh, Kaan Dogrusoz, Oncel Tuzel, and Erdrin Azemi.
\newblock Subject-aware contrastive learning for biosignals.
\newblock \emph{arXiv preprint arXiv:2007.04871}, 2020.

\bibitem[Chung et~al.(2020)Chung, Kung, Chang, and Tsai]{chung2020demographics}
Wei-Sheng Chung, Pei-Tseng Kung, Hui-Yun Chang, and Wen-Chen Tsai.
\newblock Demographics and medical disorders associated with smoking: a population-based study.
\newblock \emph{BMC Public Health}, 20:\penalty0 1--8, 2020.

\bibitem[Crump et~al.(2023)Crump, Sundquist, McLaughlin, Dolan, Govindarajulu, Sieh, and Sundquist]{crump2023adverse}
Casey Crump, Jan Sundquist, Mary~Ann McLaughlin, Siobhan~M Dolan, Usha Govindarajulu, Weiva Sieh, and Kristina Sundquist.
\newblock Adverse pregnancy outcomes and long term risk of ischemic heart disease in mothers: national cohort and co-sibling study.
\newblock \emph{bmj}, 380, 2023.

\bibitem[Dem{\v{s}}ar(2006)]{demvsar2006statistical}
Janez Dem{\v{s}}ar.
\newblock Statistical comparisons of classifiers over multiple data sets.
\newblock \emph{The Journal of Machine learning research}, 7:\penalty0 1--30, 2006.

\bibitem[Ding et~al.(2024)Ding, Guo, Chen, Lee, Rudin, and Hu]{ding2024siamquality}
Cheng Ding, Zhicheng Guo, Zhaoliang Chen, Randall~J Lee, Cynthia Rudin, and Xiao Hu.
\newblock Siamquality: a convnet-based foundation model for photoplethysmography signals.
\newblock \emph{Physiological Measurement}, 45\penalty0 (8):\penalty0 085004, 2024.

\bibitem[Dubey et~al.(2024)Dubey, Jauhri, Pandey, Kadian, Al-Dahle, Letman, Mathur, Schelten, Yang, Fan, et~al.]{dubey2024llama}
Abhimanyu Dubey, Abhinav Jauhri, Abhinav Pandey, Abhishek Kadian, Ahmad Al-Dahle, Aiesha Letman, Akhil Mathur, Alan Schelten, Amy Yang, Angela Fan, et~al.
\newblock The llama 3 herd of models.
\newblock \emph{arXiv preprint arXiv:2407.21783}, 2024.

\bibitem[Elgendi(2016)]{elgendi2016optimal}
Mohamed Elgendi.
\newblock Optimal signal quality index for photoplethysmogram signals.
\newblock \emph{Bioengineering}, 3\penalty0 (4):\penalty0 21, 2016.

\bibitem[Facco et~al.(2015)Facco, Parker, Reddy, Silver, Louis, Basner, Chung, Schubert, Pien, Redline, et~al.]{facco2015numom2b}
Francesca~L Facco, Corette~B Parker, Uma~M Reddy, Robert~M Silver, Judette~M Louis, Robert~C Basner, Judith~H Chung, Frank~P Schubert, Grace~W Pien, Susan Redline, et~al.
\newblock Numom2b sleep-disordered breathing study: objectives and methods.
\newblock \emph{American journal of obstetrics and gynecology}, 212\penalty0 (4):\penalty0 542--e1, 2015.

\bibitem[Feli et~al.(2024)Feli, Azimi, Sarhaddi, Sharifi-Heris, Niela-Vilen, Liljeberg, Axelin, and Rahmani]{feli2024preterm}
Mohammad Feli, Iman Azimi, Fatemeh Sarhaddi, Zahra Sharifi-Heris, Hannakaisa Niela-Vilen, Pasi Liljeberg, Anna Axelin, and Amir~M Rahmani.
\newblock Preterm birth risk stratification through longitudinal heart rate and hrv monitoring in daily life.
\newblock 2024.

\bibitem[Gao et~al.(2021)Gao, Cui, Wan, Zheng, and Gu]{gao2021ecsmp}
Zhilin Gao, Xingran Cui, Wang Wan, Wenming Zheng, and Zhongze Gu.
\newblock Ecsmp: A dataset on emotion, cognition, sleep, and multi-model physiological signals.
\newblock \emph{Data in Brief}, 39:\penalty0 107660, 2021.

\bibitem[Garde et~al.(2014)Garde, Dehkordi, Karlen, Wensley, Ansermino, and Dumont]{garde2014development}
Ainara Garde, Parastoo Dehkordi, Walter Karlen, David Wensley, J~Mark Ansermino, and Guy~A Dumont.
\newblock Development of a screening tool for sleep disordered breathing in children using the phone oximeter™.
\newblock \emph{PloS one}, 9\penalty0 (11):\penalty0 e112959, 2014.

\bibitem[Goswami et~al.(2024)Goswami, Szafer, Choudhry, Cai, Li, and Dubrawski]{goswami2024moment}
Mononito Goswami, Konrad Szafer, Arjun Choudhry, Yifu Cai, Shuo Li, and Artur Dubrawski.
\newblock Moment: A family of open time-series foundation models.
\newblock \emph{arXiv preprint arXiv:2402.03885}, 2024.

\bibitem[Grill et~al.(2020)Grill, Strub, Altch{\'e}, Tallec, Richemond, Buchatskaya, Doersch, Pires, Guo, Azar, et~al.]{grill2020bootstrap}
Jean-Bastien Grill, Florian Strub, Florent Altch{\'e}, Corentin Tallec, Pierre~H Richemond, Elena Buchatskaya, Carl Doersch, Bernardo~Avila Pires, Zhaohan~Daniel Guo, Mohammad~Gheshlaghi Azar, et~al.
\newblock Bootstrap your own latent-a new approach to self-supervised learning.
\newblock \emph{Advances in neural information processing systems}, 33:\penalty0 21271--21284, 2020.

\bibitem[Gruver et~al.(2024)Gruver, Finzi, Qiu, and Wilson]{gruver2024large}
Nate Gruver, Marc Finzi, Shikai Qiu, and Andrew~G Wilson.
\newblock Large language models are zero-shot time series forecasters.
\newblock \emph{Advances in Neural Information Processing Systems}, 36, 2024.

\bibitem[Haddad et~al.(2021)Haddad, Boukhayma, and Caizzone]{haddad2021continuous}
Serj Haddad, Assim Boukhayma, and Antonino Caizzone.
\newblock Continuous ppg-based blood pressure monitoring using multi-linear regression.
\newblock \emph{IEEE journal of biomedical and health informatics}, 26\penalty0 (5):\penalty0 2096--2105, 2021.

\bibitem[He et~al.(2020)He, Fan, Wu, Xie, and Girshick]{he2020momentum}
Kaiming He, Haoqi Fan, Yuxin Wu, Saining Xie, and Ross Girshick.
\newblock Momentum contrast for unsupervised visual representation learning.
\newblock In \emph{Proceedings of the IEEE/CVF conference on computer vision and pattern recognition}, pp.\  9729--9738, 2020.

\bibitem[He et~al.(2022)He, Chen, Xie, Li, Doll{\'a}r, and Girshick]{he2022masked}
Kaiming He, Xinlei Chen, Saining Xie, Yanghao Li, Piotr Doll{\'a}r, and Ross Girshick.
\newblock Masked autoencoders are scalable vision learners.
\newblock \emph{Proceedings of the IEEE/CVF conference on computer vision and pattern recognition}, pp.\  16000--16009, 2022.

\bibitem[Johnson et~al.(2016)Johnson, Pollard, Shen, Lehman, Feng, Ghassemi, Moody, Szolovits, Anthony~Celi, and Mark]{johnson2016mimic}
Alistair~EW Johnson, Tom~J Pollard, Lu~Shen, Li-wei~H Lehman, Mengling Feng, Mohammad Ghassemi, Benjamin Moody, Peter Szolovits, Leo Anthony~Celi, and Roger~G Mark.
\newblock Mimic-iii, a freely accessible critical care database.
\newblock \emph{Scientific data}, 3\penalty0 (1):\penalty0 1--9, 2016.

\bibitem[Kiyasseh et~al.(2020)Kiyasseh, Tadesse, Thwaites, Zhu, Clifton, et~al.]{kiyasseh2020plethaugment}
Dani Kiyasseh, Girmaw~Abebe Tadesse, Louise Thwaites, Tingting Zhu, David Clifton, et~al.
\newblock Plethaugment: Gan-based ppg augmentation for medical diagnosis in low-resource settings.
\newblock \emph{IEEE journal of biomedical and health informatics}, 24\penalty0 (11):\penalty0 3226--3235, 2020.

\bibitem[Kiyasseh et~al.(2021)Kiyasseh, Zhu, and Clifton]{kiyasseh2021clocs}
Dani Kiyasseh, Tingting Zhu, and David~A Clifton.
\newblock Clocs: Contrastive learning of cardiac signals across space, time, and patients.
\newblock In \emph{International Conference on Machine Learning}, pp.\  5606--5615. PMLR, 2021.

\bibitem[Koteska et~al.(2022)Koteska, Bodanova, Mitrova, Sidorenko, and Lehocki]{koteska2022deep}
Bojana Koteska, Ana~Madevska Bodanova, Hristina Mitrova, Marija Sidorenko, and Fedor Lehocki.
\newblock A deep learning approach to estimate spo2 from ppg signals.
\newblock In \emph{Proceedings of the 9th International Conference on Bioinformatics Research and Applications}, pp.\  142--148, 2022.

\bibitem[Lai et~al.(2023)Lai, Tan, Wang, Ji, Guo, Han, Shi, Feng, and Yang]{lai2023practical}
Jiewei Lai, Huixin Tan, Jinliang Wang, Lei Ji, Jun Guo, Baoshi Han, Yajun Shi, Qianjin Feng, and Wei Yang.
\newblock Practical intelligent diagnostic algorithm for wearable 12-lead ecg via self-supervised learning on large-scale dataset.
\newblock \emph{Nature Communications}, 14\penalty0 (1):\penalty0 3741, 2023.

\bibitem[Lapitan et~al.(2024)Lapitan, Rogatkin, Molchanova, and Tarasov]{lapitan2024estimation}
Denis~G Lapitan, Dmitry~A Rogatkin, Elizaveta~A Molchanova, and Andrey~P Tarasov.
\newblock Estimation of phase distortions of the photoplethysmographic signal in digital iir filtering.
\newblock \emph{Scientific Reports}, 14\penalty0 (1):\penalty0 6546, 2024.

\bibitem[Lee et~al.(2022)Lee, Park, Yoon, Yang, Park, and Jung]{lee2022vitaldb}
Hyung-Chul Lee, Yoonsang Park, Soo~Bin Yoon, Seong~Mi Yang, Dongnyeok Park, and Chul-Woo Jung.
\newblock Vitaldb, a high-fidelity multi-parameter vital signs database in surgical patients.
\newblock \emph{Scientific Data}, 9\penalty0 (1):\penalty0 279, 2022.

\bibitem[Liang et~al.(2018{\natexlab{a}})Liang, Chen, Liu, and Elgendi]{liang2018new}
Yongbo Liang, Zhencheng Chen, Guiyong Liu, and Mohamed Elgendi.
\newblock A new, short-recorded photoplethysmogram dataset for blood pressure monitoring in china.
\newblock \emph{Scientific data}, 5\penalty0 (1):\penalty0 1--7, 2018{\natexlab{a}}.

\bibitem[Liang et~al.(2018{\natexlab{b}})Liang, Chen, Ward, and Elgendi]{liang2018hypertension}
Yongbo Liang, Zhencheng Chen, Rabab Ward, and Mohamed Elgendi.
\newblock Hypertension assessment using photoplethysmography: a risk stratification approach.
\newblock \emph{Journal of clinical medicine}, 8\penalty0 (1):\penalty0 12, 2018{\natexlab{b}}.

\bibitem[Liang et~al.(2018{\natexlab{c}})Liang, Elgendi, Chen, and Ward]{liang2018optimal}
Yongbo Liang, Mohamed Elgendi, Zhencheng Chen, and Rabab Ward.
\newblock An optimal filter for short photoplethysmogram signals.
\newblock \emph{Scientific data}, 5\penalty0 (1):\penalty0 1--12, 2018{\natexlab{c}}.

\bibitem[Lyu et~al.(2015)Lyu, Luo, Zhou, Yu, Miao, Wang, Shi, and Kameyama]{lyu2015measuring}
Yongqiang Lyu, Xiaomin Luo, Jun Zhou, Chun Yu, Congcong Miao, Tong Wang, Yuanchun Shi, and Ken-ichi Kameyama.
\newblock Measuring photoplethysmogram-based stress-induced vascular response index to assess cognitive load and stress.
\newblock In \emph{Proceedings of the 33rd annual ACM conference on human factors in computing systems}, pp.\  857--866, 2015.

\bibitem[McKeen et~al.(2024)McKeen, Oliva, Masood, Toma, Rubin, and Wang]{mckeen2024ecg}
Kaden McKeen, Laura Oliva, Sameer Masood, Augustin Toma, Barry Rubin, and Bo~Wang.
\newblock Ecg-fm: An open electrocardiogram foundation model.
\newblock \emph{arXiv preprint arXiv:2408.05178}, 2024.

\bibitem[Moody et~al.(2020)Moody, Moody, Villarroel, Clifford, and Silva]{moody2020mimic}
Benjamin Moody, George Moody, Mauricio Villarroel, Gari~D. Clifford, and Ikaro Silva.
\newblock Mimic-iii waveform database matched subset (version 1.0), 2020.
\newblock URL \url{https://doi.org/10.13026/c2294b}.

\bibitem[Moon et~al.(2023)Moon, Madotto, Lin, Nagarajan, Smith, Jain, Yeh, Murugesan, Heidari, Liu, et~al.]{moon2023anymal}
Seungwhan Moon, Andrea Madotto, Zhaojiang Lin, Tushar Nagarajan, Matt Smith, Shashank Jain, Chun-Fu Yeh, Prakash Murugesan, Peyman Heidari, Yue Liu, et~al.
\newblock Anymal: An efficient and scalable any-modality augmented language model.
\newblock \emph{arXiv preprint arXiv:2309.16058}, 2023.

\bibitem[Narayanswamy et~al.(2024)Narayanswamy, Liu, Ayush, Yang, Xu, Liao, Garrison, Tailor, Sunshine, Liu, et~al.]{narayanswamy2024scaling}
Girish Narayanswamy, Xin Liu, Kumar Ayush, Yuzhe Yang, Xuhai Xu, Shun Liao, Jake Garrison, Shyam Tailor, Jake Sunshine, Yun Liu, et~al.
\newblock Scaling wearable foundation models.
\newblock \emph{arXiv preprint arXiv:2410.13638}, 2024.

\bibitem[Oord et~al.(2018)Oord, Li, and Vinyals]{oord2018representation}
Aaron van~den Oord, Yazhe Li, and Oriol Vinyals.
\newblock Representation learning with contrastive predictive coding.
\newblock \emph{arXiv preprint arXiv:1807.03748}, 2018.

\bibitem[Orphanidou(2018)]{orphanidou2018quality}
Christina Orphanidou.
\newblock Quality assessment for the photoplethysmogram (ppg).
\newblock \emph{Signal Quality Assessment in Physiological Monitoring: State of the Art and Practical Considerations}, pp.\  41--63, 2018.

\bibitem[Parikh et~al.(2021)Parikh, Gonzalez, Anderson, Judd, Rexrode, Hlatky, Gunderson, Stuart, Vaidya, on~Epidemiology, Prevention; Council~on Arteriosclerosis, on~Cardiovascular, Nursing;, and the Stroke~Council]{parikh2021adverse}
Nisha~I Parikh, Juan~M Gonzalez, Cheryl~AM Anderson, Suzanne~E Judd, Kathryn~M Rexrode, Mark~A Hlatky, Erica~P Gunderson, Jennifer~J Stuart, Dhananjay Vaidya, American Heart Association~Council on~Epidemiology, Thrombosis Prevention; Council~on Arteriosclerosis, Vascular Biology;~Council on~Cardiovascular, Stroke Nursing;, and the Stroke~Council.
\newblock Adverse pregnancy outcomes and cardiovascular disease risk: unique opportunities for cardiovascular disease prevention in women: a scientific statement from the american heart association.
\newblock \emph{Circulation}, 143\penalty0 (18):\penalty0 e902--e916, 2021.

\bibitem[Paszke et~al.(2019)Paszke, Gross, Massa, Lerer, Bradbury, Chanan, Killeen, Lin, Gimelshein, Antiga, et~al.]{paszke2019pytorch}
Adam Paszke, Sam Gross, Francisco Massa, Adam Lerer, James Bradbury, Gregory Chanan, Trevor Killeen, Zeming Lin, Natalia Gimelshein, Luca Antiga, et~al.
\newblock Pytorch: An imperative style, high-performance deep learning library.
\newblock \emph{Advances in neural information processing systems}, 32, 2019.

\bibitem[Perez-Pozuelo et~al.(2021)Perez-Pozuelo, Spathis, Gifford-Moore, Morley, and Cowls]{perez2021digital}
Ignacio Perez-Pozuelo, Dimitris Spathis, Jordan Gifford-Moore, Jessica Morley, and Josh Cowls.
\newblock Digital phenotyping and sensitive health data: Implications for data governance.
\newblock \emph{Journal of the American Medical Informatics Association}, 28\penalty0 (9):\penalty0 2002--2008, 2021.

\bibitem[Reiss et~al.(2019)Reiss, Indlekofer, Schmidt, and Van~Laerhoven]{reiss2019deep}
Attila Reiss, Ina Indlekofer, Philip Schmidt, and Kristof Van~Laerhoven.
\newblock Deep ppg: Large-scale heart rate estimation with convolutional neural networks.
\newblock \emph{Sensors}, 19\penalty0 (14):\penalty0 3079, 2019.

\bibitem[Rice(2008)]{rice2008sisg}
Ken Rice.
\newblock \emph{Linear Models and Generalized Linear Models}, 2008.
\newblock URL \url{https://faculty.washington.edu/kenrice/sisg/SISG-08-06.pdf}.
\newblock SISG-08.

\bibitem[Ruehland et~al.(2009)Ruehland, Rochford, O’Donoghue, Pierce, Singh, and Thornton]{ruehland2009new}
Warren~R Ruehland, Peter~D Rochford, Fergal~J O’Donoghue, Robert~J Pierce, Parmjit Singh, and Andrew~T Thornton.
\newblock The new aasm criteria for scoring hypopneas: impact on the apnea hypopnea index.
\newblock \emph{sleep}, 32\penalty0 (2):\penalty0 150--157, 2009.

\bibitem[Sadad et~al.(2022)Sadad, Bukhari, Munir, Ghani, El-Sherbeeny, and Rauf]{sadad2022detection}
Tariq Sadad, Syed Ahmad~Chan Bukhari, Asim Munir, Anwar Ghani, Ahmed~M El-Sherbeeny, and Hafiz~Tayyab Rauf.
\newblock Detection of cardiovascular disease based on ppg signals using machine learning with cloud computing.
\newblock \emph{Computational Intelligence and Neuroscience}, 2022\penalty0 (1):\penalty0 1672677, 2022.

\bibitem[Sarkar \& Etemad(2020)Sarkar and Etemad]{sarkar2020self}
Pritam Sarkar and Ali Etemad.
\newblock Self-supervised ecg representation learning for emotion recognition.
\newblock \emph{IEEE Transactions on Affective Computing}, 13\penalty0 (3):\penalty0 1541--1554, 2020.

\bibitem[Schmidt et~al.(2018)Schmidt, Reiss, Duerichen, Marberger, and Van~Laerhoven]{schmidt2018introducing}
Philip Schmidt, Attila Reiss, Robert Duerichen, Claus Marberger, and Kristof Van~Laerhoven.
\newblock Introducing wesad, a multimodal dataset for wearable stress and affect detection.
\newblock In \emph{Proceedings of the 20th ACM international conference on multimodal interaction}, pp.\  400--408, 2018.

\bibitem[Schrumpf et~al.(2021)Schrumpf, Frenzel, Aust, Osterhoff, and Fuchs]{schrumpf2021assessment}
Fabian Schrumpf, Patrick Frenzel, Christoph Aust, Georg Osterhoff, and Mirco Fuchs.
\newblock Assessment of deep learning based blood pressure prediction from ppg and rppg signals.
\newblock In \emph{Proceedings of the IEEE/CVF conference on computer vision and pattern recognition}, pp.\  3820--3830, 2021.

\bibitem[Sohn(2016)]{sohn2016improved}
Kihyuk Sohn.
\newblock Improved deep metric learning with multi-class n-pair loss objective.
\newblock \emph{Advances in neural information processing systems}, 29, 2016.

\bibitem[Song et~al.(2024)Song, Jang, Lee, Hong, Kwon, and Jo]{song2024foundation}
Junho Song, Jong-Hwan Jang, Byeong~Tak Lee, DongGyun Hong, Joon-myoung Kwon, and Yong-Yeon Jo.
\newblock Foundation models for electrocardiograms.
\newblock \emph{arXiv preprint arXiv:2407.07110}, 2024.

\bibitem[Spathis \& Kawsar(2024)Spathis and Kawsar]{spathis2024first}
Dimitris Spathis and Fahim Kawsar.
\newblock The first step is the hardest: Pitfalls of representing and tokenizing temporal data for large language models.
\newblock \emph{Journal of the American Medical Informatics Association}, 31\penalty0 (9):\penalty0 2151--2158, 2024.

\bibitem[Spathis et~al.(2021)Spathis, Perez-Pozuelo, Brage, Wareham, and Mascolo]{spathis2021self}
Dimitris Spathis, Ignacio Perez-Pozuelo, Soren Brage, Nicholas~J Wareham, and Cecilia Mascolo.
\newblock Self-supervised transfer learning of physiological representations from free-living wearable data.
\newblock In \emph{Proceedings of the Conference on Health, Inference, and Learning}, pp.\  69--78, 2021.

\bibitem[Spathis et~al.(2022)Spathis, Perez-Pozuelo, Gonzales, Wu, Brage, Wareham, and Mascolo]{spathis2022longitudinal}
Dimitris Spathis, Ignacio Perez-Pozuelo, Tomas~I Gonzales, Yu~Wu, Soren Brage, Nicholas Wareham, and Cecilia Mascolo.
\newblock Longitudinal cardio-respiratory fitness prediction through wearables in free-living environments.
\newblock \emph{NPJ Digital Medicine}, 5\penalty0 (1):\penalty0 176, 2022.

\bibitem[Tang et~al.(2020)Tang, Perez-Pozuelo, Spathis, and Mascolo]{tang2020exploring}
Chi~Ian Tang, Ignacio Perez-Pozuelo, Dimitris Spathis, and Cecilia Mascolo.
\newblock Exploring contrastive learning in human activity recognition for healthcare.
\newblock \emph{arXiv preprint arXiv:2011.11542}, 2020.

\bibitem[Temko(2017)]{temko2017accurate}
Andriy Temko.
\newblock Accurate heart rate monitoring during physical exercises using ppg.
\newblock \emph{IEEE Transactions on Biomedical Engineering}, 64\penalty0 (9):\penalty0 2016--2024, 2017.

\bibitem[Tonekaboni et~al.(2021)Tonekaboni, Eytan, and Goldenberg]{tonekaboni2021unsupervised}
Sana Tonekaboni, Danny Eytan, and Anna Goldenberg.
\newblock Unsupervised representation learning for time series with temporal neighborhood coding.
\newblock \emph{arXiv preprint arXiv:2106.00750}, 2021.

\bibitem[Toye(2023)]{toye2023vital}
Pieter-Jan Toye.
\newblock Vital videos: A dataset of videos with ppg and blood pressure ground truths.
\newblock \emph{arXiv preprint arXiv:2306.11891}, 2023.

\bibitem[Trammel \& Sapra(2020)Trammel and Sapra]{trammel2020physiology}
Jacob~E Trammel and Amit Sapra.
\newblock Physiology, systemic vascular resistance.
\newblock 2020.

\bibitem[Wang et~al.(2009)Wang, Pickwell-MacPherson, Liang, and Zhang]{wang2009noninvasive}
L~Wang, Emma Pickwell-MacPherson, YP~Liang, and Yuan~Ting Zhang.
\newblock Noninvasive cardiac output estimation using a novel photoplethysmogram index.
\newblock In \emph{2009 annual international conference of the IEEE engineering in medicine and biology society}, pp.\  1746--1749. IEEE, 2009.

\bibitem[Weng et~al.(2024)Weng, Baur, Daswani, Chen, Harrell, Kakarmath, Jabara, Behsaz, McLean, Matias, et~al.]{weng2024predicting}
Wei-Hung Weng, Sebastien Baur, Mayank Daswani, Christina Chen, Lauren Harrell, Sujay Kakarmath, Mariam Jabara, Babak Behsaz, Cory~Y McLean, Yossi Matias, et~al.
\newblock Predicting cardiovascular disease risk using photoplethysmography and deep learning.
\newblock \emph{PLOS Global Public Health}, 4\penalty0 (6):\penalty0 e0003204, 2024.

\bibitem[Wu et~al.(2020)Wu, Wan, Gu, Mou, Dong, Luo, Zhang, and Hua]{wu2020gestational}
Yuelin Wu, Sheng Wan, Shengyi Gu, Zhengqian Mou, Lingling Dong, Zhongcheng Luo, Jun Zhang, and Xiaolin Hua.
\newblock Gestational weight gain and adverse pregnancy outcomes: a prospective cohort study.
\newblock \emph{BMJ open}, 10\penalty0 (9):\penalty0 e038187, 2020.

\bibitem[Y{\`e}che et~al.(2021)Y{\`e}che, Dresdner, Locatello, H{\"u}ser, and R{\"a}tsch]{yeche2021neighborhood}
Hugo Y{\`e}che, Gideon Dresdner, Francesco Locatello, Matthias H{\"u}ser, and Gunnar R{\"a}tsch.
\newblock Neighborhood contrastive learning applied to online patient monitoring.
\newblock In \emph{International Conference on Machine Learning}, pp.\  11964--11974. PMLR, 2021.

\bibitem[Yuan et~al.(2024{\natexlab{a}})Yuan, Chan, Creagh, Tong, Acquah, Clifton, and Doherty]{yuan2024self}
Hang Yuan, Shing Chan, Andrew~P Creagh, Catherine Tong, Aidan Acquah, David~A Clifton, and Aiden Doherty.
\newblock Self-supervised learning for human activity recognition using 700,000 person-days of wearable data.
\newblock \emph{NPJ digital medicine}, 7\penalty0 (1):\penalty0 91, 2024{\natexlab{a}}.

\bibitem[Yuan et~al.(2024{\natexlab{b}})Yuan, Zhang, Chen, Gu, and Yang]{yuan2024brant}
Zhizhang Yuan, Daoze Zhang, Junru Chen, Geifei Gu, and Yang Yang.
\newblock Brant-2: Foundation model for brain signals.
\newblock \emph{arXiv preprint arXiv:2402.10251}, 2024{\natexlab{b}}.

\bibitem[Yun et~al.(2024)Yun, Cosentino, Behsaz, McCaw, Hill, Luben, Lai, Bates, Yang, Schwantes-An, et~al.]{yun2024unsupervised}
Taedong Yun, Justin Cosentino, Babak Behsaz, Zachary~R McCaw, Davin Hill, Robert Luben, Dongbing Lai, John Bates, Howard Yang, Tae-Hwi Schwantes-An, et~al.
\newblock Unsupervised representation learning on high-dimensional clinical data improves genomic discovery and prediction.
\newblock \emph{Nature Genetics}, pp.\  1--10, 2024.

\bibitem[Zhang et~al.(2018)Zhang, Cui, Mueller, Tao, Kim, Rueschman, Mariani, Mobley, and Redline]{zhang2018national}
Guo-Qiang Zhang, Licong Cui, Remo Mueller, Shiqiang Tao, Matthew Kim, Michael Rueschman, Sara Mariani, Daniel Mobley, and Susan Redline.
\newblock The national sleep research resource: towards a sleep data commons.
\newblock \emph{Journal of the American Medical Informatics Association}, 25\penalty0 (10):\penalty0 1351--1358, 2018.

\bibitem[Zhang et~al.(2022)Zhang, Zhao, Tsiligkaridis, and Zitnik]{zhang2022self}
Xiang Zhang, Ziyuan Zhao, Theodoros Tsiligkaridis, and Marinka Zitnik.
\newblock Self-supervised contrastive pre-training for time series via time-frequency consistency.
\newblock \emph{Advances in Neural Information Processing Systems}, 35:\penalty0 3988--4003, 2022.

\bibitem[Zhang et~al.(2019)Zhang, Lyu, Qu, Qiu, Luo, Zhang, Fan, and Shi]{zhang2019photoplethysmogram}
Xiao Zhang, Yongqiang Lyu, Tong Qu, Pengfei Qiu, Xiaomin Luo, Jingyu Zhang, Shunjie Fan, and Yuanchun Shi.
\newblock Photoplethysmogram-based cognitive load assessment using multi-feature fusion model.
\newblock \emph{ACM Transactions on Applied Perception (TAP)}, 16\penalty0 (4):\penalty0 1--17, 2019.

\bibitem[Zhou et~al.(2017)Zhou, Arshad, Luo, Yu, Berkovsky, and Chen]{zhou2017indexing}
Jianlong Zhou, Syed~Z Arshad, Simon Luo, Kun Yu, Shlomo Berkovsky, and Fang Chen.
\newblock Indexing cognitive load using blood volume pulse features.
\newblock In \emph{Proceedings of the 2017 CHI Conference Extended Abstracts on Human Factors in Computing Systems}, pp.\  2269--2275, 2017.

\bibitem[Zhou et~al.(2024)Zhou, Cosentino, Yun, Biradar, Shreibati, Lai, Schwantes-An, Luben, McCaw, Engmann, et~al.]{zhou2024utilizing}
Yuchen Zhou, Justin Cosentino, Taedong Yun, Mahantesh~I Biradar, Jacqueline Shreibati, Dongbing Lai, Tae-Hwi Schwantes-An, Robert Luben, Zachary McCaw, Jorgen Engmann, et~al.
\newblock Utilizing multimodal ai to improve genetic analyses of cardiovascular traits.
\newblock \emph{medRxiv}, 2024.

\end{thebibliography}
\bibliographystyle{iclr2025_conference}

\newpage
\appendix
\section*{APPENDIX}
\section{Training and Inference Details} \label{appendix:hyperparameters}
\noindent\textbf{Architecture \& Pre-training.} The architecture of our ResNet 18-block encoder is described in Tables \ref{tab:resnet}, \ref{tab:resnet_block1}, and \ref{tab:resnet_block2}. Each 1D convolution layer is configured with a kernel size of 3 and a stride of 2, while the max-pooling layer utilizes a kernel size of 3 with a stride of 1. We start with a filter size of 32, which doubles every 4 blocks to capture progressively more complex features. Dropout is applied with a probability of 0.5 to prevent overfitting. This backbone architecture is used across different methods in our experiments to ensure consistency and make for a fair comparison during evaluation. Additionally, our SSL baselines use the same batch size, learning rate, input sampling frequency, and training steps as \model{}. We use the same augmentation types and intensity for BYOL \citep{grill2020bootstrap}, SimCLR \cite{chen2020simple}, and \model{}-P. Furthermore, we investigated 0.07 and 0.5 temperatures as MoCo \citep{he2020momentum} and SimCLR \citep{chen2020simple}, respectively. In contrast to a smaller embedding size of 256 adopted by \citep{abbaspourazad2023large}, we project the learned representations to a 512-dimensional embedding after the convolutional block (we investigated larger embedding sizes of 768 and 1024 and found no significant performance changes). This embedding is then passed through two Mixture of Experts (MoE) blocks, each containing three experts. Each expert block consists of two sequential linear layers, with sizes 256 and 1, which are used for IPA and SQI prediction tasks. It is noteworthy that both BYOL and TF-C require multiple encoders and different projection heads, resulting in variations in model sizes. For these methods, we use existing implementations available online\footnote{\url{https://github.com/chengding0713/SiamQuality}}\footnote{\url{https://github.com/mims-harvard/TFC-pretraining}}, but apply our encoder as the backbone to ensure consistency. Our models are pre-trained for 15,000 steps using the Adam optimizer, with a learning rate of $10^{-4}$. We use a batch size of 128 for training since after various trials we did not observe significant differences in performance with batch sizes of 64 and 256. We performed five iterations of pre-training and selected the best-performing model for each downstream task. For SimCLR and \model{}-P, a single model consistently achieves the best performance across all tasks. For BYOL, we select two models that perform best across all tasks. Similarly, for TF-C and \model{}-S, we choose three models with the highest performance. We use this approach as some models excel in certain task groups while others perform better in the rest. Note that a more robust approach would involve broader hyperparameter tuning with k-fold validation to obtain the optimal model. However, this requires substantial computational resources for pre-training. Additionally, we did not perform an exhaustive evaluation of different augmentation settings but instead used transformations and values based on prior research \citep{abbaspourazad2023large, tang2020exploring}. For model training, we primarily used PyTorch \citep{paszke2019pytorch}. The NTXentLoss implementation was sourced from the PyTorch Metric Learning package\footnote{\url{https://github.com/KevinMusgrave/pytorch-metric-learning}}.

\begin{table}[h]
\centering
\begin{minipage}{0.48\linewidth}
    \centering
    \caption{ResNet-style CNN encoder architecture used in \model{}.}
    \begin{tabular}{@{}ll@{}}
    \toprule
    \textbf{Layer} & \textbf{Output Shape} \\ \midrule
    Conv1     & [32, 32, 1250]                         \\
    Batch Norm &[32, 32, 1250]  \\
    ReLU &[32, 32, 1250] \\
    Basic Block Type 1 & [32, 32, 1250]          \\
    (Basic Block Type 2) $\times$ 3 & [32, 32, 313]       \\ 
    (Basic Block Type 2) $\times$ 4 & [32, 64, 79]\\
    (Basic Block Type 2) $\times$ 4 & [32, 128, 20]  \\
    (Basic Block Type 2) $\times$ 4 & [32, 256, 5] \\
    (Basic Block Type 2) $\times$ 2 & [32, 512, 3] \\
    BatchNorm & [32, 512, 3] \\
    ReLU & [32, 512, 3]\\
    Linear & [32, 512] \\
    \bottomrule
    \end{tabular}
    \label{tab:resnet}
\end{minipage}
\hfill
\begin{minipage}{0.22\linewidth}
    \centering
    \caption{Basic Block Type 1}
    \begin{tabular}{@{}l@{}}
    \toprule
    \textbf{Layer} \\ \midrule
    Conv1D \\
    BatchNorm \\
    ReLU \\
    Dropout \\
    Conv1D \\
    \bottomrule
    \end{tabular}
    \label{tab:resnet_block1}
\end{minipage}
\hfill
\begin{minipage}{0.22\linewidth}
    \centering
    \caption{Basic Block Type 2}
    \begin{tabular}{@{}l@{}}
    \toprule
    \textbf{Layer} \\ \midrule
    BatchNorm \\
    ReLU \\
    Dropout \\
    Conv1D \\
    BatchNorm \\
    ReLU \\
    Dropout \\
    Conv1D \\
    Maxpool \\
    \bottomrule
    \end{tabular}
    \label{tab:resnet_block2}
\end{minipage}
\end{table}

\noindent\textbf{Parameters: Training and Inference}. This section outlines the training and inference parameters used in our methods. Inference parameters are those utilized for feature extraction. 
\begin{itemize}
    \item \model{}-P (5M) and SimCLR (5M): Both training and inference involve 5M parameters. For SimCLR, it is worth noting that we use the projection features during inference instead of using the encoder only.
    \item BYOL (5M): During training, the online and target encoders each have 5M parameters, and the projector is 800K. At inference, only the online encoder is used for feature extraction, totaling 5M parameters.
    \item TF-C (10M): The time and frequency encoders each have 5M parameters, followed by a smaller projector ($<$ 100K). Since both encoders and projectors are required for inference, the total parameter count is 10M.
    \item \model{}-S (5M): The encoder consists of 5M parameters, while the expert heads contribute approximately 400K each. As the expert heads are not used for feature extraction, the inference parameter total remains 5M.
\end{itemize}


\noindent\textbf{Feature Extraction \& Linear Evaluation}
We extracted the projected embedding for linear evaluation. For Moment and Chronos, we extract the default embedding size, which is 1024 and 768, respectively. We use cross-validated grid search to identify the best parameters for our linear probes. The hyperparameters chosen for each model are as follows: (1) Logistic Regression: \{'penalty': ['l1', 'l2'], 'C': [0.01, 0.1, 1, 10, 100], 'solver': ['lbfgs'], 'max\_iter': [100, 200]\}. (2) Linear Regression: \{'alpha': [0.1, 1.0, 10.0, 100.0], 'solver': ['auto', 'cholesky', 'sparse\_cg']\}. (3) Random Forest: \{'n\_estimators': [100, 200], 'max\_features': ['sqrt', 'log2'], 'max\_depth': [10, 20, 30], 'min\_samples\_split': [2, 5], 'min\_samples\_leaf': [1, 2]\}

\section{Datasets and Tasks}
\label{appendix:datasets_tasks}
\begin{table}[h]
\centering
\caption{The task evaluation benchmark of \model{}. Datasets highlighted in gray are unseen during training, thus, the corresponding tasks are out-of-domain. The rest were used for pre-training but their test sets and labels are held out. For task type, B/M/R refer to Binary classification, Multi-class classification (\#classes), and Regression, respectively.}
\label{tab:downstream_tasks}
\scalebox{0.65}{
\begin{tabular}{@{}lllllll@{}}
\toprule
\textbf{\#ID} &\textbf{Dataset} &\textbf{SR (Hz)} &\textbf{Collected by} & \textbf{Task} & \textbf{Task Type} &\textbf{\#Participants (\#Samples)} \\ \midrule
T1 & VitalDB \citep{lee2022vitaldb} &500 &ICU monitor &ICU admission (Yes/No) &B &5866 \\ 
T2 & & & &Operation Type &M (11) &5866 \\
T3 &MIMIC-III \citep{moody2020mimic} &125 &ICU Monitor &Mortality &B &5596 \\
T4 &MESA \citep{zhang2018national} &256 &Polysomnography finger &Smoker &B &2055 \\
T5 & & & &AHI $>$ 3\% Oxygen Desat. &R &2055 \\
T6 & & & &AHI $>$ 4\% Oxygen Desat. &R &2055 \\
T7 &\cellcolor{gray!10}nuMom2B \citep{facco2015numom2b} &75 &Polysomnography finger &Pregnancy stage (early/late) &B &3163 (5337) \\ 
T8 &\cellcolor{gray!10} & & &Gestation Age &R &3163 (5337) \\
T9 &\cellcolor{gray!10}VV (Skin Tone) \citep{toye2023vital} &60 &Finger &Systolic BP &R &231 \\ 
T10 &\cellcolor{gray!10} & & &Diastolic BP &R & 231 \\

T11 &\cellcolor{gray!10}PPG-BP \citep{liang2018new}& 1000 &Finger Pulse Ox & Systolic BP &R &219\\
T12 &\cellcolor{gray!10} & & &Diastolic BP &R &219 \\
T13 &\cellcolor{gray!10}& & &Average Heart Rate &R &219 \\
T14 &\cellcolor{gray!10} & & & Hypertension &B &219 \\
T15 &\cellcolor{gray!10}SDB \citep{garde2014development} &62.5 &Finger Pulse Ox &Sleep Disordered Breathing &B &146 \\
T16 &\cellcolor{gray!10}ECSMP \citep{gao2021ecsmp} &64 & Wrist &Mood Disturbance &B &89 \\
T17 &\cellcolor{gray!10}WESAD \citep{schmidt2018introducing} &64 &Wrist &Valence &B &15 (4497) \\
T18 &\cellcolor{gray!10} & & &Arousal &B &15 (4497) \\
T19 &\cellcolor{gray!10}PPG-DaLiA \citep{reiss2019deep} &64 &Wrist &Heart Rate &R &15 (64697) \\
T20 &\cellcolor{gray!10} & & &Activity  &M (9) & 15 (64697) \\
\bottomrule
\end{tabular}
}
\end{table}

\begin{figure}[h]
    \centering
    \begin{subfigure}[b]{0.45\textwidth}
         \includegraphics[width=\textwidth]{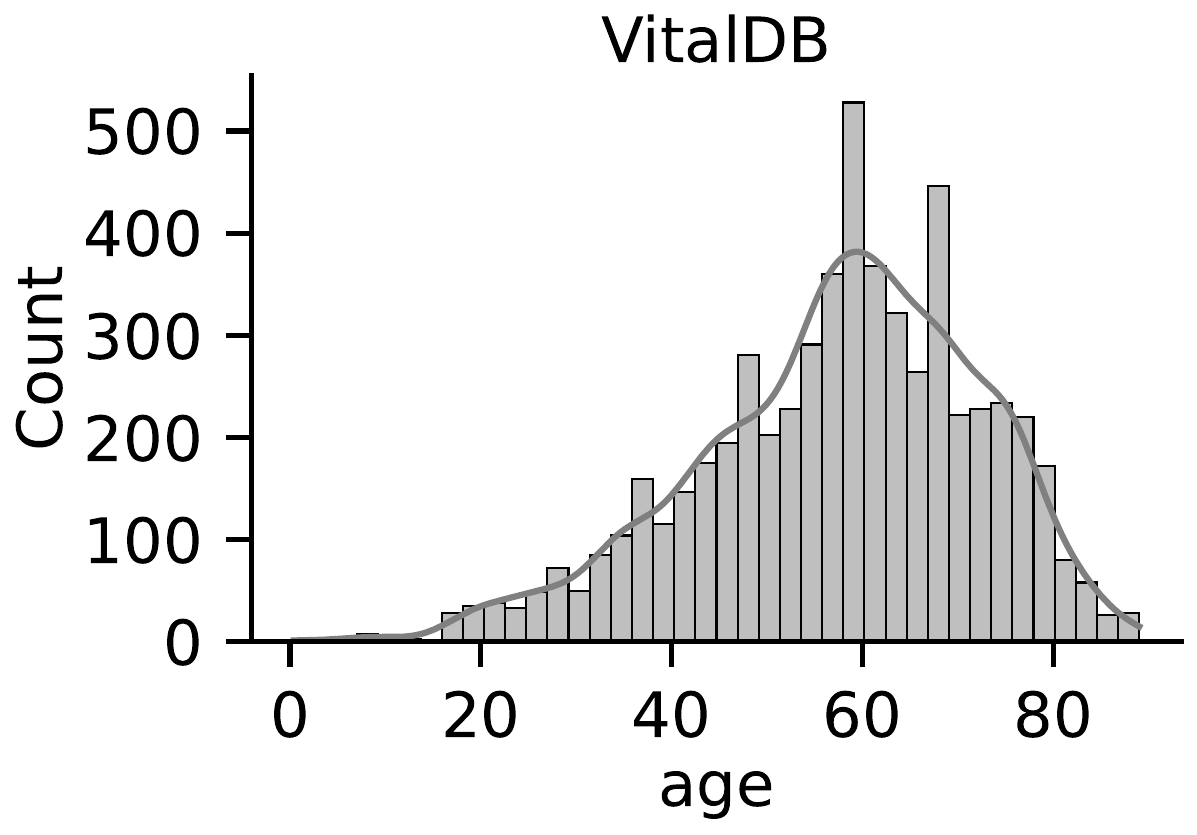}
     \end{subfigure}
     \hfill
     \begin{subfigure}[b]{0.25\textwidth}
         \includegraphics[width=\textwidth]{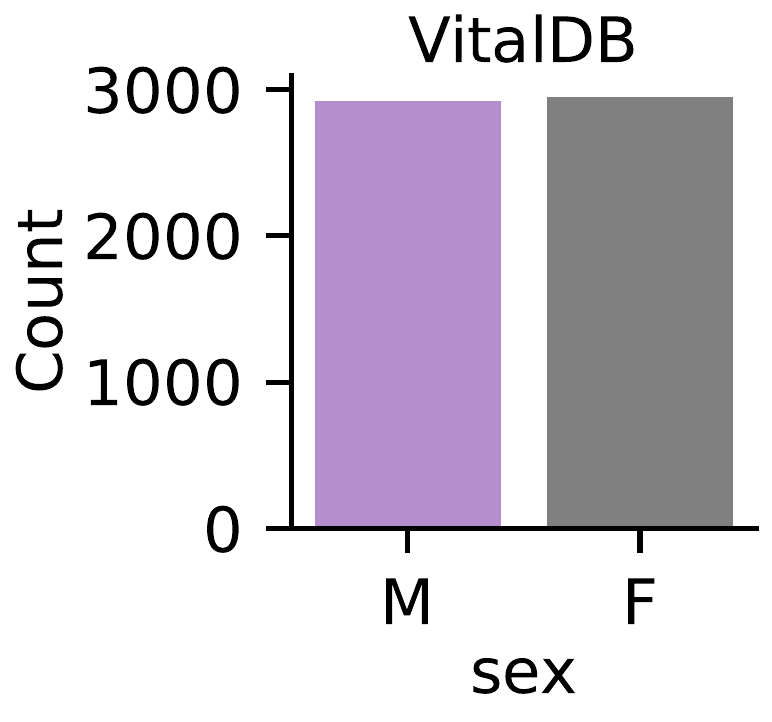}
     \end{subfigure}
     \hfill
    \begin{subfigure}[b]{0.25\textwidth}
         \includegraphics[width=\textwidth]{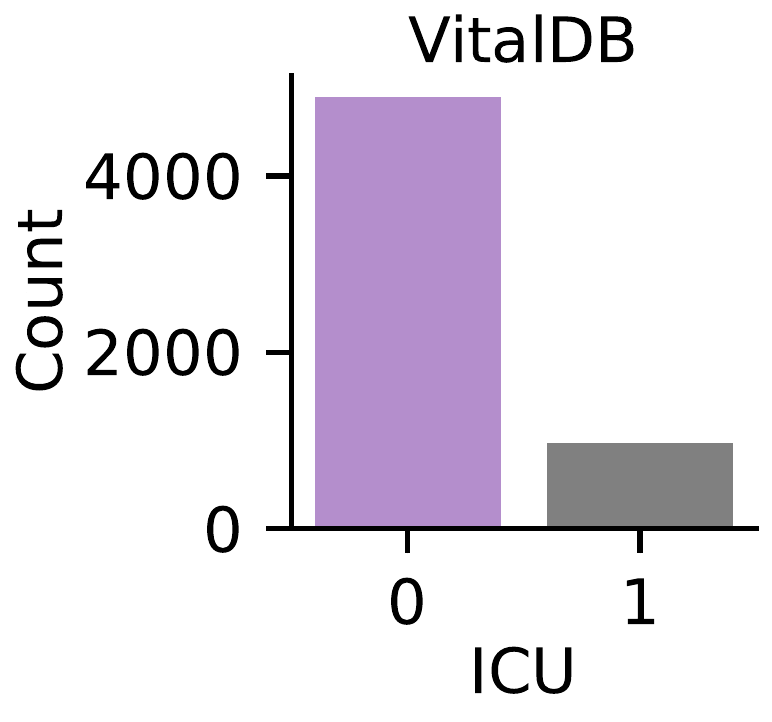}
     \end{subfigure}
     \hfill
    \caption{VitalDB dataset descriptive statistics.}
    \label{fig:vital_dataset}
\end{figure}

\noindent\textbf{VitalDB.} The VitalDB dataset provides comprehensive monitoring of vital signs and physiological parameters from 6,388 surgical cases. This high-resolution dataset includes a wide range of intraoperative monitoring variables such as heart rate, blood pressure, oxygen saturation, and other critical physiological signals, collected at frequent intervals throughout surgery. The surgical operation belongs to one of the eleven categories: colorectal, biliary/pancreas, stomach, major resection, minor resection, breast, transplantation, thyroid, hepatic, vascular, and others. After the data cleaning process, we narrowed the dataset down to 5,866 participants with complete and usable information. As depicted in Figure \ref{fig:vital_dataset}, we observe that the gender distribution is relatively balanced, with nearly equal representation of male and female patients. Additionally, the majority of the participants fall within the age range of 50 to 70, with a significant proportion being around 60 years old. The ICU label corresponds to whether the person was admitted to the ICU or not. 

\begin{figure}
    \centering
    \begin{subfigure}[b]{0.45\textwidth}
         \includegraphics[width=\textwidth]{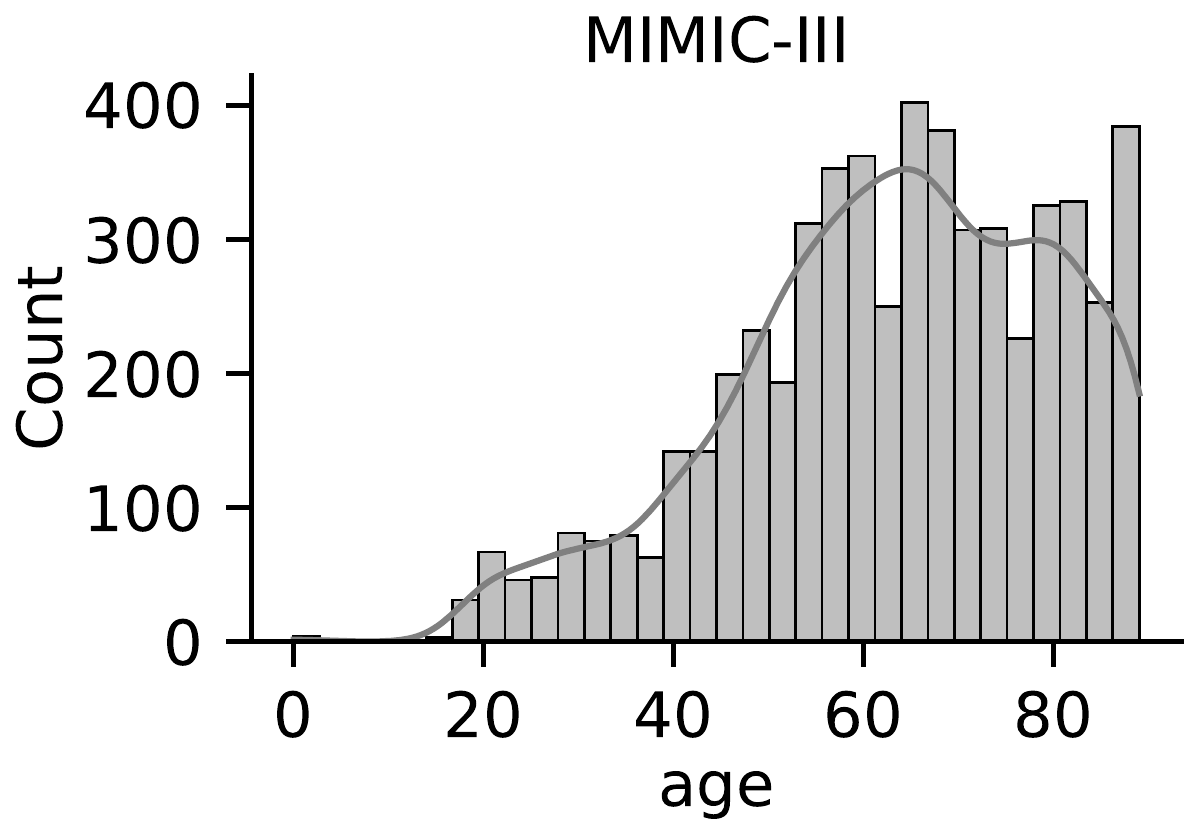}
     \end{subfigure}
     \hfill
     \begin{subfigure}[b]{0.45\textwidth}
         \includegraphics[width=\textwidth]{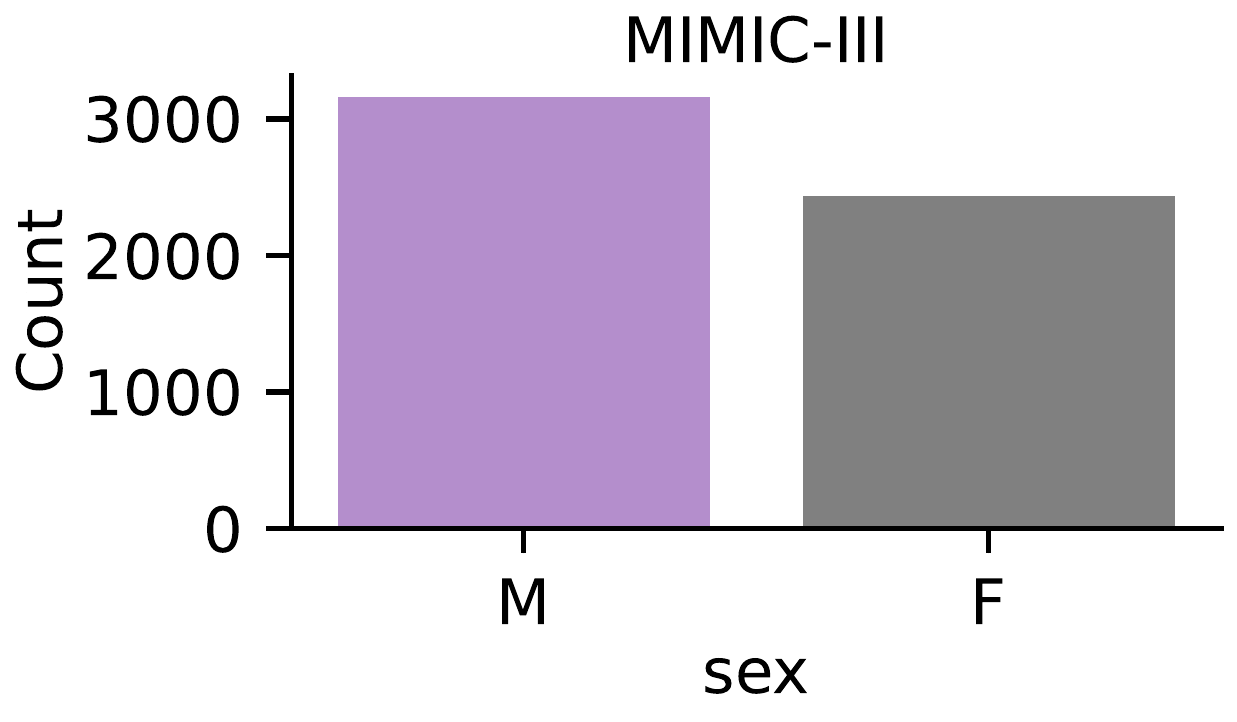}
     \end{subfigure}
     \hfill
    \caption{MIMIC-III dataset descriptive statistics.}
    \label{fig:mimic_dataset}
\end{figure}

\noindent\textbf{MIMIC-III.} In our analysis, we utilize the MIMIC-III waveform database matched subset, which comprises data from 10,282 ICU patients. From this dataset, we focus specifically on extracting photoplethysmogram (PPG) data, provided it is available for each patient. To ensure the quality of the data, we set a criterion of at least 1 minute of usable PPG signal that must be present. After performing a thorough data cleaning process, we end up with a cohort of 5,596 participants with reliable PPG data. As illustrated in Figure \ref{fig:mimic_dataset}, the dataset shows a gender imbalance, with a higher proportion of male patients compared to female patients. Additionally, the majority of participants are aged 60 years or older, reflecting a typical ICU population that often includes elderly patients with critical health conditions. 

\begin{figure}[h]
    \centering
    \begin{subfigure}[b]{0.32\textwidth}
         \includegraphics[width=\textwidth]{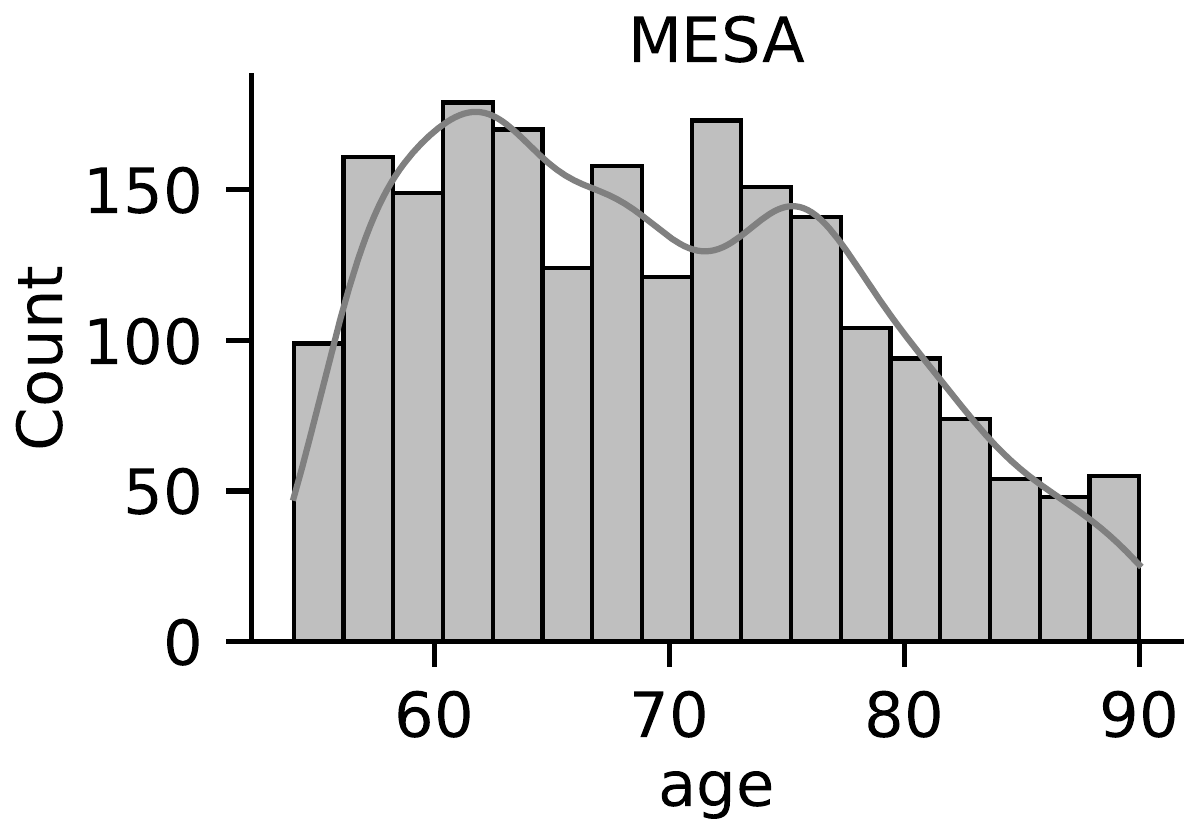}
     \end{subfigure}
     \hfill
    \begin{subfigure}[b]{0.32\textwidth}
         \includegraphics[width=\textwidth]{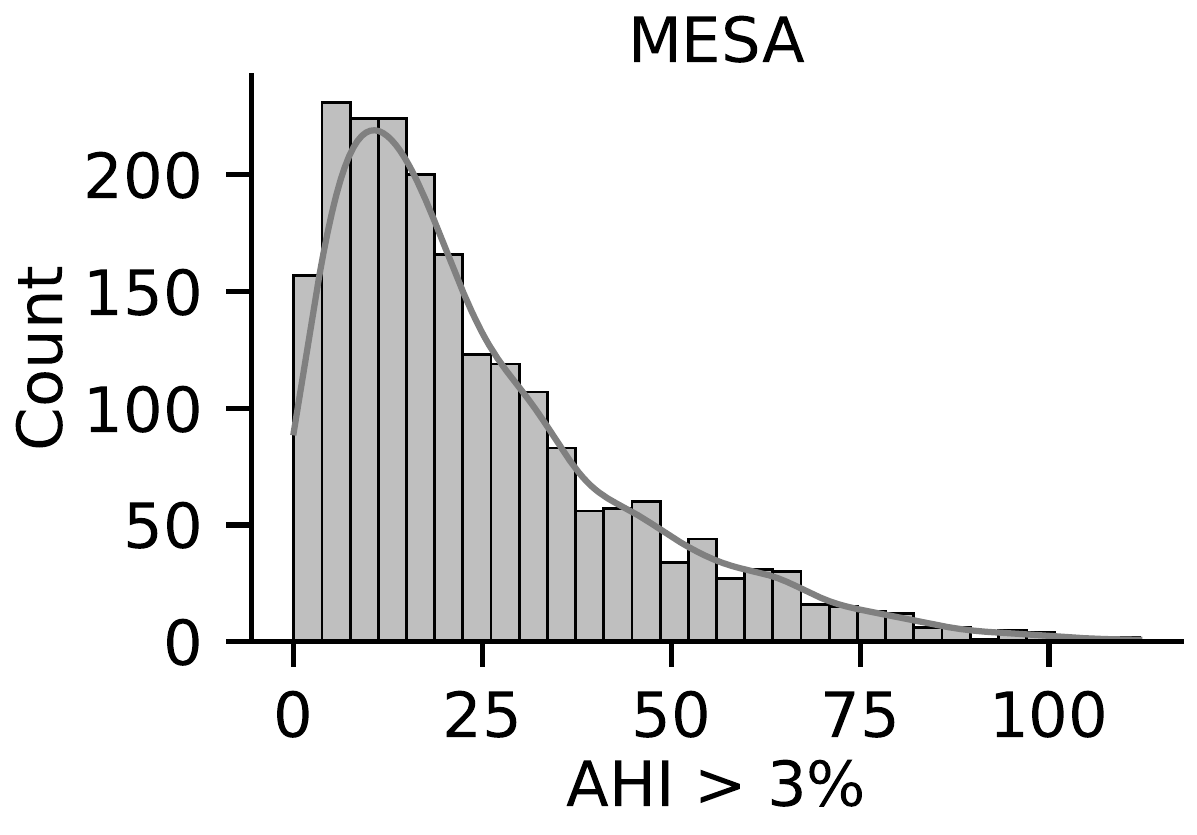}
     \end{subfigure}
     \hfill
    \begin{subfigure}[b]{0.32\textwidth}
         \includegraphics[width=\textwidth]{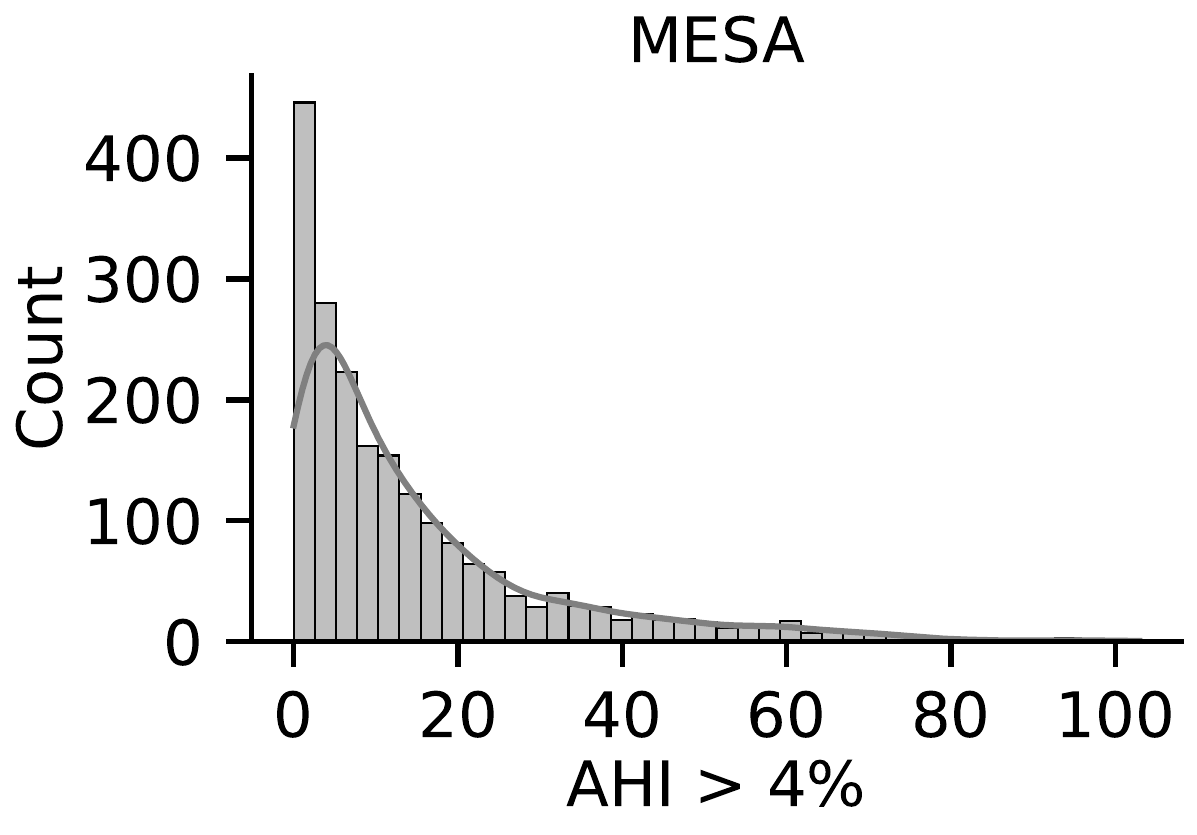}
     \end{subfigure} 
     \hfill
          \begin{subfigure}[b]{0.40\textwidth}
         \includegraphics[width=\textwidth]{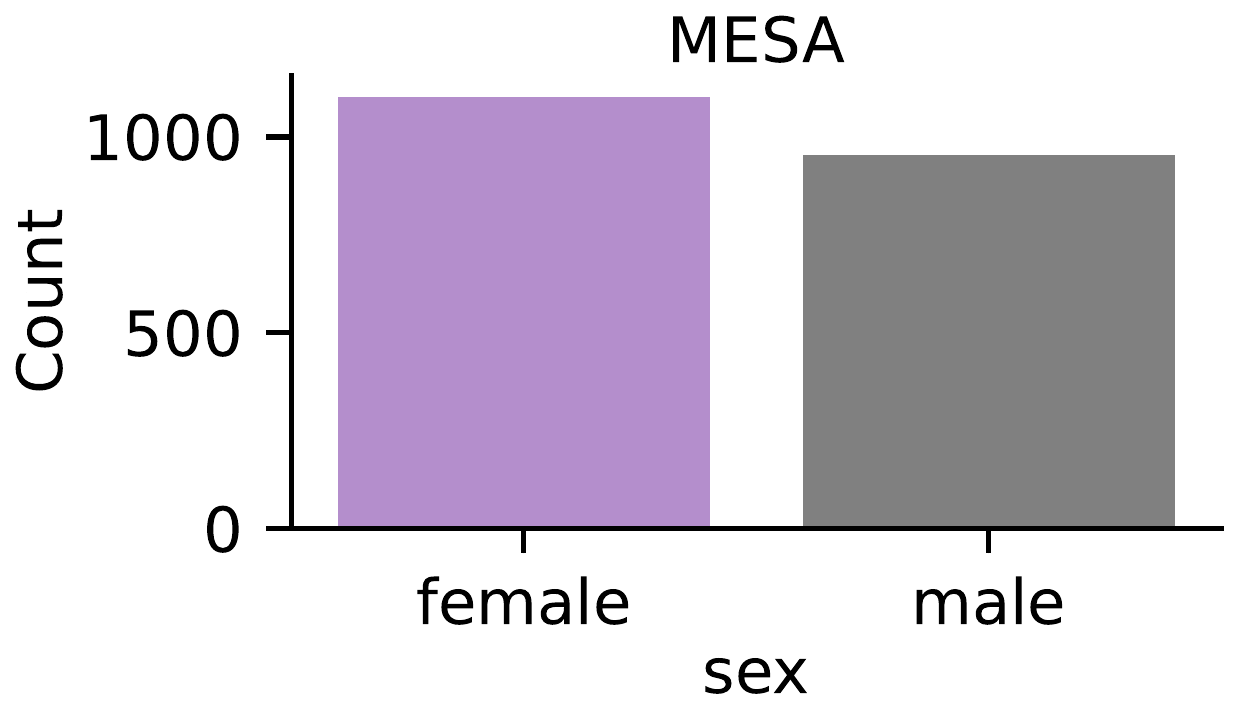}
     \end{subfigure}
     \hfill 
     \begin{subfigure}[b]{0.40\textwidth}
         \includegraphics[width=\textwidth]{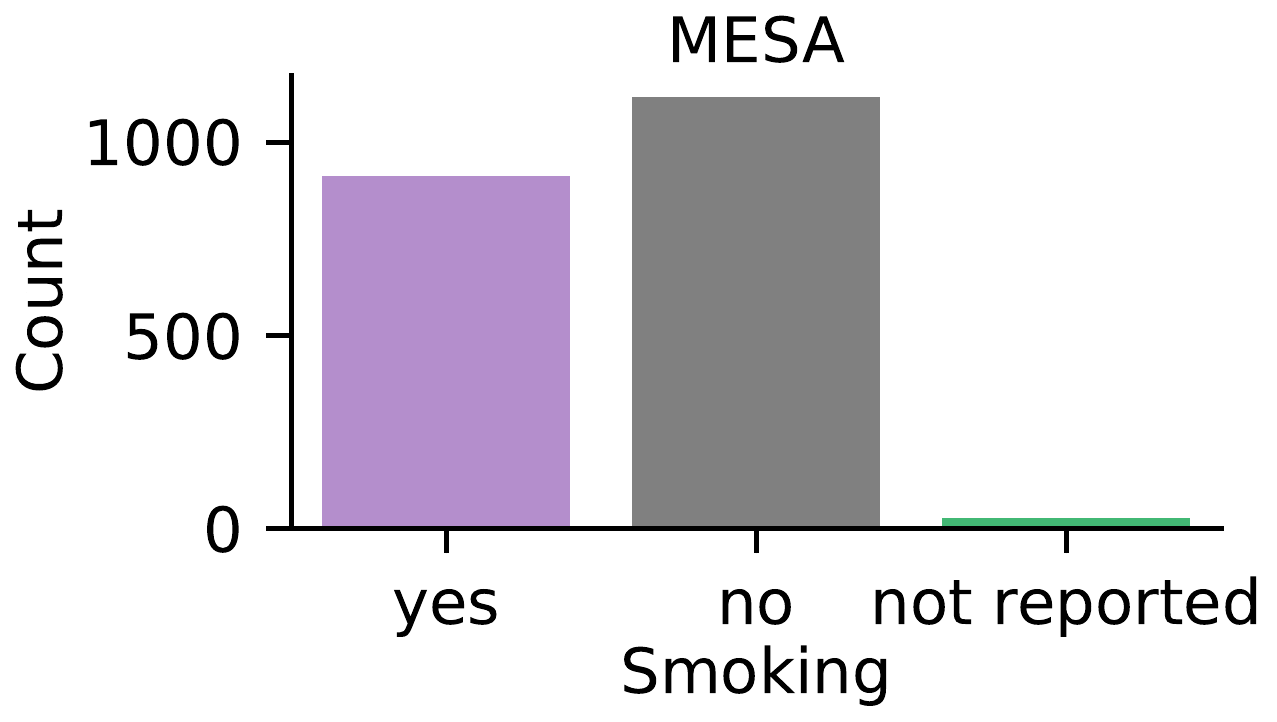}
     \end{subfigure}
     \hfill
    \caption{MESA dataset descriptive statistics.}
    \label{fig:mesa_dataset}
\end{figure}

\noindent\textbf{MESA.} 
The Multi-Ethnic Study of Atherosclerosis (MESA) sleep sub-study gathered data from 2,237 participants through overnight, unattended polysomnography to assess various sleep parameters. After the data cleaning process, we retained 2,055 participants for analysis. As shown in Figure \ref{fig:mesa_dataset}, the dataset shows a slightly larger proportion of female participants. The age distribution reveals that most participants are between 60 and 80 years old, reflecting an older adult population, which is commonly studied concerning sleep disorders and cardiovascular risks.

In this study, we use the Apnea-Hypopnea Index (AHI) with at least 3\% and 4\% oxygen desaturation as the primary measure for diagnosing sleep apnea, as recommended by the American Academy of Sleep Medicine~\citep{ruehland2009new}. These thresholds indicate the severity of sleep apnea, with oxygen desaturation during apneas/hypopneas being a critical factor. We predict these AHI values directly in our regression models. Additionally, we classify participants with any history of smoking as smokers. This approach allows us to account for both current and former smokers, capturing a broader range of smoking-related health risks within our analysis.

\begin{figure}
    \centering
    \begin{subfigure}[b]{0.35\textwidth}
         \includegraphics[width=\textwidth]{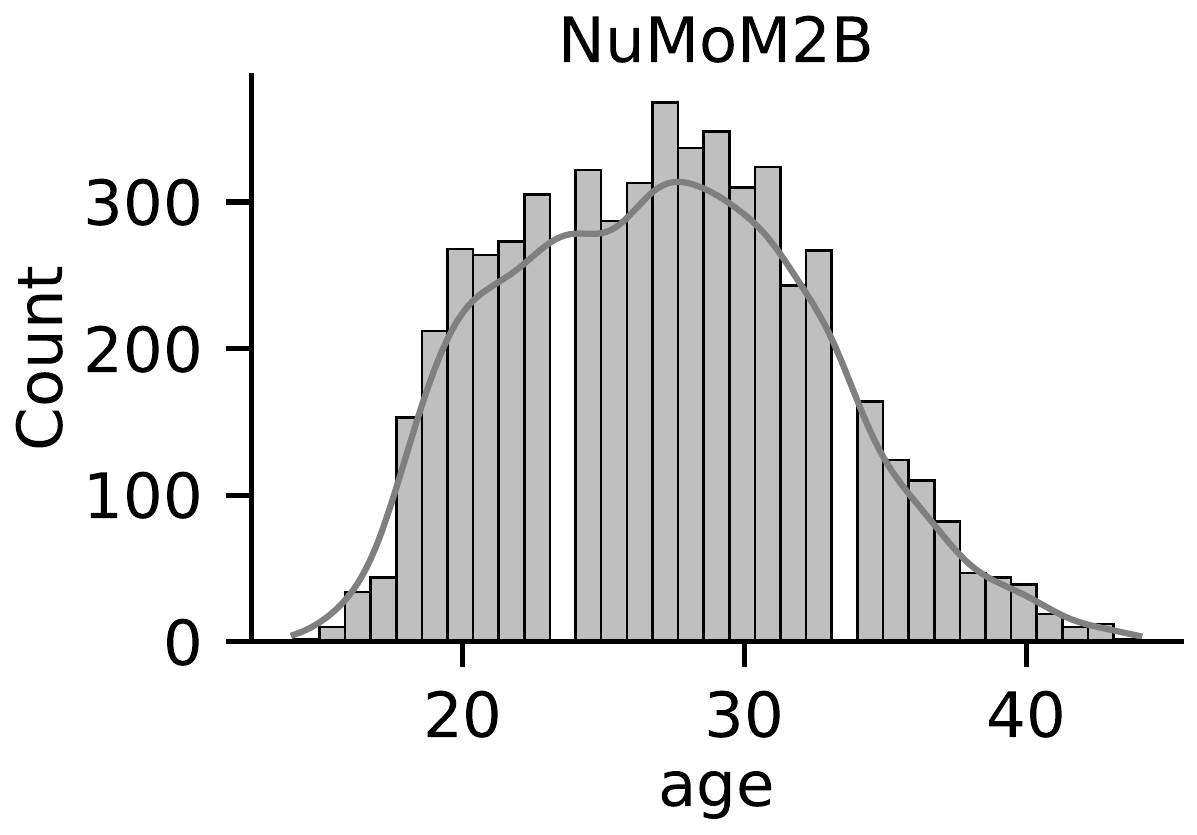}
     \end{subfigure}
     \hfill
    \begin{subfigure}[b]{0.35\textwidth}
         \includegraphics[width=\textwidth]{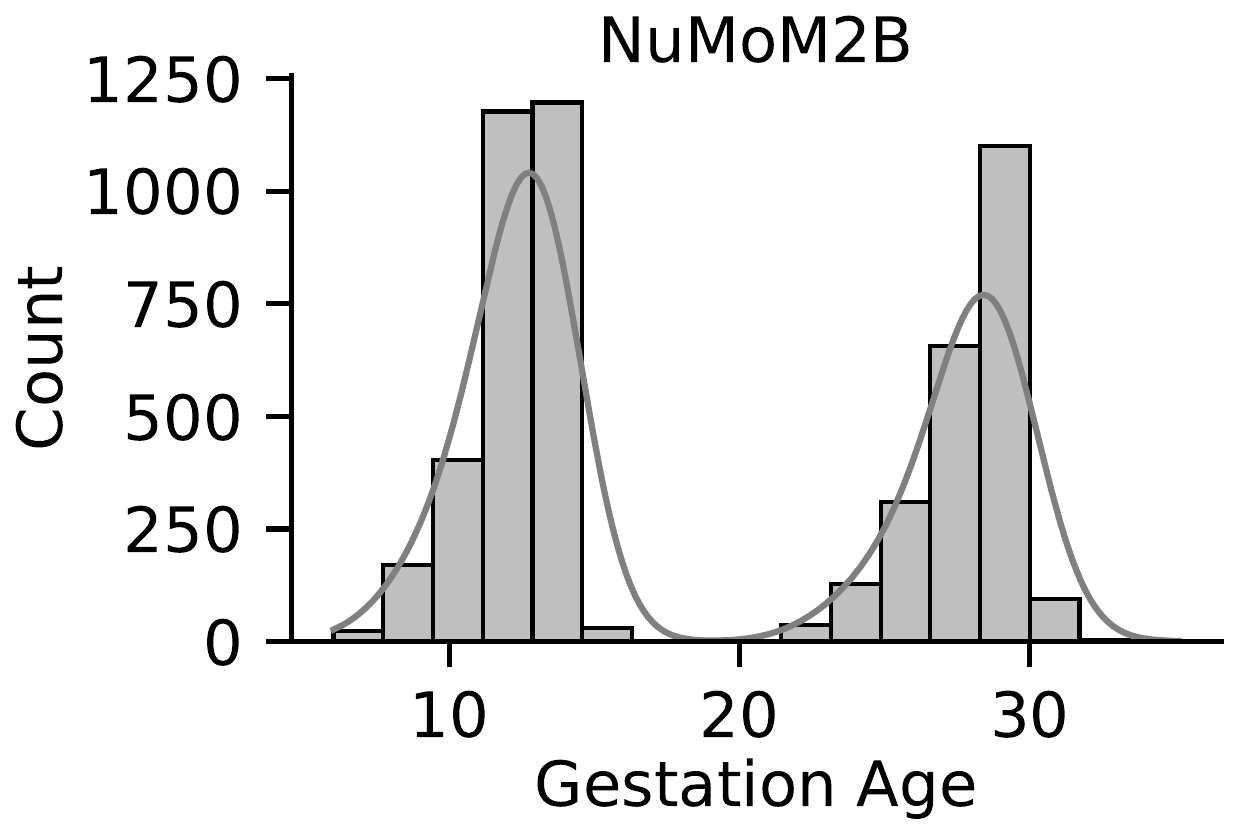}
     \end{subfigure}
     \hfill
    \begin{subfigure}[b]{0.25\textwidth}
         \includegraphics[width=\textwidth]{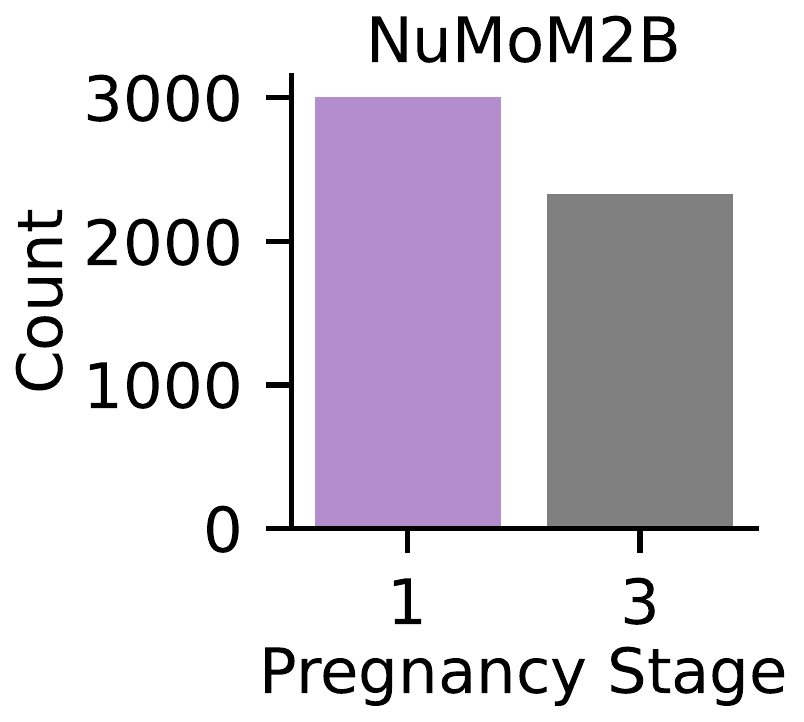}
     \end{subfigure} 
     \hfill
    \caption{NuMoM2B dataset descriptive statistics.}
    \label{fig:numom2b_dataset}
\end{figure}

\noindent\textbf{NuMoM2B.} Changes in gestational age and pregnancy stage are risk factors associated with adverse pregnancy outcomes such as hypertensive disorders and small-for-gestational-age delivery \citep{bouariu2022first, parikh2021adverse, wu2020gestational, crump2023adverse}. These diseases affect heart function that can be measured using the PPG sensor \citep{feli2024preterm}. The Nulliparous Pregnancy Outcomes Study: monitoring mothers-to-be (nuMoM2B) sub-study examines the relationship between adverse pregnancy outcomes and sleep disorders. In particular, an overnight polysomnograph that collects PPG data is administered to the women at their homes during 6-15 weeks (early) and 22-31 weeks (late) of pregnancy. Therefore, our tasks  are to classify between early and late-stage pregnancy as well as predict the gestation age of the fetus. In Figure \ref{fig:numom2b_dataset}, we observe that maternal age peaks around 28 years. The gestational age distribution is bimodal, which we use as a predictor in our regression task. For pregnancy stage, we classify visit 1 as early and visit 3 as late.

\begin{figure}[h]
    \centering
    \begin{subfigure}[b]{0.45\textwidth}
         \includegraphics[width=\textwidth]{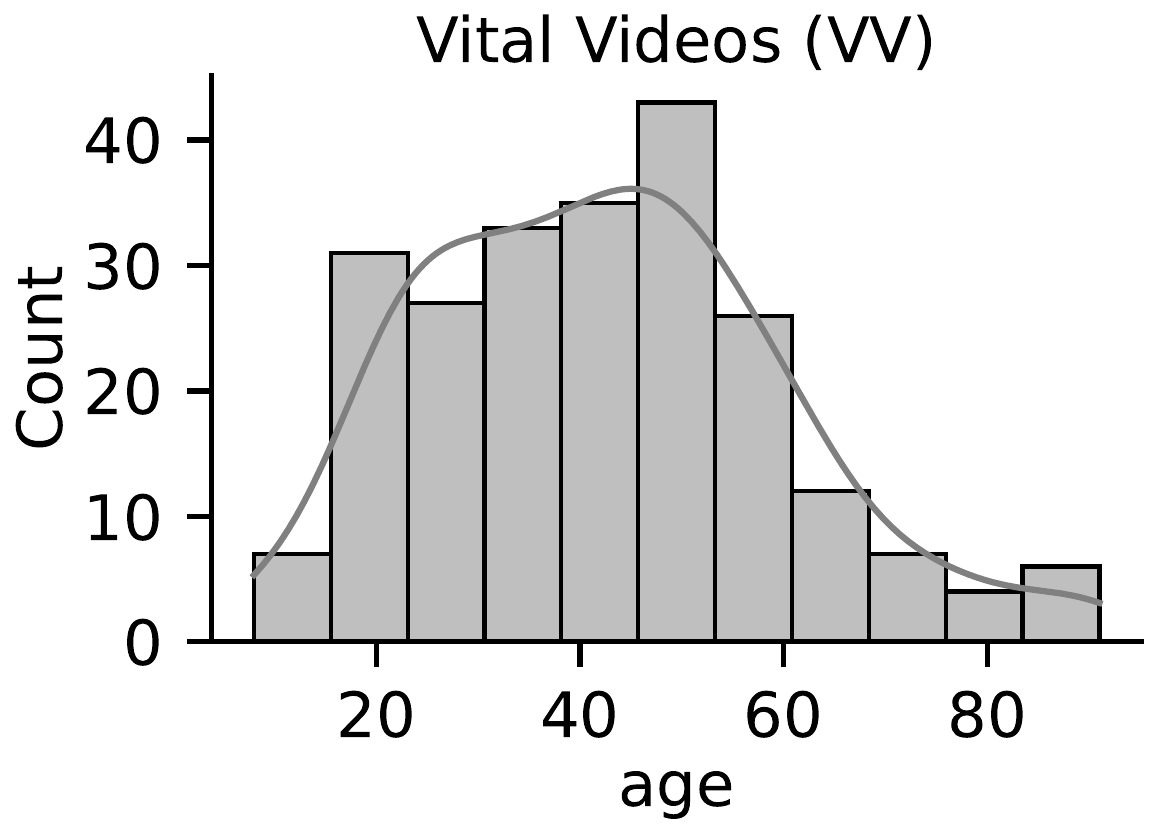}
     \end{subfigure}
     \hfill
    \begin{subfigure}[b]{0.30\textwidth}
         \includegraphics[width=\textwidth]{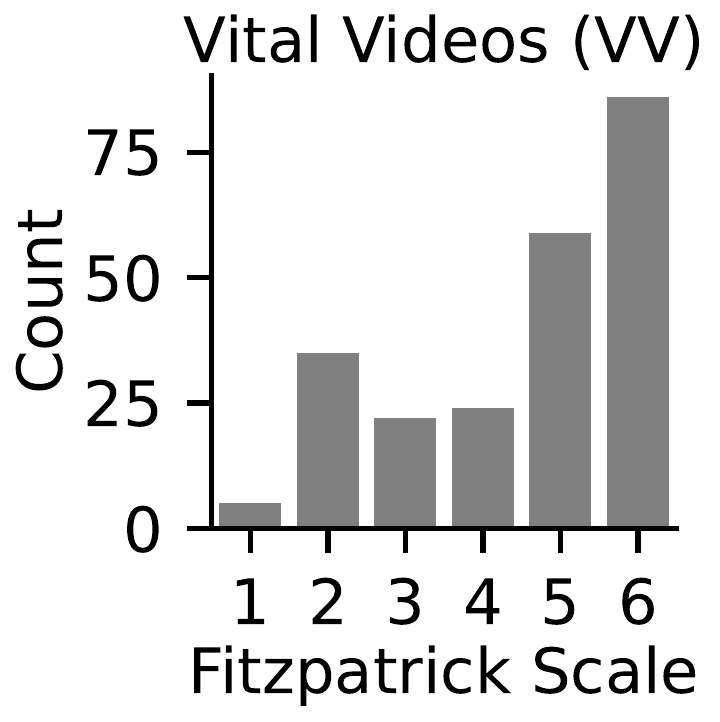}
     \end{subfigure} 
     \hfill \\
    \begin{subfigure}[b]{0.32\textwidth}
         \includegraphics[width=\textwidth]{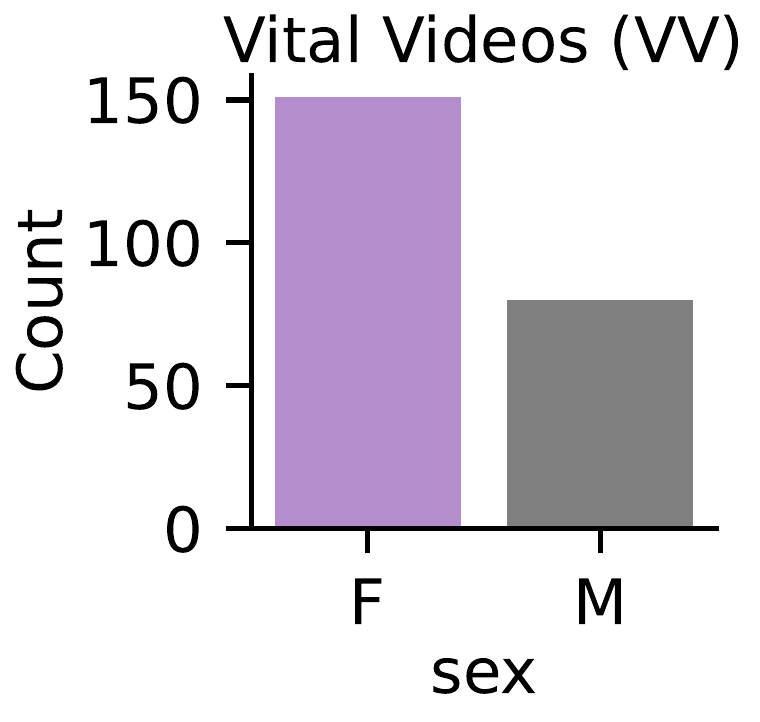}
     \end{subfigure}
     \hfill
    \begin{subfigure}[b]{0.32\textwidth}
         \includegraphics[width=\textwidth]{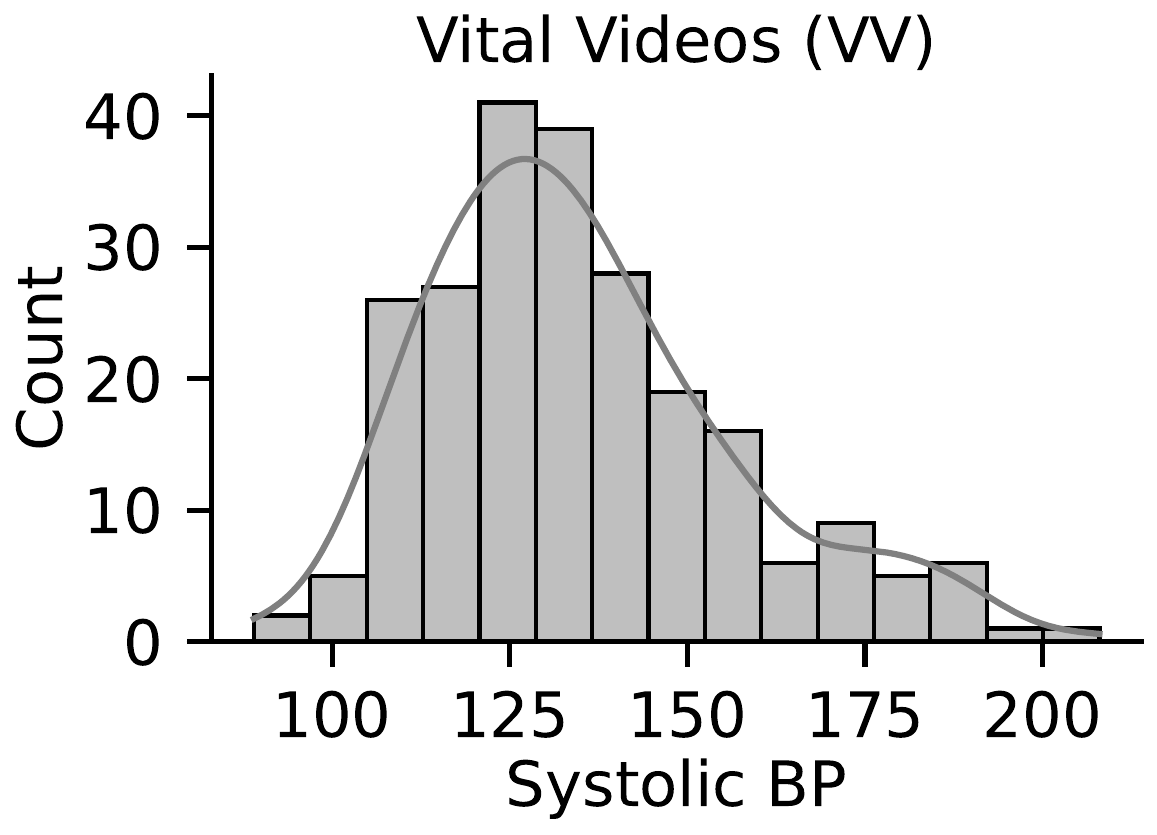}
     \end{subfigure} 
     \hfill
     \begin{subfigure}[b]{0.32\textwidth}
         \includegraphics[width=\textwidth]{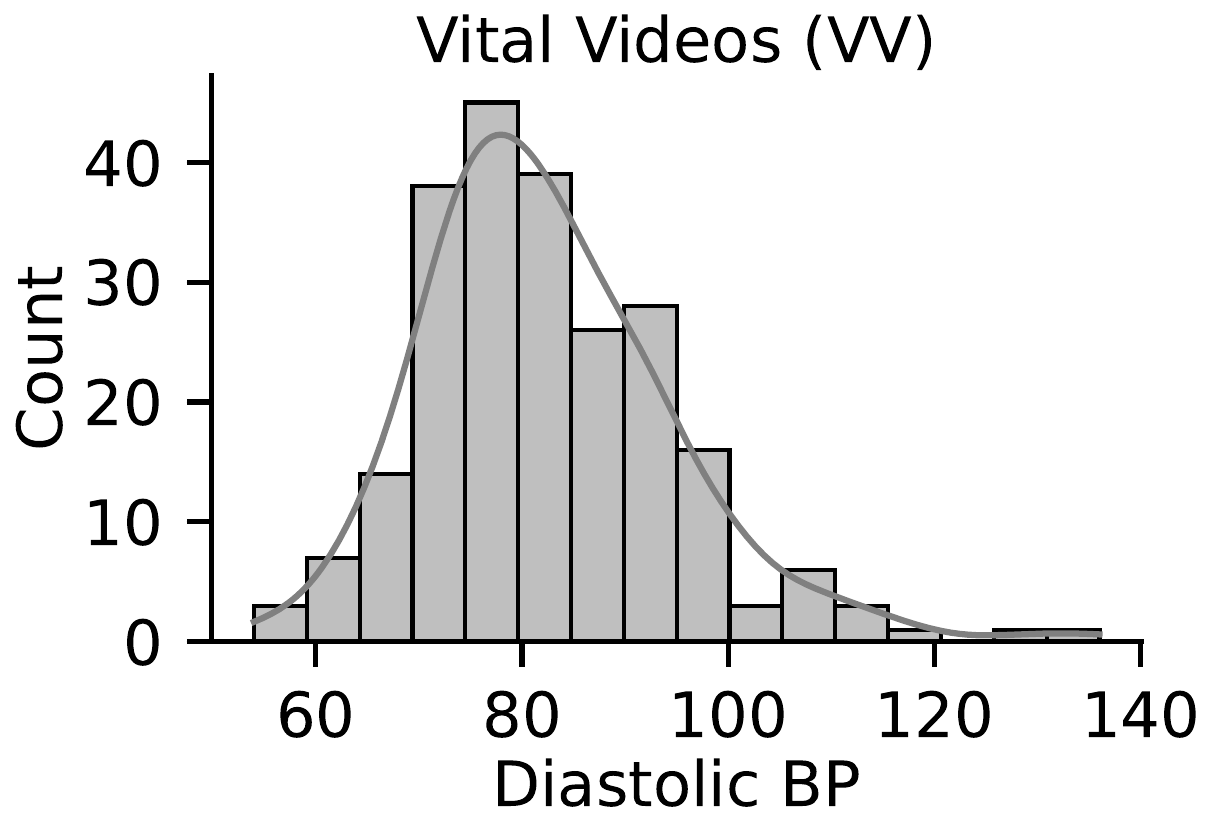}
     \end{subfigure} 
     \hfill
    \caption{Vital Videos dataset descriptive statistics.}
    \label{fig:vv_dataset}
\end{figure}

\noindent\textbf{VitalVideos (VV) (Skin Tone).} The Vital Videos study is an ongoing project that collects data on vital signs, videos, and blood pressure across a variety of conditions, including variations in lighting, background, and skin tone. For our analysis, we used data from two groups, totalling 231 participants, from Europe and Sub-Saharan Africa. As shown in Figure \ref{fig:vv_dataset}, most participants have a Fitzpatrick skin tone of 5 or 6, indicating darker skin. The dataset is primarily composed of female participants, with an age range between 40 and 60 years. Additionally, the majority of participants had a systolic blood pressure of around 125 and a diastolic pressure of around 80, suggesting that most individuals in the study were relatively healthy.

\begin{figure}
    \centering
    \begin{subfigure}[b]{0.32\textwidth}
         \includegraphics[width=\textwidth]{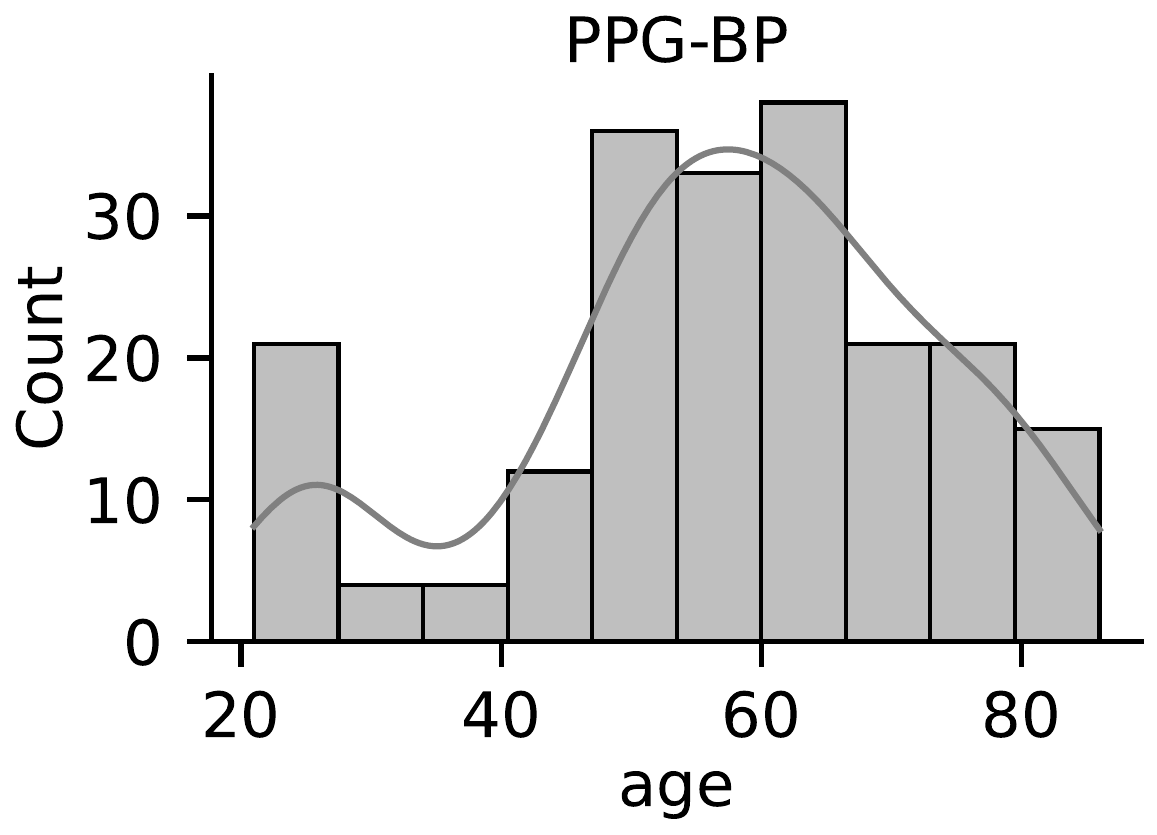}
     \end{subfigure}
     \hfill
    \begin{subfigure}[b]{0.32\textwidth}
         \includegraphics[width=\textwidth]{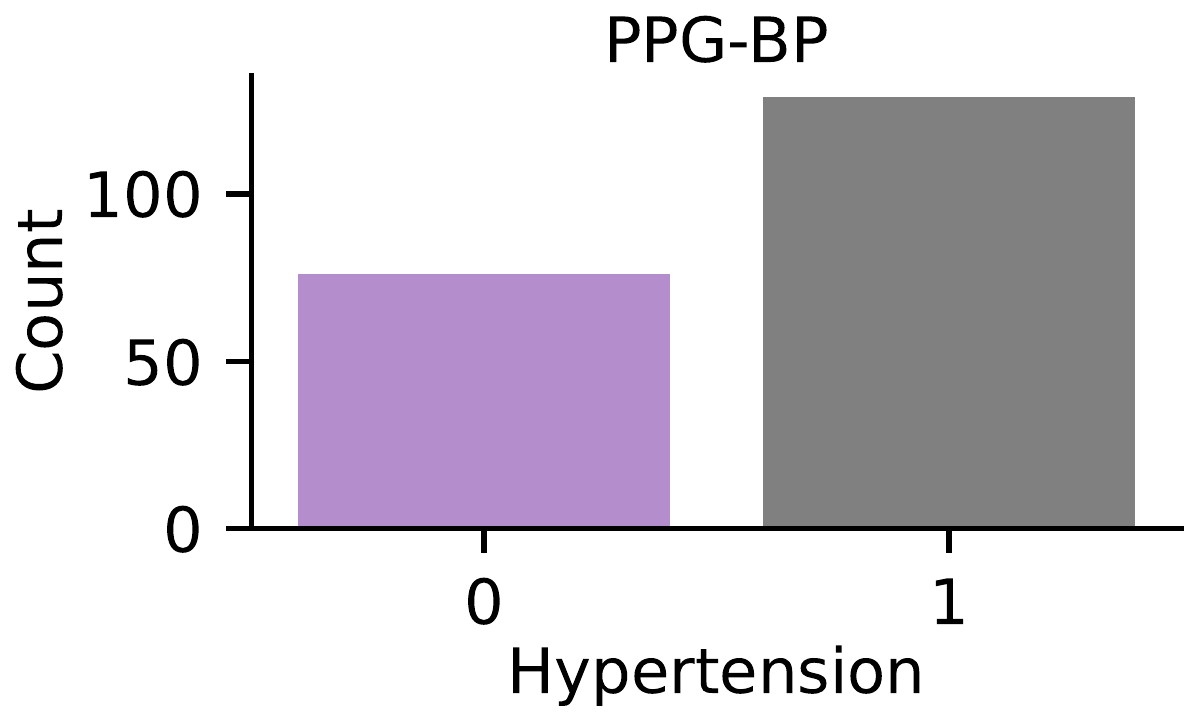}
     \end{subfigure} 
     \hfill 
    \begin{subfigure}[b]{0.32\textwidth}
         \includegraphics[width=\textwidth]{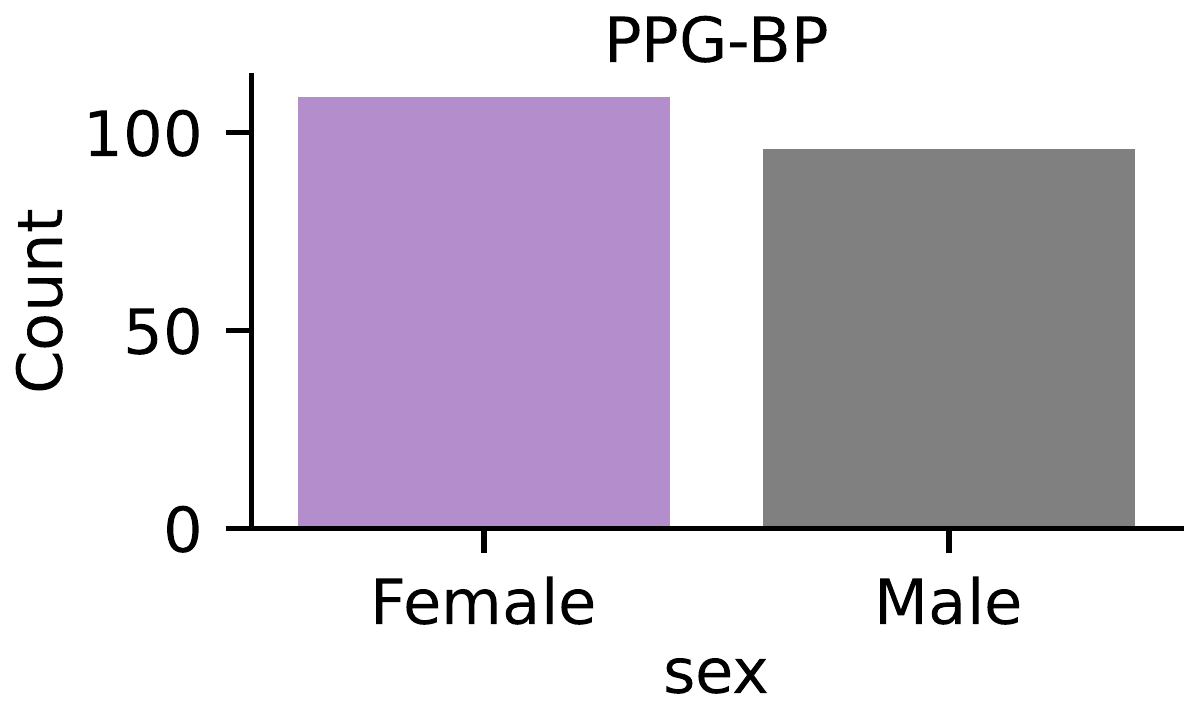}
     \end{subfigure}
     \hfill
    \begin{subfigure}[b]{0.32\textwidth}
         \includegraphics[width=\textwidth]{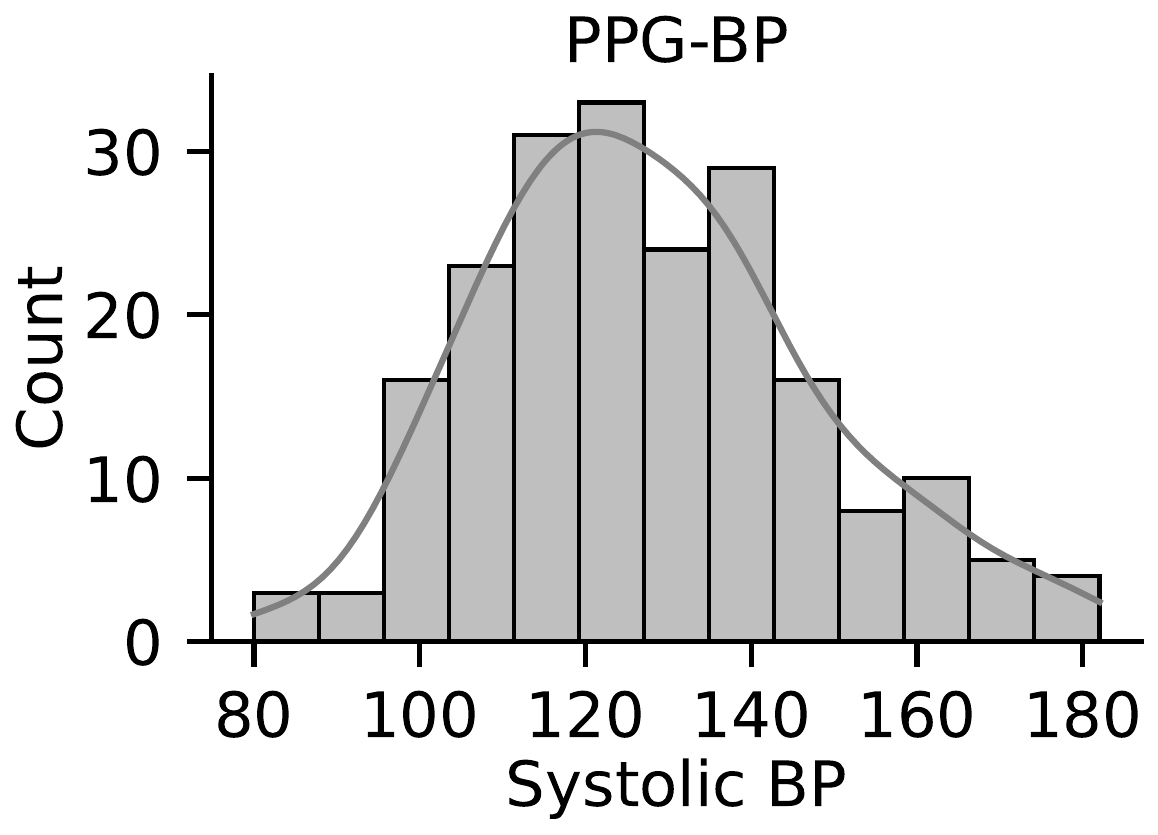}
     \end{subfigure} 
     \hfill
     \begin{subfigure}[b]{0.32\textwidth}
         \includegraphics[width=\textwidth]{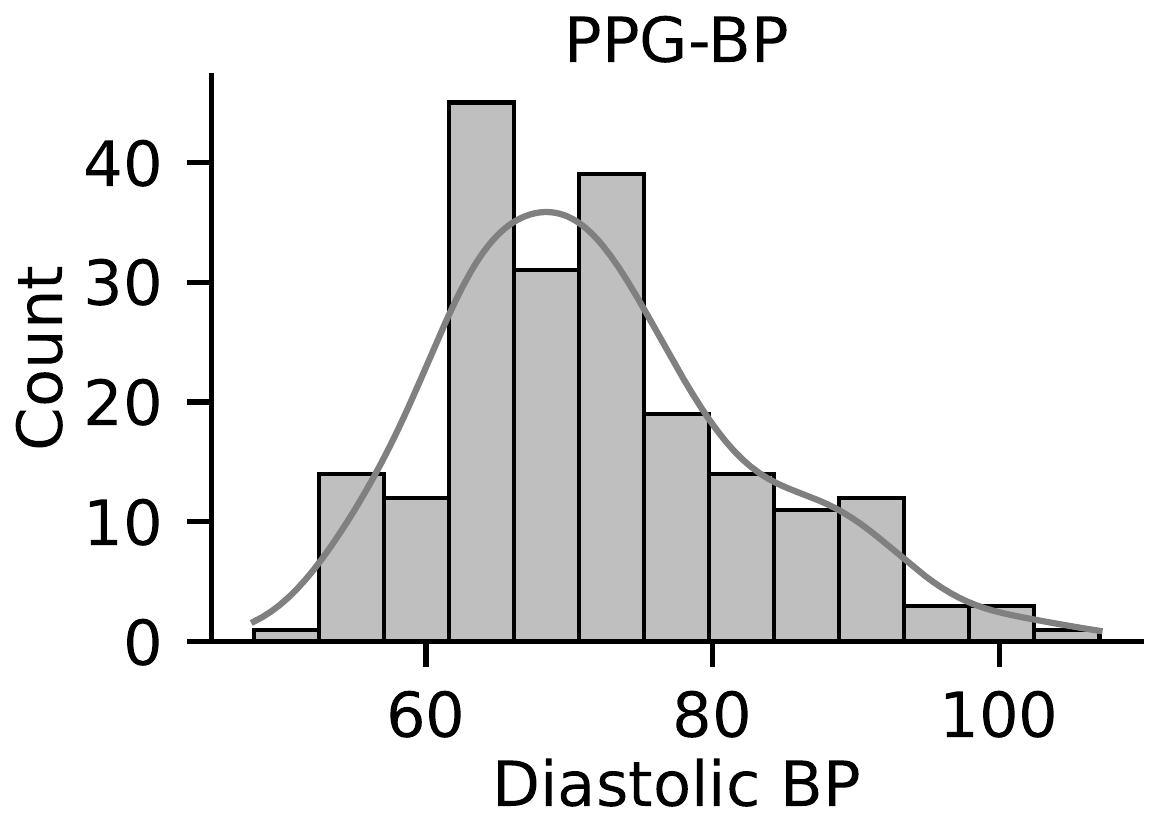}
     \end{subfigure} 
     \hfill
    \begin{subfigure}[b]{0.32\textwidth}
         \includegraphics[width=\textwidth]{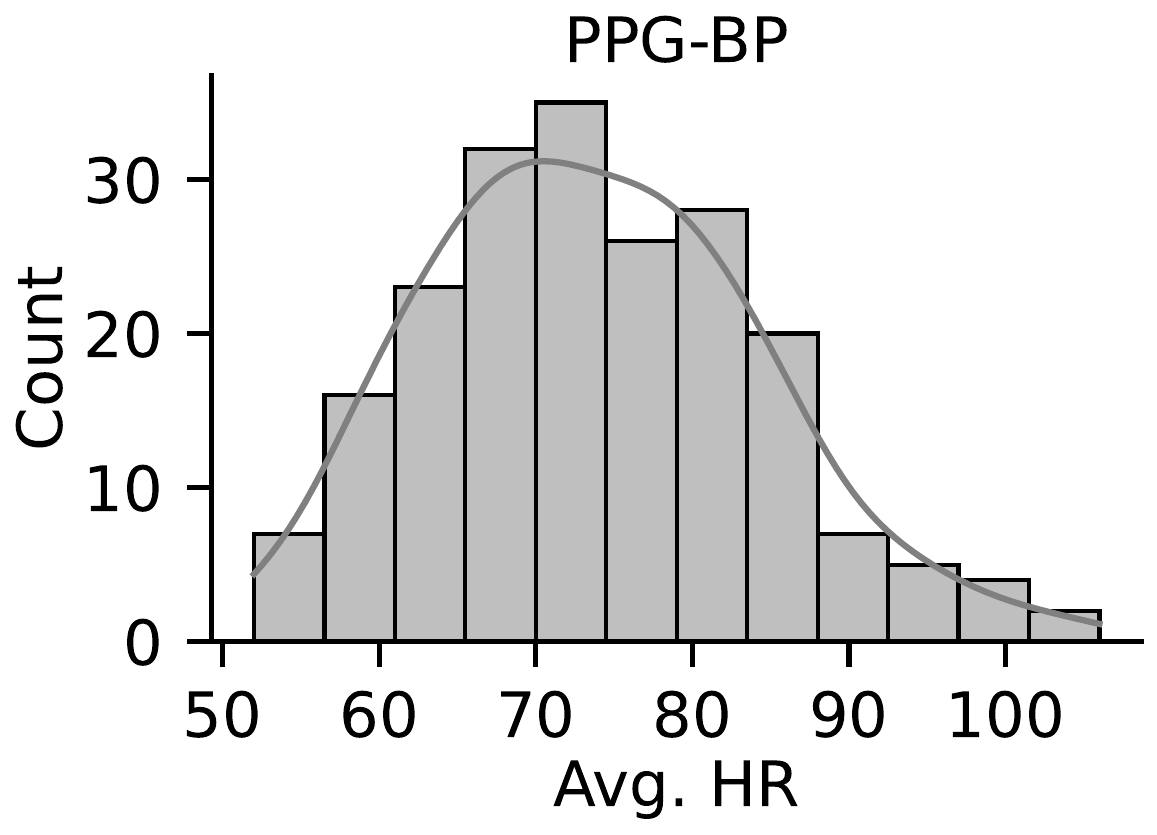}
     \end{subfigure} 
     \hfill
    \caption{PPG-BP dataset descriptive statistics.}
    \label{fig:ppg_bp_dataset}
\end{figure}

\noindent\textbf{PPG-BP.} The PPG-BP consists of short PPG recordings from 219 participants collected at 1000Hz. For each subject, there are three 2.1s recordings. For our analysis, we zero pad them to 10s. In Figure \ref{fig:ppg_bp_dataset}, the age distribution shows that most participants are between 40 and 80 years old, with fewer participants under 40. Furthermore, the majority of individuals have hypertension. In terms of gender, the dataset has slightly more females than males. The distribution of systolic blood pressure is centered around 120-140, indicating a population with normal to moderately elevated blood pressure, while diastolic blood pressure predominantly falls between 70 and 90. Lastly, the average heart rate for most participants ranges between 70 and 90 beats per minute.

\begin{wrapfigure}{r}{0.4\textwidth}
    \centering
    \includegraphics[width=0.7\linewidth]{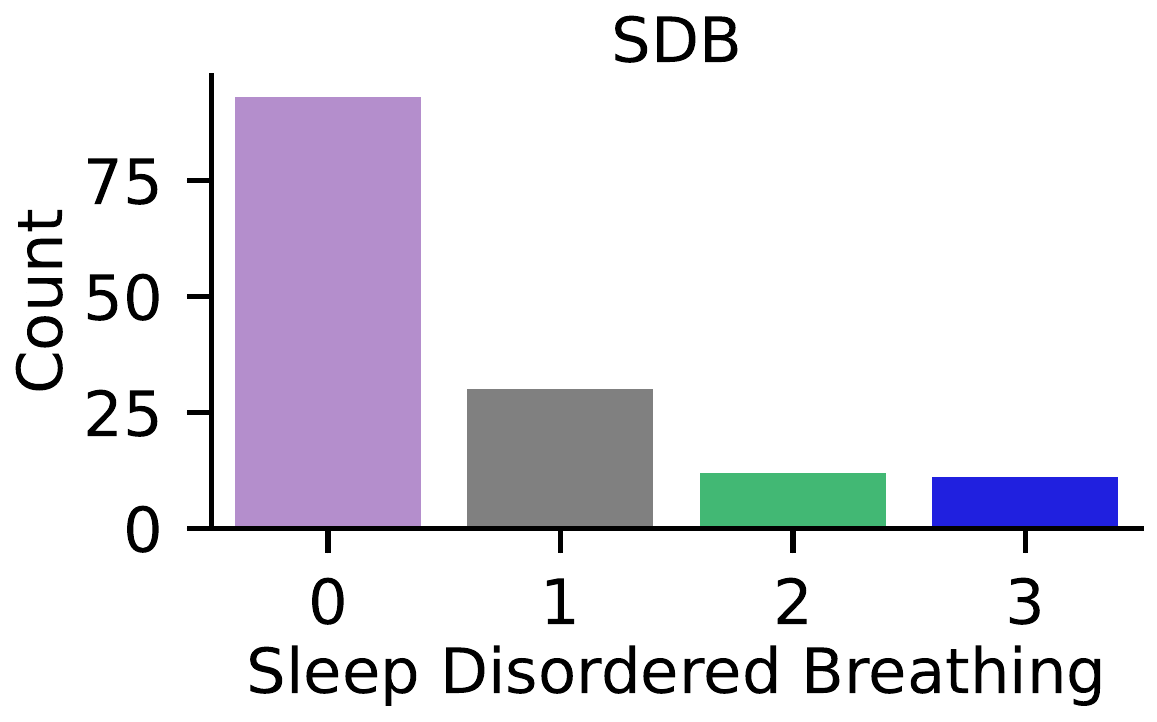}
    \caption{SDB dataset descriptive statistics.}
    \label{fig:sdb_dataset}
\end{wrapfigure}

\noindent\textbf{SDB.} The sleep-disordered breathing dataset includes data from 146 children, collected through polysomnography with finger recordings lasting over three hours. Ground truth labels are provided as Apnea-Hypopnea Index (AHI) values, categorized into four levels: 0 (normal), 1 (mild, AHI between 5 and 15), 2 (moderate, AHI between 15 and 30), and 3 (severe, AHI over 30) as shown in Figure \ref{fig:sdb_dataset}. For our classification task, we group AHI 0 as indicating no sleep breathing disorder, while AHI levels 1 through 3 are classified as the presence of a sleep breathing disorder.

\begin{figure}[h]
    \centering
    \begin{subfigure}[b]{0.32\textwidth}
         \includegraphics[width=\textwidth]{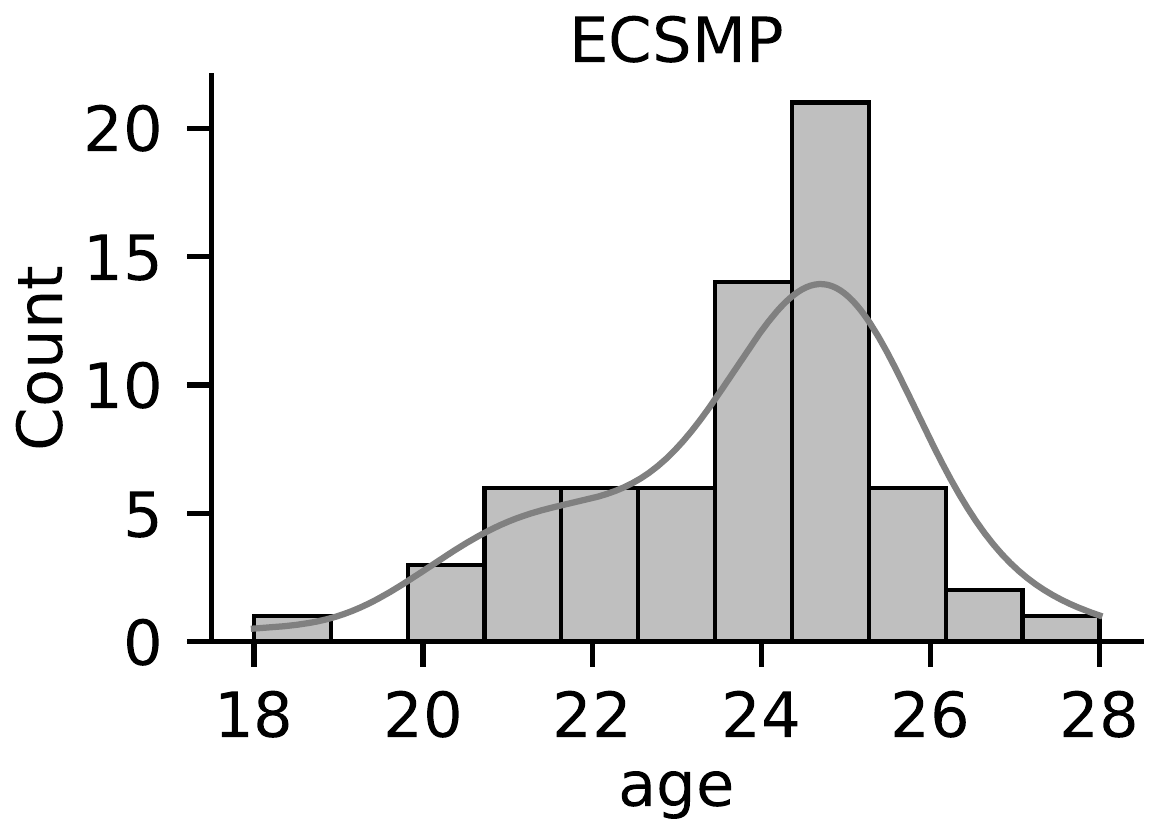}
     \end{subfigure}
     \hfill
    \begin{subfigure}[b]{0.32\textwidth}
         \includegraphics[width=\textwidth]{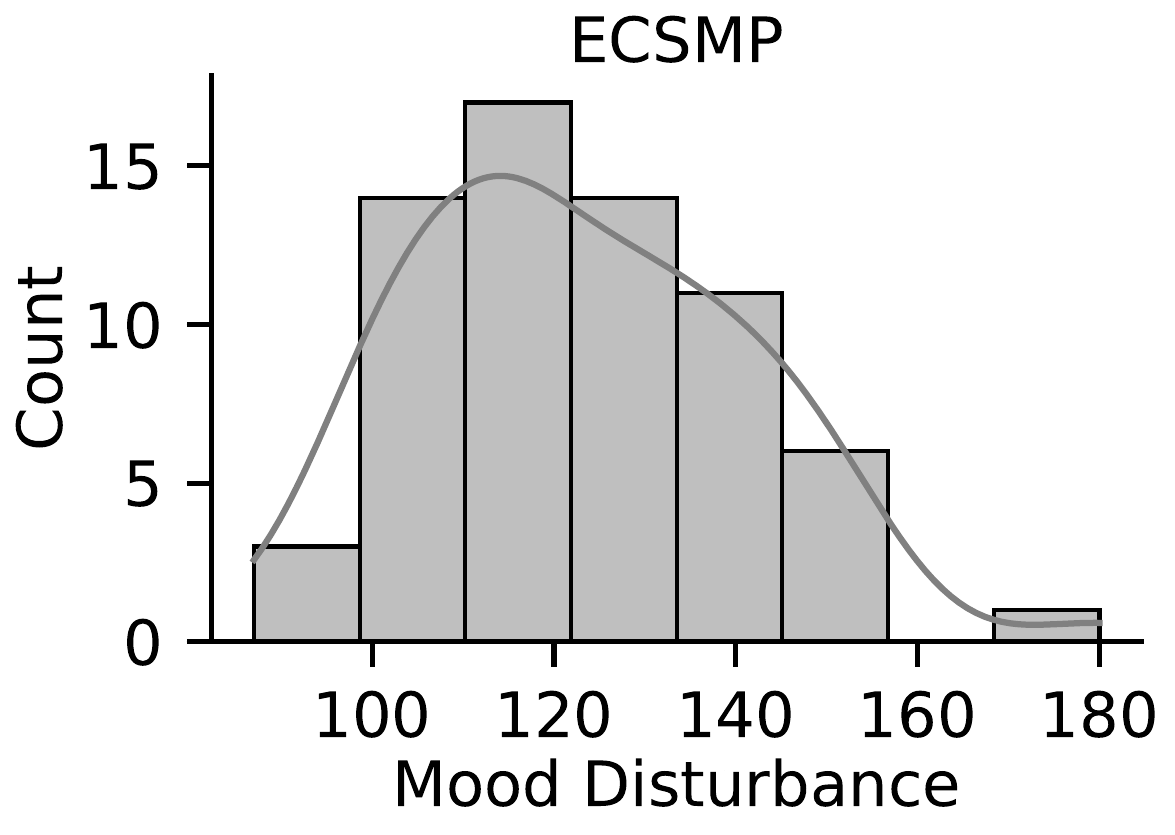}
     \end{subfigure} 
     \hfill 
    \begin{subfigure}[b]{0.32\textwidth}
         \includegraphics[width=\textwidth]{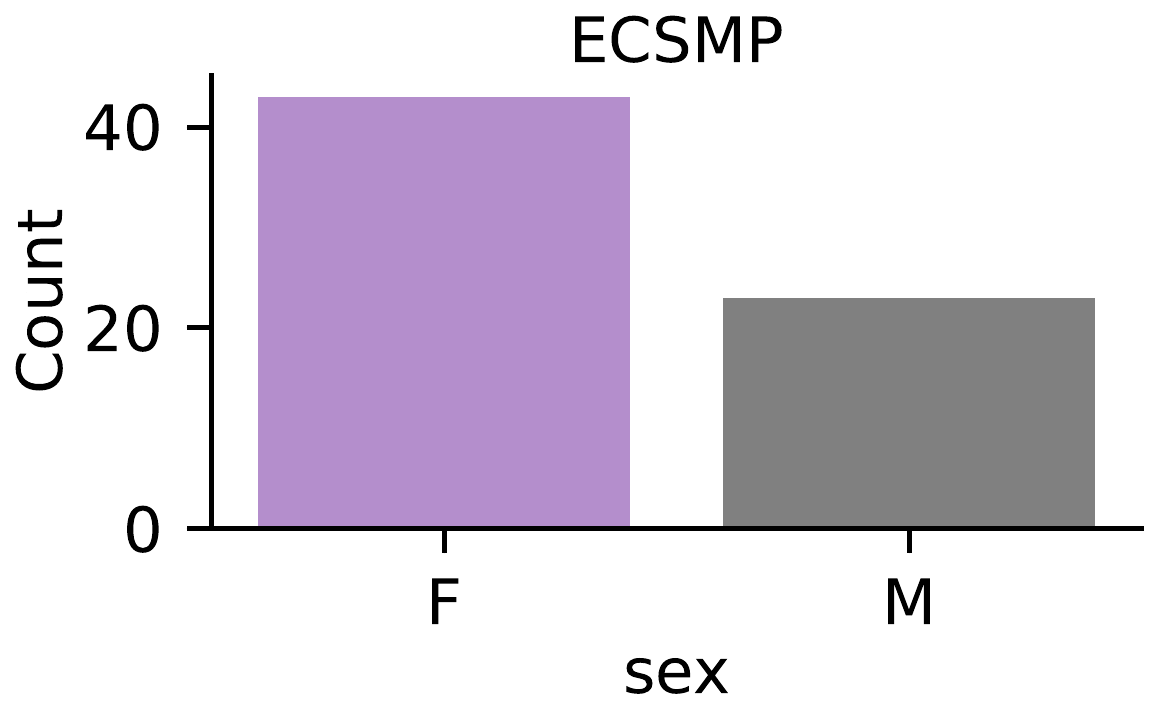}
     \end{subfigure}
     \hfill
    \caption{ECSMP dataset descriptive statistics.}
    \label{fig:ecsmp_dataset}
\end{figure}

\noindent\textbf{ECSMP.} The ECSMP dataset was gathered to study the relationship between emotion, cognition, and sleep in 89 participants. As shown in Figure \ref{fig:ecsmp_dataset}, the majority of the participants are young adult females, with an average age of around 25 years. Mood disturbances were measured using the Profile of Mood States (POMS) scale, which captures various aspects of emotional states. To classify participants into high versus low mood disturbance categories, we binarized the Total Mood Disturbance (TMD) values by using the median as the cutoff point.

\begin{figure}[h]
    \centering
    \begin{subfigure}[b]{0.45\textwidth}
         \includegraphics[width=\textwidth]{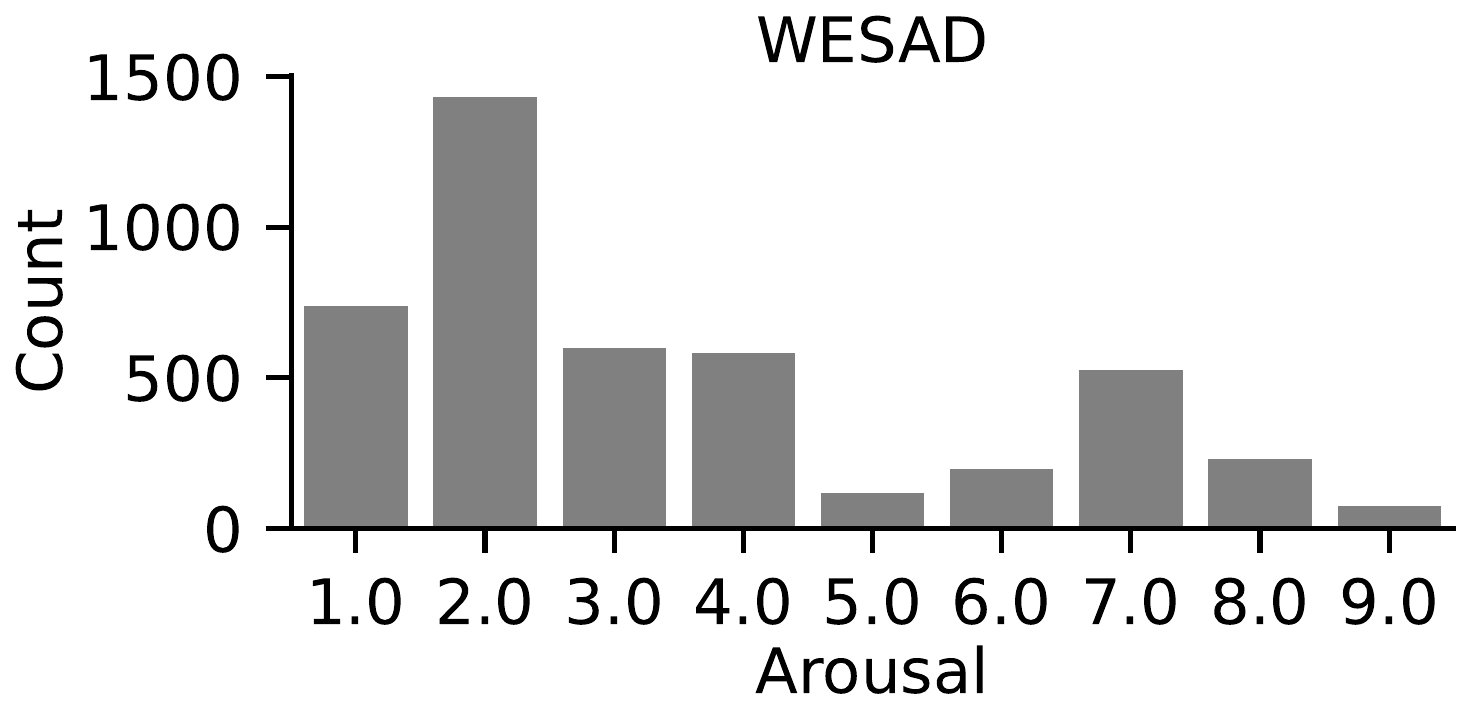}
     \end{subfigure}
     \hfill
    \begin{subfigure}[b]{0.45\textwidth}
         \includegraphics[width=\textwidth]{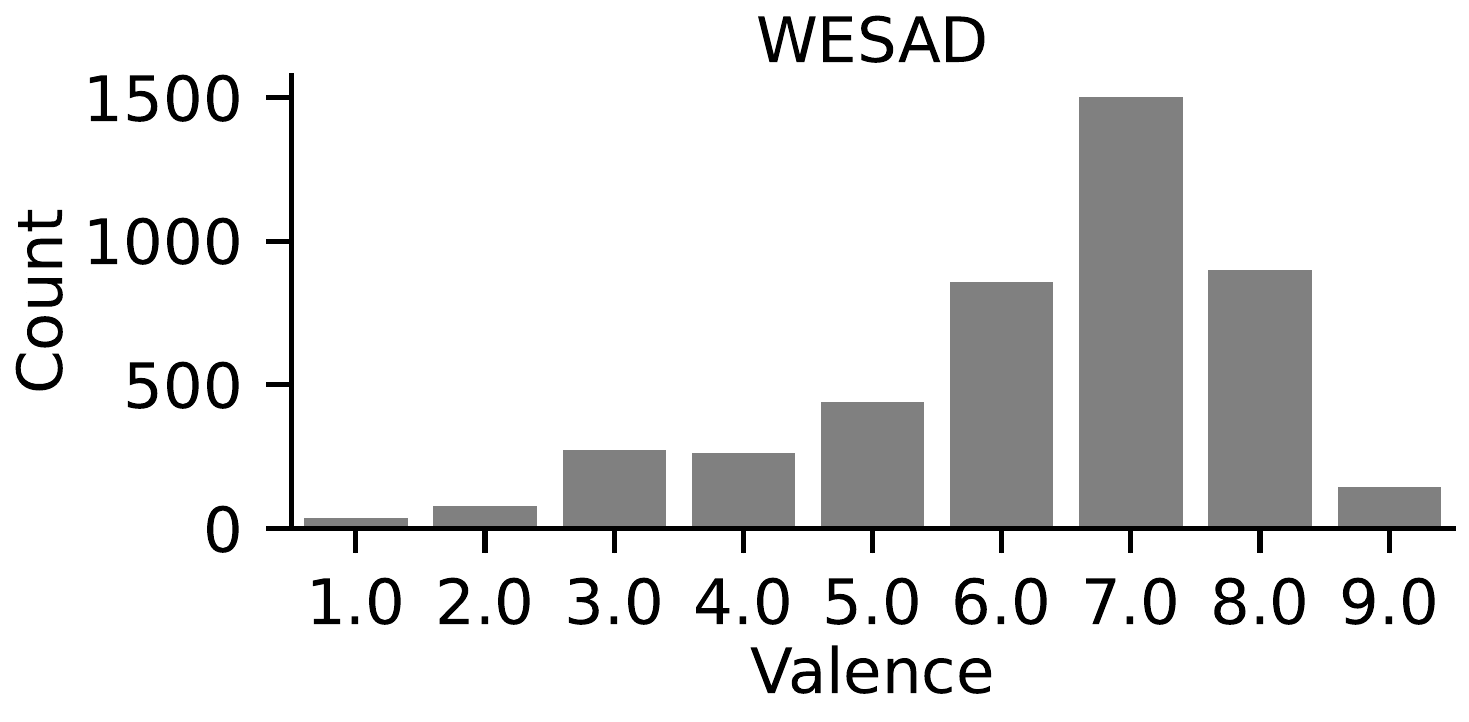}
     \end{subfigure} 
     \hfill 
    \caption{WESAD dataset descriptive statistics.}
    \label{fig:wesad_dataset}
\end{figure}

\noindent\textbf{WESAD.} The wearable stress and affect detection dataset is a multi-modal dataset collected from 15 participants using various sensor modalities. In this study, participants were exposed to videos designed to elicit different affective states, such as amusement, meditation, stress, and baseline conditions. Following each session, participants completed the Self-Assessment Manikins (SAM) questionnaire \citep{bradley1994measuring}, which provided the ground-truth values for valence and arousal. In our analysis, we binarized these values by categorizing valence and arousal as low (1) when less than 5 and high (0) otherwise. Then, we perform regression at the segment level. As shown in Figure \ref{fig:wesad_dataset}, arousal levels are generally low, while valence tends to be high in most cases.

\noindent\textbf{PPG-DaLiA.} This dataset collects PPG signals from 15 participants for heart rate estimation while performing various daily activities. These activities include sitting, ascending/descending stairs, table soccer, cycling, driving, lunch break, walking, and working. As a result, the dataset captures a wide range of heart rates, varying from 60 to 150 beats per minute, depending on the specific activity being performed. To align the PPG signal with the activity labels, we use a 8s window with 6s and 2s overlap and shift, respectively. After this, we resample and pad the signal to facilitate modeling. 

\begin{figure}[h]
    \centering
    \begin{subfigure}[b]{0.45\textwidth}
         \includegraphics[width=\textwidth]{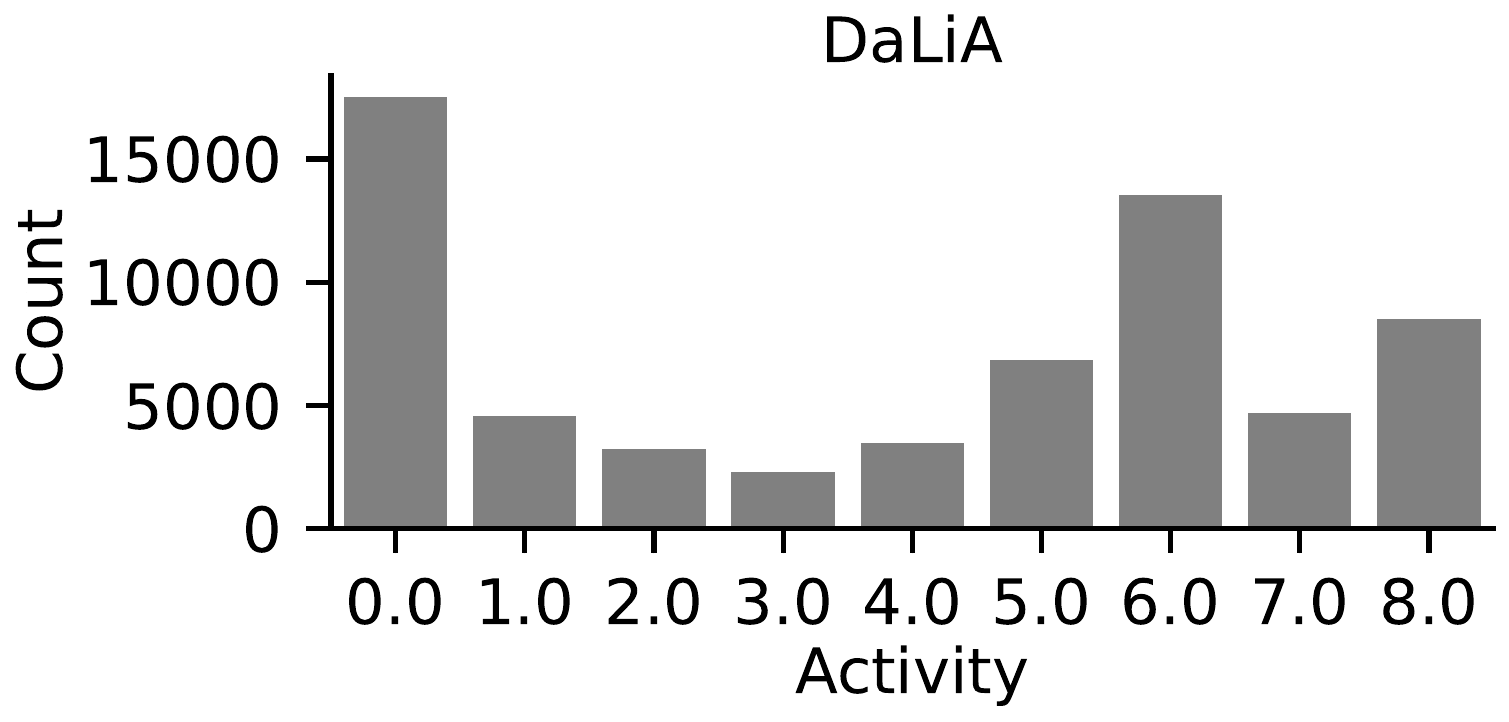}
     \end{subfigure}
     \hfill
    \begin{subfigure}[b]{0.45\textwidth}
         \includegraphics[width=\textwidth]{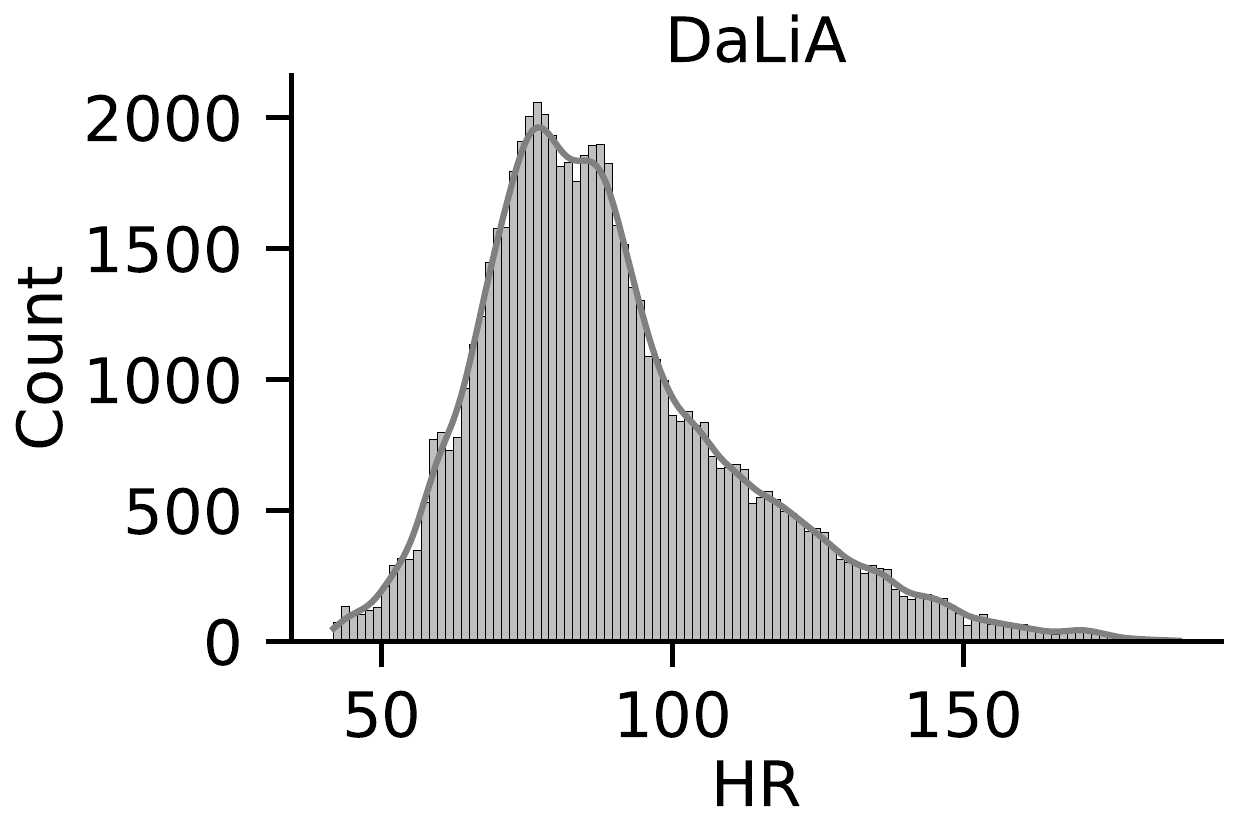}
     \end{subfigure} 
     \hfill 
    \caption{PPG-DaLiA dataset descriptive statistics.}
    \label{fig:ppg_dalia_dataset}
\end{figure}

\newpage
\section{Representative Signals from Pre-training Datasets}
\begin{figure}[h]
    \centering
    \includegraphics[width=0.8\linewidth]{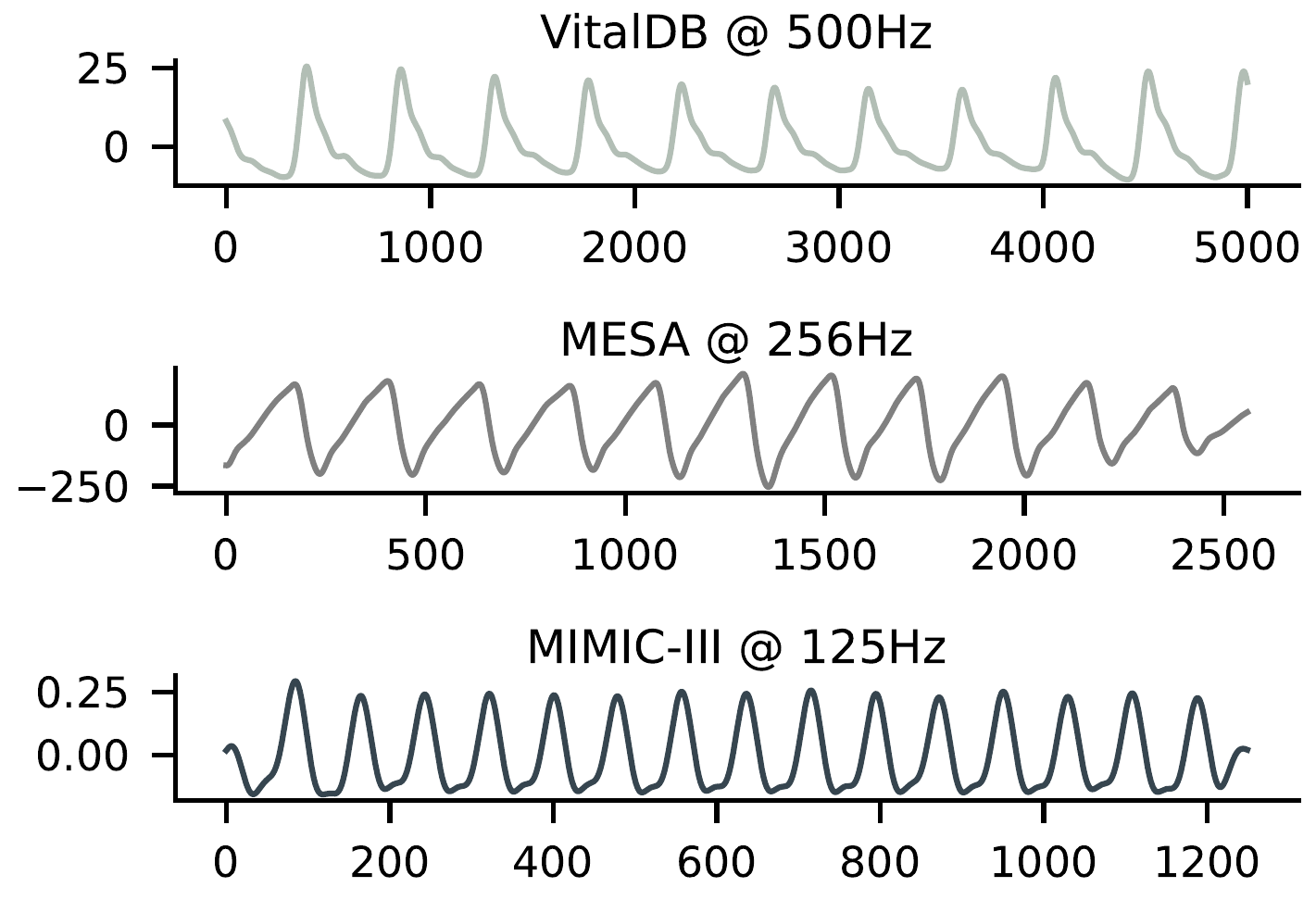}
    \caption{Representative 10-second \textit{raw} PPG segments from VitalDB, MESA, and MIMIC-III. We observe that each signal's amplitude (y-axis) and sampling rate differ.}
    \label{fig:representative_signals_raw}
\end{figure}
\begin{figure}[h]
    \centering
    \includegraphics[width=0.8\linewidth]{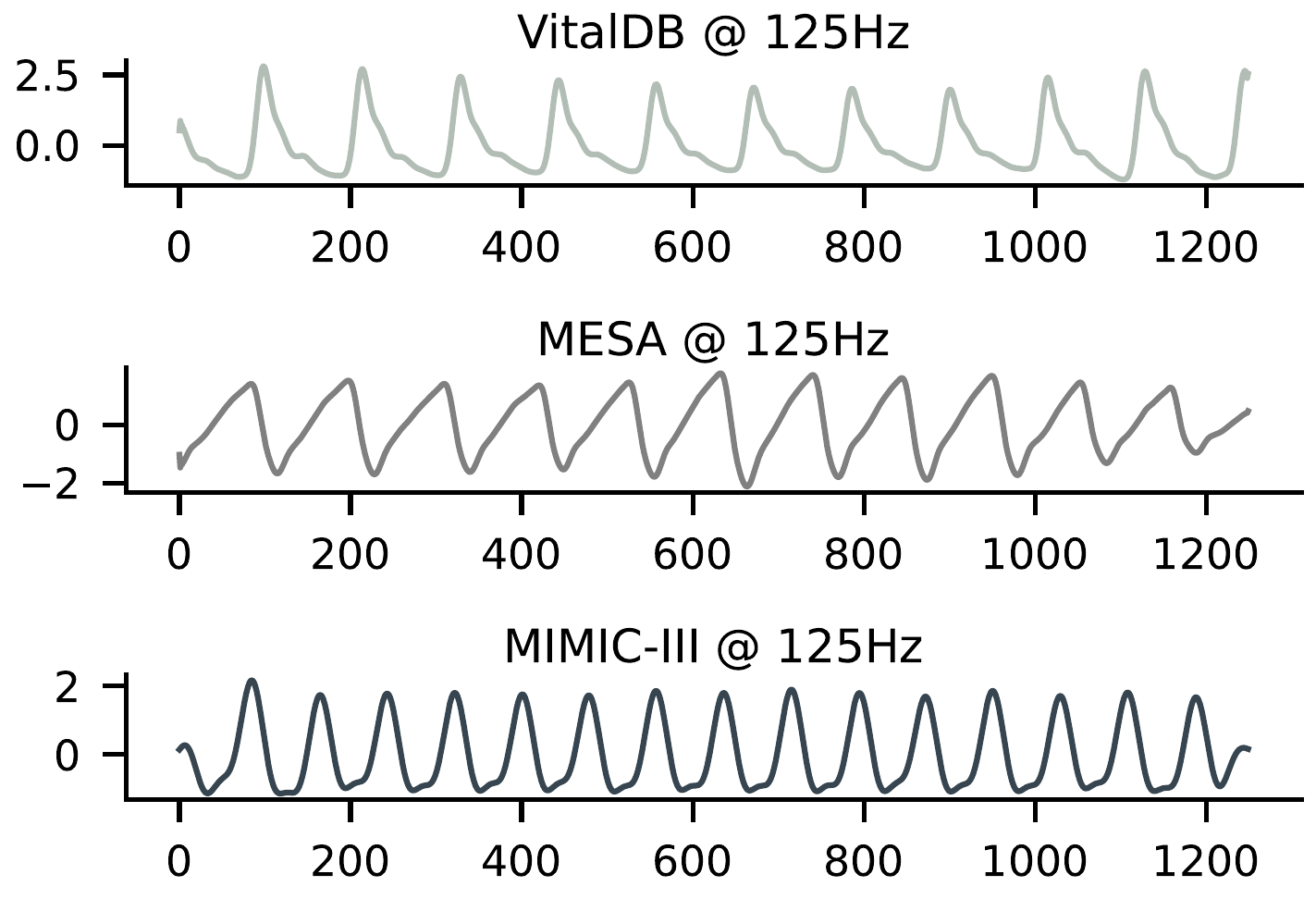}
    \caption{Normalized and resampled 10-second \textit{pre-processed} PPG segments from VitalDB, MESA, and MIMIC-III. These signals represent the final form before being fed to our models. We observe that the signal characteristics across datasets are more consistent.}
    \label{fig:representative_signals_training}
\end{figure}

\newpage
\section{Additional Results} \label{appendix:additional_results}

\subsection{Multi-class Classification}
\begin{table}[h]
\centering
\caption{\textbf{Multi-class classification comparison against pre-trained models.} Feature extraction parameters are indicated next to each name.. 95\% CIs are reported in square brackets and the best value is \textbf{bolded}.}
\label{tab:multi_classification_fm}
\scalebox{0.76}{
\begin{tabular}{@{}l|lll|ll@{}}
\toprule
 &\textbf{REGLE} \smat{(0.07M)} &\textbf{Chronos} \smat{(200M)} &\textbf{Moment}  \smat{(385M)} &\textbf{\model{}-P} \smat{(5M)} &\textbf{\model{}-S} \smat{(5M)}  \\ 
\textbf{Classification} - ACC ($\uparrow$) &\smat{\citep{yun2024unsupervised}} &\smat{\citep{ansari2024chronos}} &\smat{\citep{goswami2024moment}} & & \\\midrule
Operation Type &0.21 \smat{[0.18-0.23]} &0.25 \smat{[0.22-0.29]} &0.27 \smat{[0.23-0.31]} &0.30 \smat{[0.26-0.33]} & \textbf{0.30} \smat{[0.27-0.32]} \\
Activity &0.29 \smat{[0.28-0.29]} &\textbf{0.41} \smat{[0.40-0.42]} &\textbf{0.41} \smat{[0.40-0.42]} &0.38 \smat{[0.37-0.39]} &0.37 \smat{[0.36-0.37]} \\
\midrule
Average &0.25 $\pm$ 0.04 &0.33 $\pm$ 0.08 &\textbf{0.34 $\pm$ 0.07} &\textbf{0.34 $\pm$ 0.04} &0.33 $\pm$ 0.03 \\
     \bottomrule
\end{tabular}}
\end{table}

\begin{table}[h]
\centering
\caption{\textbf{Multi-class classification comparison against CL methods}. Feature extraction parameters are indicated next to each name.. 95\% CIs are reported in square brackets and the best value is \textbf{bolded}.}
\label{tab:multi_classification_ssl}
\scalebox{0.7}{
\begin{tabular}{@{}l|l|lll|ll@{}}
\toprule
 &\textbf{Stat. Features} &\textbf{SimCLR} \smat{(5M)} &\textbf{BYOL} \smat{(5M)} &\textbf{TF-C} \smat{(10M)} &\textbf{\model{}-P} \smat{(5M)} &\textbf{\model{}-S} \smat{(5M)} \\ 
 \textbf{Classification} - ACC ($\uparrow$) & &\smat{\citep{chen2020simple}} &\smat{\citep{grill2020bootstrap}} &\smat{\citep{zhang2022self}} & & \\ \midrule
Operation Type &0.27 \smat{[0.24-0.32]} &0.27 \smat{[0.24-0.29]} &\textbf{0.31} \smat{0.27-0.34} &0.27 \smat{[0.27-0.29]} &0.30 \smat{[0.26-0.33]} & 0.30 \smat{[0.27-0.32]} \\
Activity &0.37 \smat{[0.36-0.38]} &0.36 \smat{[0.35-0.37]} &0.34 \smat{[0.33-0.35]} &0.37 \smat{[0.36-0.38]} &\textbf{0.38} \smat{[0.37-0.39]} &0.37 \smat{[0.36-0.37]} \\
\midrule
Average &0.32 $\pm$ 0.05 &0.31 $\pm$ 0.04 &0.32 $\pm$ 0.01 &0.32 $\pm$ 0.05 &\textbf{0.34 $\pm$ 0.04} &0.33 $\pm$ 0.03 \\
     \bottomrule
\end{tabular}
}
\end{table}

\subsection{F1-Score and $R^2$ evaluation metrics.}
\begin{table}[h]
\centering
\caption{\textbf{Downstream comparison against pre-trained models (additional metrics: F1-score and $R^2$).} Feature extraction parameters are indicated next to each name. 95\% CIs are reported in square brackets and the best value is \textbf{bolded}.}
\label{tab:binary_classification_fm_f1}
\scalebox{0.76}{
\begin{tabular}{@{}l|lll|ll@{}}
\toprule
 &\textbf{REGLE} \smat{(0.07M)} &\textbf{Chronos} \smat{(200M)} &\textbf{Moment}  \smat{(385M)} &\textbf{\model{}-P} \smat{(5M)} &\textbf{\model{}-S} \smat{(5M)}  \\ 
\textbf{Classification} - F1-Score ($\uparrow$) &\smat{\citep{yun2024unsupervised}} &\smat{\citep{ansari2024chronos}} &\smat{\citep{goswami2024moment}} & & \\\midrule
ICU Admission &0.00 \smat{[0.00-0.00]} & 0.20 \smat{[0.11-0.30]} &0.12 \smat{[0.04-0.20]} &0.12 \smat{[0.04-0.20]} &\textbf{0.26} \smat{[0.18-0.33]} \\
Mortality &0.00 \smat{[0.00-0.00]} &0.14 \smat{[0.09-0.19]} &0.16 \smat{[0.11-0.21]} &\textbf{0.22} \smat{[0.16-0.27]} &0.17 \smat{[0.13-0.22]} \\
Smoker &0.16 \smat{[0.10-0.23]} &\textbf{0.51} \smat{[0.44-0.58]} &0.40 \smat{[0.33-0.47]} &0.45 \smat{[0.38-0.51]} &0.45 \smat{[0.37-0.50]} \\
Pregnancy stage &0.49 \smat{[0.47-0.52]} &\textbf{0.69} \smat{[0.67-0.71]} &0.63 \smat{[0.60-0.65]} &0.62 \smat{[0.59-0.64]} &0.65 \smat{[0.62-0.67]} \\
Hypertension &0.77 \smat{[0.70-0.84]} &0.68 \smat{[0.58-0.77]} &0.75 \smat{[0.66-0.84]} &\textbf{0.84} \smat{[0.72-0.92]} &0.78 \smat{[0.70-0.86]} \\
Sleep Disordered Breathing &0.00 \smat{[0.00-0.00]} &0.33 \smat{[0.00-0.60]} &0.22 \smat{[0.00-0.47]} &0.32 \smat{[0.00-0.60]} &\textbf{0.47} \smat{[0.23-0.67]} \\
Mood Disturbance &0.00 \smat{[0.00-0.00]} &0.36 \smat{[0.10-0.59]} &0.23 \smat{[0.00-0.47]} &\textbf{0.37} \smat{[0.00-0.66]} &0.32 \smat{[0.00-0.58]} \\
Valence &0.00 \smat{[0.00-0.00]} &0.10 \smat{[0.07-0.14]} &0.12 \smat{[0.09-0.16]} &\textbf{0.17} \smat{[0.13-0.21]} &0.03 \smat{[0.01-0.04]} \\
Arousal &0.83 \smat{[0.81-0.84]} &0.82 \smat{[0.80-0.83]} &0.81 \smat{[0.79-0.82]} &0.81 \smat{[0.79-0.82]} &\textbf{0.83} \smat{[0.81-0.84]} \\
\midrule
Average &0.25 $\pm$ 0.33 &0.42 $\pm$ 0.24 &0.38 $\pm$ 0.26 &0.43 $\pm$ 0.25 &\textbf{0.44 $\pm$ 0.25}\\
\midrule
\textbf{Regression} - $R^2$ ($\uparrow$)& & & & & \\
\midrule
Apnea/Hypopnea Index $>$ 3\% &0.02 \smat{[0.00-0.03]} &0.18 \smat{[0.08-0.26]} &0.14 \smat{[0.06-0.22]} &0.15 \smat{[0.05-0.24]} &\textbf{0.29} \smat{[0.22-0.36]} \\
Apnea/Hypopnea Index $>$ 4\% &0.01 \smat{[0.00-0.03]} &0.16 \smat{[0.08-0.22]} &0.13 \smat{[0.05-0.20]} &0.12 \smat{[0.03-0.22]} &\textbf{0.28} \smat{[0.20-0.34]} \\
Gestation Age &0.04 \smat{[0.02-0.06]} &\textbf{0.28} \smat{[0.24-0.31]} &0.20 \smat{[0.17-0.23]} &0.18 \smat{[0.14-0.22]} &0.22 \smat{[0.19-0.25]} \\
Systolic BP (VV) &-0.03 \smat{[-0.18-0.01]} &-0.24 \smat{[-0.72-0.03]} &0.06 \smat{[-0.25-0.28]} &-0.41 \smat{[-0.77-(-0.15)]} &\textbf{0.15} \smat{[-0.09-0.30]} \\
Diastolic BP (VV) &0.01 \smat{[-0.09-0.06]} &-0.29 \smat{[-0.87-(-0.01)]} &0.01 \smat{[-0.25-0.14]} &-0.48 \smat{[-1.02-(-0.20)]} &\textbf{0.10} \smat{[-0.11-0.23]} \\
Systolic BP (PPG-BP) &-0.07 \smat{[-0.21-0.04]} &-0.13 \smat{[-0.36-0.06]} &0.07 \smat{[-0.31 -0.31]} &\textbf{0.36} \smat{[0.16-0.49]} &0.20 \smat{[0.02-0.31]}\\
Diastolic BP (PPG-BP) &0.01 \smat{[-0.05-0.02]} &-0.10 \smat{[-0.45-0.07]} &-0.03 \smat{[-0.31-0.13]} &\textbf{0.22} \smat{[-0.13-0.40]} &0.08 \smat{[-0.07-0.17]} \\
Average HR &0.37 \smat{[0.17-0.51]} &0.02 \smat{[-0.16-0.17]} &0.68 \smat{[0.45-0.80]} &\textbf{0.79} \smat{[0.57-0.90]} &0.78 \smat{[0.69-0.83]} \\
HR &0.00 \smat{[0.00-0.01]} &0.57 \smat{[0.56-0.59]} &\textbf{0.63} \smat{[0.61-0.64]} &0.52 \smat{[0.51-0.53]} &0.48 \smat{[0.42-0.46]} \\
\midrule
Average &0.04 $\pm$ 0.12 &0.05 $\pm$ 0.25 &0.21 $\pm$ 0.24 &0.16 $\pm$ 0.38 &\textbf{0.28 $\pm$ 0.20} \\
\bottomrule
\end{tabular}}
\end{table}

\begin{table}[h]
\centering
\caption{\textbf{Downstream comparison against CL models (additional metrics: F1-score and $R^2$).} Feature extraction parameters are indicated next to each name. 95\% CIs are reported in square brackets and the best value is \textbf{bolded}.}
\label{tab:binary_classification_ssl_f1}
\scalebox{0.70}{
\begin{tabular}{@{}l|l|lll|ll@{}}
\toprule
 &\textbf{Stat. Features} &\textbf{SimCLR} \smat{(5M)} &\textbf{BYOL} \smat{(5M)} &\textbf{TF-C} \smat{(10M)} &\textbf{\model{}-P} \smat{(5M)} &\textbf{\model{}-S} \smat{(5M)}  \\ 
\textbf{Classification} - F1-Score ($\uparrow$) & & & & & & \\\midrule
ICU Admission &\textbf{0.30} \smat{[0.18-0.40]} &0.19 \smat{[0.12-0.26]} &0.17 \smat{[0.11-0.22]} &0.10 \smat{[0.05-0.16]} &0.12 \smat{[0.04-0.20]} &0.26 \smat{[0.18-0.33]} \\
Mortality &0.03 \smat{[0.01-0.06]} &0.15 \smat{[0.10-0.20]} &0.13 \smat{[0.08-0.17]} &0.15 \smat{[0.10-0.20]} &\textbf{0.22} \smat{[0.16-0.27]} &0.17 \smat{[0.13-0.22]} \\
Smoker &0.47 \smat{[0.40-0.53]} &0.43 \smat{[0.35-0.49]} &\textbf{0.49} \smat{[0.42-0.56]} &0.37 \smat{[0.30-0.44]} &0.45 \smat{[0.38-0.51]} &0.45 \smat{[0.37-0.50]} \\
Pregnancy stage &0.43 \smat{[0.41-0.47]} &0.60 \smat{[0.57-0.63]} &0.60 \smat{[0.57-0.63]} &0.59 \smat{[0.56-0.62]} &0.62 \smat{[0.59-0.64]} &\textbf{0.65} \smat{[0.62-0.67]} \\
Hypertension &0.73 \smat{[0.58-0.85]} &0.82 \smat{[0.73-0.89]} &0.81 \smat{[0.73-0.88]} &0.81 \smat{[0.72-0.89]} &\textbf{0.84} \smat{[0.72-0.92]} &0.78 \smat{[0.70-0.86]} \\
Sleep Disordered Breathing &0.00 \smat{[0.00-0.00]} &0.46 \smat{[0.21-0.64]} &0.45 \smat{[0.23-0.62]} &0.19 \smat{[0.00-0.36]} &0.32 \smat{[0.00-0.60]} &\textbf{0.47} \smat{[0.23-0.67]} \\
Mood Disturbance &0.21 \smat{[0.00-0.47]} &0.21 \smat{[0.00-0.44]} & 0.37 \smat{[0.00-0.67]} &\textbf{0.56} \smat{[0.27-0.80]} &0.37 \smat{[0.00-0.66]} &0.32 \smat{[0.00-0.58]} \\
Valence &0.04 \smat{[0.02-0.07]} &0.09 \smat{[0.06-0.12]} &0.01 \smat{[0.00-0.03]} &0.07 \smat{[0.04-0.09]} &\textbf{0.17} \smat{[0.13-0.21]} &0.03 \smat{[0.01-0.04]} \\
Arousal &0.82 \smat{[0.81-0.83]} &0.81 \smat{[0.79-0.82]} &\textbf{0.83} \smat{[0.81-0.84]} &0.81 \smat{[0.80-0.83]} &0.81 \smat{[0.79-0.82]} &\textbf{0.83} \smat{[0.81-0.84]} \\
\midrule
Average &0.33 $\pm$ 0.28 &0.42 $\pm$ 0.26 &0.43 $\pm$ 0.27 &0.40 $\pm$ 0.27 &0.43 $\pm$ 0.25 &\textbf{0.44 $\pm$ 0.25}\\
\midrule
\textbf{Regression} - $R^2$ ($\uparrow$)& & & & & & \\
\midrule
Apnea/Hypopnea Index $>$ 3\% &-0.00 \smat{[-0.06-0.03]} &0.16 \smat{[0.07-0.23]} &0.16 \smat{[0.08-0.22]} &0.06 \smat{[-0.00-0.13]} &0.15 \smat{[0.05-0.24]} &\textbf{0.29} \smat{[0.22-0.36]} \\
Apnea/Hypopnea Index $>$ 4\% &-0.01 \smat{[-0.07-0.03]} &0.13 \smat{[0.06-0.21]} &0.13 \smat{[0.05-0.19]} &0.13 \smat{[-0.06-0.26]} &0.12 \smat{[0.03-0.22]} &\textbf{0.28} \smat{[0.20-0.34]} \\
Gestation Age &0.07 \smat{[0.04-0.10]} &0.19 \smat{[0.15-0.21]} &0.19 \smat{[0.15-0.22]} &0.18 \smat{[0.15-0.21]} &0.18 \smat{[0.14-0.22]} &\textbf{0.22} \smat{[0.19-0.25]} \\
Systolic BP (VV) &-0.10 \smat{[-0.51-0.10]} &-0.05 \smat{[-0.44-0.21]} &-0.03 \smat{[-0.37-0.18]} &-0.05 \smat{[-0.36-0.12]} &-0.41 \smat{[-0.77-(-0.15)]} &\textbf{0.15} \smat{[-0.09-0.30]} \\
Diastolic BP (VV) &-0.15 \smat{[-0.31-0.11]} &-0.14 \smat{[-0.29-0.08]} &-0.01 \smat{[-0.40-0.20]} &-0.09 \smat{[-0.45-0.16]} &-0.48 \smat{[-1.02-(-0.20)]} &\textbf{0.10} \smat{[-0.11-0.23]} \\
Systolic BP (PPG-BP) &0.12 \smat{[-0.04-0.21]} &0.09 \smat{[-0.20-0.31]} &0.10 \smat{[-0.16-0.30]} &0.13 \smat{[-0.06-0.26]} &\textbf{0.36} \smat{[0.16-0.49]} &0.20 \smat{[0.02-0.31]}\\
Diastolic BP (PPG-BP) &0.01 \smat{[-0.18-0.14]} &0.00 \smat{[-0.20-0.18]} &0.05 \smat{[-0.11-0.17]} &0.02 \smat{[-0.15-0.12]} &\textbf{0.22} \smat{[-0.13-0.40]} &0.08 \smat{[-0.07-0.17]} \\
Average HR &0.15 \smat{[-0.10-0.33]} &0.74 \smat{[0.64-0.80]} &0.65 \smat{[0.50-0.77]} &\textbf{0.82} \smat{[0.73-0.88]} &0.79 \smat{[0.57-0.90]} &0.78 \smat{[0.69-0.83]} \\
HR &0.34 \smat{[0.32-0.36]} &0.45 \smat{[0.44-0.47]} &0.36 \smat{[0.35-0.37]} &\textbf{0.54} \smat{[0.53-0.55]} &0.52 \smat{[0.51-0.53]} &0.48 \smat{[0.42-0.46]} \\
\midrule
Average &0.05 $\pm$ 0.14 &0.17 $\pm$ 0.25 &0.18 $\pm$ 0.20 &0.19 $\pm$ 0.28 &0.16 $\pm$ 0.38 &\textbf{0.28 $\pm$ 0.20} \\
\bottomrule
\end{tabular}}
\end{table}

\subsection{Ablation Results}
In this section, we provide the numeric results for the scaling analysis (Table \ref{tab:scaling_analysis}) and \model{}-S component analysis (Table \ref{tab:component_analysis}).

\begin{table}[]
\centering
\caption{\textbf{Scaling: Downstream comparison for different \model{}-S models}. 95\% CIs are reported in square brackets and the best value is \textbf{bolded}.}
\label{tab:scaling_analysis}
\scalebox{0.76}{
\begin{tabular}{@{}l|llll@{}}
\toprule
 &\textbf{\model{}-S-5M} &\textbf{\model{}-S-35M} &\textbf{\model{}-S-139M}  \\ 
\textbf{Classification} - AUROC ($\uparrow$) & & &  \\\midrule
ICU Admission &\textbf{0.79} \smat{[0.75-0.82]} & 0.72 \smat{[0.68-0.75]} & 0.77 \smat{[0.73-0.80]} \\
Mortality &\textbf{0.67} \smat{[0.63-0.70]} & 0.66 \smat{[0.63-0.70]} & 0.66 \smat{[0.63-0.69]} \\
Smoker &\textbf{0.61} \smat{[0.56-0.66]} &0.58 \smat{[0.52-0.64]}  &0.59 \smat{[0.54-0.65]} \\
Pregnancy stage &\textbf{0.78} \smat{[0.75-0.80]} & 0.77 \smat{[0.75-0.79]} &0.76 \smat{[0.74-0.78]} \\
Hypertension &\textbf{0.77} \smat{[0.68-0.87]} & 0.75 \smat{[0.64-0.85]} & \textbf{0.77} \smat{[0.65-0.87]} \\
Sleep Disordered Breathing &\textbf{0.70} \smat{[0.57-0.84]} & 0.59 \smat{[0.44-0.74]} & 0.62 \smat{[0.46-0.78]} \\
Mood Disturbance &\textbf{0.56} \smat{[0.33-0.77]} & 0.53 \smat{[0.30-0.73]} & 0.54 \smat{[0.29-0.78]} \\
Valence &\textbf{0.56} \smat{[0.54-0.59]} & 0.53 \smat{[0.50-0.56]} & 0.54 \smat{[0.51-0.56]} \\
Arousal &\textbf{0.55} \smat{[0.52-0.57]} & 0.52 \smat{[0.49-0.55]} & \textbf{0.55} \smat{[0.52-0.58]} \\
\midrule
Average &\textbf{0.67 $\pm$ 0.09} & 0.63 $\pm$ 0.09 &0.63 $\pm$ 0.10 \\
\midrule
\textbf{Regression} - MAE ($\downarrow$)& & &  \\
\midrule
Apnea/Hypopnea Index $>$ 3\% &12.97 \smat{[11.87-14.05]} & 13.07 \smat{[11.92-14.25]} & \textbf{12.86} \smat{[11.79-13.94]} \\
Apnea/Hypopnea Index $>$ 4\% &\textbf{10.56} \smat{[9.59-11.62]} &10.79 \smat{[9.85-11.83]}  & 10.65 \smat{[9.62-11.68]} \\
Gestation Age &\textbf{6.05} \smat{[5.91-6.17]} & 6.10 \smat{[5.94-6.24]} & 6.17 \smat{[6.02-6.30]} \\
Systolic BP (VV) &\textbf{14.65} \smat{[12.50-16.78]} & 15.10 \smat{[13.10-17.21]} & 14.95 \smat{[12.87-17.01]} \\
Diastolic BP (VV) &\textbf{8.29} \smat{[6.61-10.22]} & 9.20 \smat{[6.93-11.12]} & 8.95 \smat{[6.72-10.95]} \\
Systolic BP (PPG-BP) &\textbf{14.39} \smat{[12.53-16.45]} &16.70 \smat{[14.25-19.38]}  &16.20 \smat{[13.73-18.85]}  \\
Diastolic BP (PPG-BP) &\textbf{8.71} \smat{[7.18-10.01]} &9.48 \smat{[8.24-10.90]}  &9.32 \smat{[7.90-10.69]}  \\
Average HR &\textbf{4.00} \smat{[3.34-4.67]} &4.76 \smat{[3.94-5.86]} &4.71 \smat{[3.86-5.60]}  \\
HR &\textbf{11.53} \smat{[11.40-11.66]} &12.86 \smat{[12.73-12.99]}  &12.20 \smat{[12.07-12.34]}  \\
\midrule
Average &\textbf{10.12 $\pm$ 3.47} &10.89 $\pm$ 3.73  &10.76 $\pm$ 3.57 \\
\bottomrule
\end{tabular}}
\end{table}

\begin{table}[]
\centering
\caption{\textbf{\model{} component ablation study results.}}
\label{tab:component_analysis}
\scalebox{0.76}{
\begin{tabular}{@{}l|llll@{}}
\toprule
 &\textbf{sVRI} &\textbf{sVRI + SQI} &\textbf{sVRI + IPA} &\textbf{Full} \\ 
\textbf{Classification} - AUROC ($\uparrow$) & & & &\\\midrule
ICU Admission & \textbf{0.79} & 0.75 & 0.78 &\textbf{0.79}\\
Mortality &\textbf{0.67} & 0.65 & \textbf{0.67} &\textbf{0.67}\\
Smoker &0.59 & \textbf{0.61} & 0.60 &\textbf{0.61}\\
Pregnancy stage &\textbf{0.78} & 0.73 & 0.72 &\textbf{0.78} \\
Hypertension &\textbf{0.77} & 0.72 & 0.75 &\textbf{0.77}\\
Sleep Disordered Breathing &0.62 & 0.53 & 0.64 &\textbf{0.70} \\
Mood Disturbance &0.53 & \textbf{0.56} & 0.55 &\textbf{0.56}\\
Valence &0.54 & 0.55 & 0.53 &\textbf{0.56}\\
Arousal &0.44 & 0.51 & 0.49 &\textbf{0.55}\\
\midrule
\textbf{Regression} - MAE ($\downarrow$)& & &  \\
\midrule
Apnea/Hypopnea Index $>$ 3\% &13.36 & 13.74 & 13.42 &\textbf{12.97}\\
Apnea/Hypopnea Index $>$ 4\% &11.01& 11.43 & 11.29 &\textbf{10.56} \\
Gestation Age &6.18 & 6.32 & 6.15 &\textbf{6.05}\\
Systolic BP (VV) &\textbf{14.62} & 15.97 & 15.33 &14.65\\
Diastolic BP (VV) &8.32 & 8.76 & 9.04 &\textbf{8.29}\\
Systolic BP (PPG-BP) &15.03 &\textbf{14.39} & 16.15 &\textbf{14.39}\\
Diastolic BP (PPG-BP) &9.12 & 8.76 & 9.06 &\textbf{8.71}\\
Average HR &\textbf{4.00} & 5.88 & 4.26 &\textbf{4.00}\\
HR &\textbf{11.51} & 11.97 & 11.88 &11.53\\
\bottomrule
\end{tabular}}
\end{table}

\subsection{Statistical Significance of Model Comparison}

In addition to confidence intervals, we perform the following steps to evaluate the significance across models on a per task basis (Tables \ref{tab:binary_classification_fm} \& \ref{tab:binary_classification_ssl}). First, we ran the Friedmann Chi Square test, and identified statistically significant differences across \model{} and the baseline models at $p < 0.05$. Next, we created critical difference (CD) diagrams to rank the best performing models, as suggested by the literature to compare models over multiple datasets\footnote{\url{https://scikit-posthocs.readthedocs.io/en/latest/tutorial.html\#critical-difference-diagrams}} \citep{demvsar2006statistical}. The CDs indicate that \model{} performs the best across both classification and regression tasks. Furthermore, it has a statistically significant average rank as indicated by the lack of horizontal line.

\begin{figure}
    \centering
    \includegraphics[width=1\linewidth]{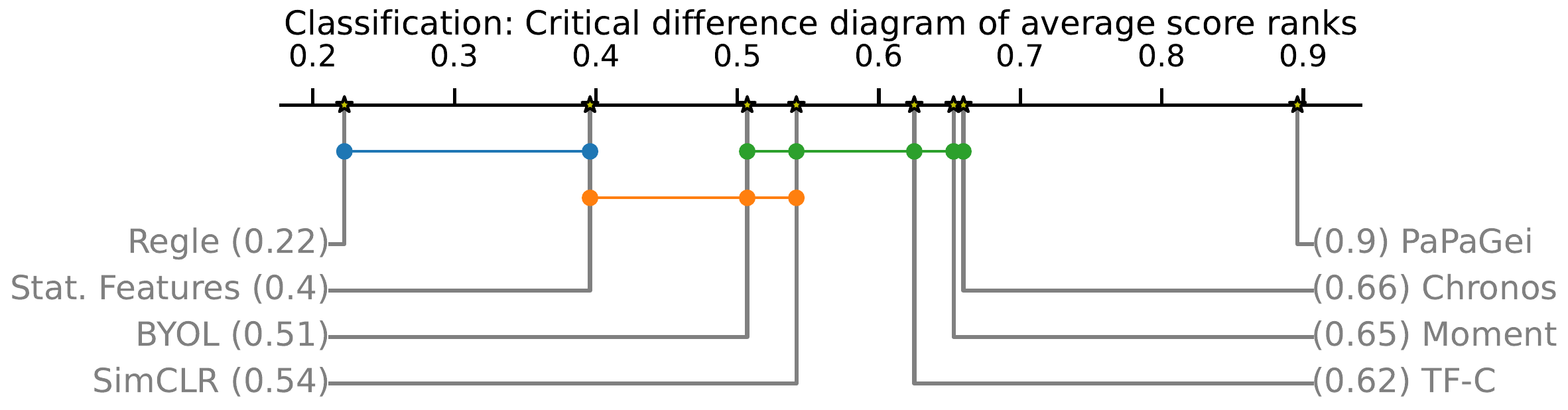}
    \caption{\textbf{Critical Difference Diagram for Classification Tasks}. The axis represents the average rank of the model. The horizontal connector lines indicate no significant differences between the models.}
    \label{fig:cd_classification}
\end{figure}

\begin{figure}
    \centering
    \includegraphics[width=1\linewidth]{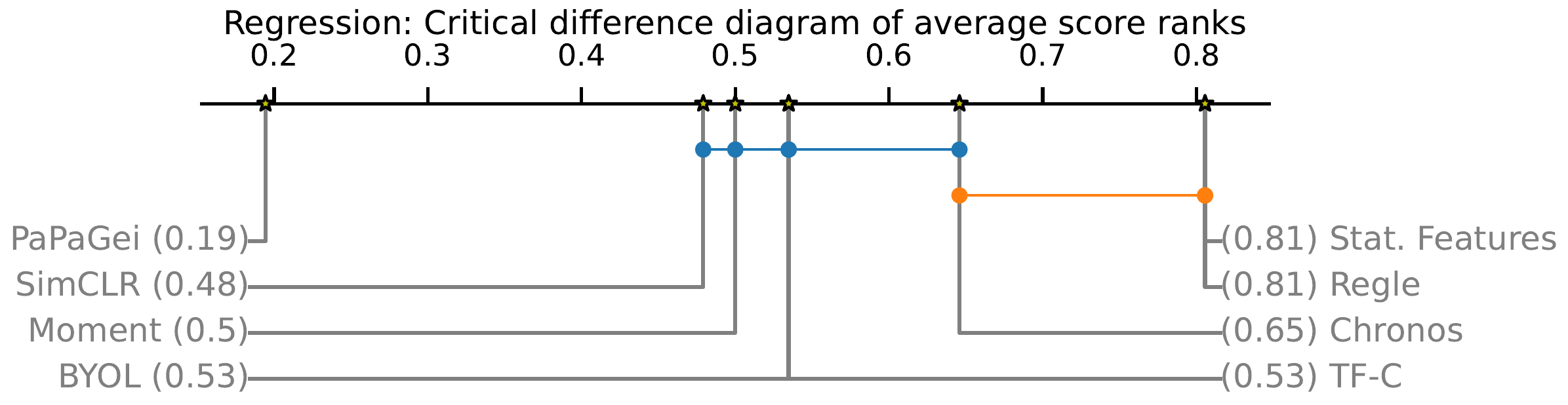}
    \caption{\textbf{Critical Difference Diagram for Regression Tasks}. The axis represents the average rank of the model. The horizontal connector lines indicate no significant differences between the models.}
    \label{fig:cd_regression}
\end{figure}

From the critical difference diagrams we observe that \model{} is significantly better across classification (Figure \ref{fig:cd_classification}) and regression (Figure \ref{fig:cd_regression}) tasks. This arises because \model{} is the highest ranking model across most tasks. Furthermore, we observe Moment is a strong model across both classification and regression tasks. Whereas Chronos and TF-C perform well for classification tasks only. 

\begin{figure}[h]
    \centering
    \begin{subfigure}[b]{0.45\textwidth}
         \includegraphics[width=\textwidth]{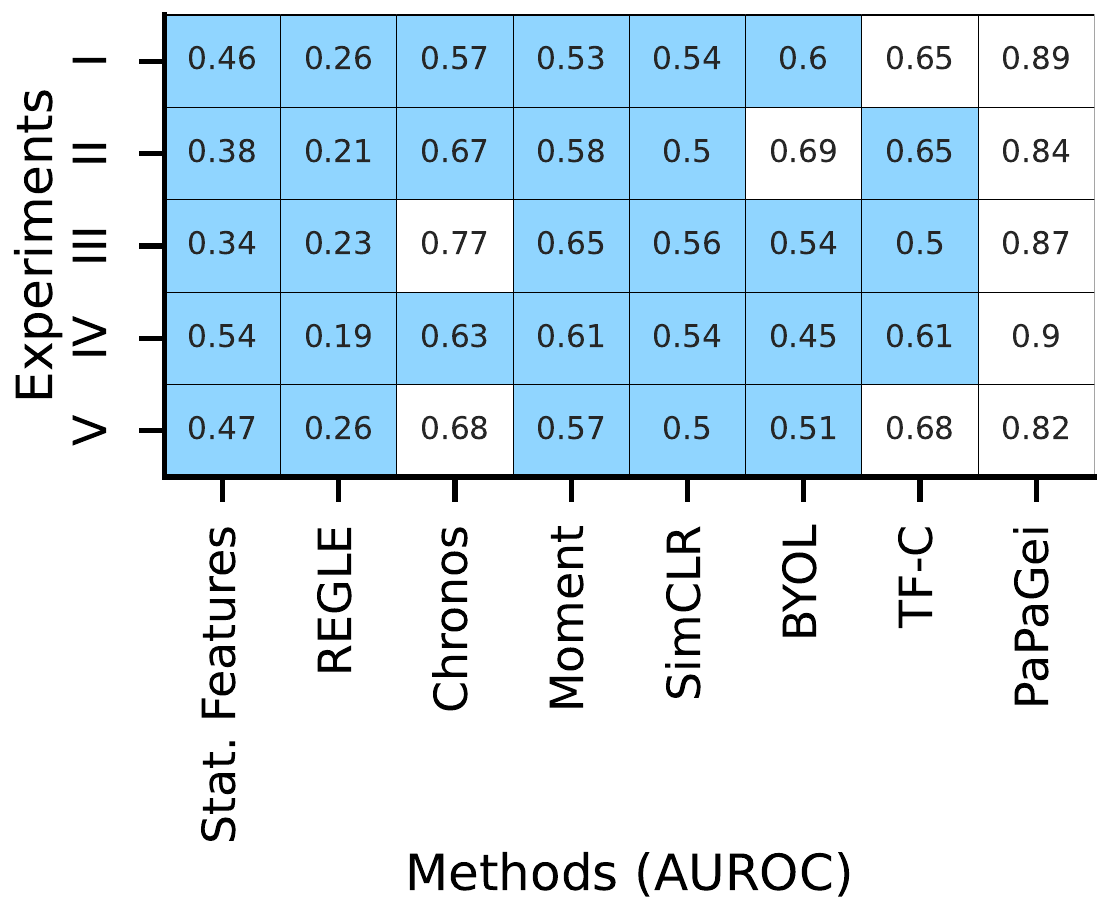}
     \end{subfigure}
     \hfill
    \begin{subfigure}[b]{0.45\textwidth}
         \includegraphics[width=\textwidth]{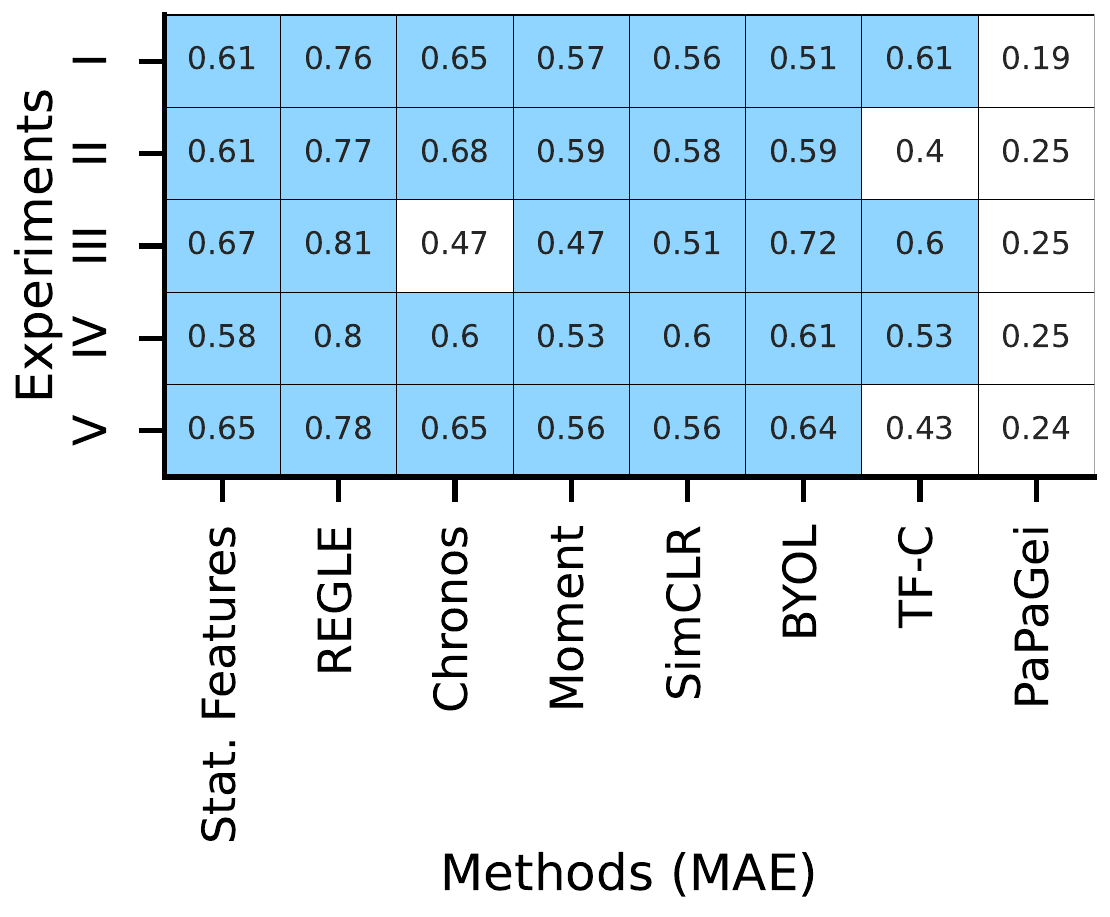}
     \end{subfigure} 
     \hfill 
    \caption{Bootstrap ranking repeated for five experiments: AUROC (left) and MAE (right). Colored cells indicate that \model{} is significant to the baseline at $p < 0.05$.}
    \label{fig:significance_heatmap}
\end{figure}

We conduct additional statistical significance comparisons using a structured approach. First, we randomly sample a score from within the confidence intervals for each task across all models. Next, we apply the CD ranking procedure to the sampled scores. This process is repeated 1,000 times, and the ranks are averaged. The entire experiment is conducted five times for Tables \ref{tab:binary_classification_fm} and \ref{tab:binary_classification_ssl}, with the results presented in Figure \ref{fig:significance_heatmap}. The colored cells indicate that \model{} is statistically significant compared to the respective model at $p < 0.05$. Our findings show that PaPaGei consistently achieves the best average rank, ranging between 0.82-0.90 for AUROC and 0.19-0.25 for MAE. Across 35 comparisons (PaPaGei vs. the other models, repeated five times), PaPaGei demonstrates significant improvements in 30 out of 35 AUROC comparisons and 32 out of 35 MAE comparisons. Among the baseline models, we acknowledge that Chronos and TF-C are strong competitors capable of performing comparably to \model{}.

\subsection{Statistical assessment between IPA and SQI} \label{appendix:statistics_ipa_sqi}
{Recall that we incorporate SQI to handle situations where the dicrotic notch cannot be computed due to poor-signal quality or different morphologies. We performed a permutation test \citep{rice2008sisg} to statistically evaluate PPG segments where IPA is unavailable. By splitting SQI values into no IPA and IPA groups and testing significance over 1000 permutations, we observed statistically significant differences ($p < 0.05$) in both mean (+0.18) and median (+0.32) SQI values, with the IPA group having larger SQI. These findings empirically motivate SQI's ability to handle limited PPG morphology.}

\section{Additional Baselines: Demographics \& PPG Morphology Features}
\label{appendix:demographics}
{In this section, we evaluate the effectiveness of demographics (Demo: age, sex) and PPG morphology (sVRI, IPA, SQI) to predict both regression (ridge) and classification (logistic regression) tasks: \textbf{(1) Ablation Study}: We compared PaPaGei with three baselines—demographics alone, PPG features alone, and demographics + PPG features. Our results show that while demographics is a stronger baseline than statistical features, PaPaGei outperforms the demographics + PPG baseline in 14 out of 18 tasks. \textbf{(2) Effect of Demographics}: We trained a model combining PaPaGei-S with demographics. The results indicate that incorporating demographic features with PaPaGei-S creates a stronger model than using PaPaGei-S alone.}

\begin{table}[h]
\centering
\caption{\textbf{Demographics \& PPG Morphology Baseline Results.}}
\label{tab:demo_analysis}
\scalebox{0.65}{
\begin{tabular}{@{}l|llll|ll@{}}
\toprule
 &\textbf{Stat. Features} &\textbf{Demo} &\textbf{PPG} &\textbf{Demo + PPG} &\textbf{\model{}-S or -P} &\textbf{\model{}-S + Demo} \\ 
\textbf{Classification} - AUROC ($\uparrow$) & & & & & & \\\midrule
ICU Admission & 0.71 \smat{[0.65-0.78]} & 0.64 \smat{[0.60-0.68]} & 0.59 \smat{[0.54-0.64]} & 0.66 \smat{[0.61-0.70]} & \textbf{0.79 \smat{[0.75-0.82]}} & 0.77 \smat{[0.74-0.81]} \\
Mortality & 0.57 \smat{[0.54-0.61]} & 0.66 \smat{[0.61-0.69]} & 0.57 \smat{[0.53-0.61]} & 0.66 \smat{[0.61-0.69]} & 0.67 \smat{[0.63-0.70]} & \textbf{0.70 \smat{[0.67-0.74]}} \\
Smoker & 0.63 \smat{[0.58-0.67]} & 0.64 \smat{[0.59-0.70]} & 0.56 \smat{[0.52-0.62]} & \textbf{0.66 \smat{[0.61-0.71]}} & 0.64 \smat{[0.58-0.69]} & 0.62 \smat{[0.56-0.68]} \\
Pregnancy stage & 0.64 \smat{[0.62-0.67]} & 0.52 \smat{[0.50-0.55]} & 0.55 \smat{[0.53-0.57]} & 0.56 \smat{[0.53-0.58]} & 0.78 \smat{[0.75-0.80]} & \textbf{0.78 \smat{[0.76-0.80]}} \\
Hypertension & 0.66 \smat{[0.47-0.83]} & 0.77 \smat{[0.65-0.88]} & 0.53 \smat{[0.40-0.68]} & 0.77 \smat{[0.65-0.88]} & 0.77 \smat{[0.68-0.87]} & \textbf{0.80 \smat{[0.70-0.89]}} \\
SDB &0.32 \smat{[0.14-0.55]} &-- &0.46 \smat{[0.31-0.62]} &-- &0.70 \smat{[0.57-0.84]} &-- \\
Mood Disturbance & 0.54 \smat{[0.31-0.77]} & 0.54 \smat{[0.30-0.80]} & \textbf{0.64 \smat{[0.42-0.85]}} & 0.63 \smat{[0.36-0.87]} & 0.56 \smat{[0.33-0.77]} & 0.52 \smat{[0.25-0.78]} \\
Valence & 0.52 \smat{[0.49-0.55]} & 0.57 \smat{[0.54-0.60]} & 0.44 \smat{[0.41-0.47]} & \textbf{0.57 \smat{[0.54-0.60]}} & 0.56 \smat{[0.54-0.59]} & 0.55 \smat{[0.53-0.58]} \\
Arousal & 0.55 \smat{[0.53-0.58]} & 0.54 \smat{[0.52-0.58]} & 0.51 \smat{[0.48-0.54]} & 0.54 \smat{[0.52-0.58]} & 0.58 \smat{[0.55-0.61]} & \textbf{0.58 \smat{[0.54-0.59]}} \\
\midrule
\textbf{Regression} - MAE ($\downarrow$)& & &  \\
\midrule
Apnea/Hypopnea Index $>$ 3\% & 15.31 \smat{[13.63-17.14]} &14.53 \smat{[13.29-15.84]} &15.09 \smat{[14.01-16.54]} &14.40 \smat{[13.12-15.61]} &12.97 \smat{[11.87-14.05]} &\textbf{12.35 \smat{[11.27-13.46]}} \\
Apnea/Hypopnea Index $>$ 4\% & 12.52 \smat{[10.92-14.14]} &12.28 \smat{[11.19-13.39]} &12.65 \smat{[11.57-13.83]} &12.17 \smat{[11.10-13.39]} &10.56 \smat{[9.59-11.62]} &1\textbf{0.47 \smat{[9.53-11.50]}} \\
Gestation Age & 7.15 \smat{[6.99-7.34]} &7.69 \smat{[7.61-7.77]} &7.61 \smat{[7.51-7.70]} &7.59 \smat{[7.51-7.68]} &6.05 \smat{[5.91-6.17]} &\textbf{6.02 \smat{[5.88-6.17]}} \\
Systolic BP (VV) & 15.76 \smat{[13.67-18.36]} &14.96 \smat{[13.21-17.35]} &15.82 \smat{[13.48-18.31]} &15.01 \smat{[13.30-17.86]} &14.65 \smat{[12.50-16.78]} &\textbf{14.27 \smat{[11.92-16.44]}} \\
Diastolic BP (VV) & 9.75 \smat{[7.16-11.27]} &8.75 \smat{[6.48-9.77]} &9.20 \smat{[7.21-10.71]} &8.78 \smat{[7.10-10.25]} &8.29 \smat{[6.61-10.22]} &\textbf{8.26 \smat{[6.64-10.16]} }\\
Systolic BP (PPG-BP) & 15.50 \smat{[11.68-20.25]} &13.71 \smat{[11.33-15.95]} &15.76 \smat{[13.36-18.30]} &13.74 \smat{[11.37-16.09]} &13.60 \smat{[10.65-16.51]} &\textbf{13.20 \smat{[11.47-15.66]}} \\
Diastolic BP (PPG-BP) & 9.35 \smat{[7.44-11.66]} &9.26 \smat{[7.89-10.68]} &9.36 \smat{[7.95-10.92]} &9.28 \smat{[8.00-10.56]} &8.71 \smat{[7.18-10.01]} &\textbf{8.61 \smat{[7.34-9.88]}} \\
Average HR & 7.01 \smat{[5.48-8.89]} &9.12 \smat{[7.86-10.61]} &8.07 \smat{[6.60-9.71]} &8.23 \smat{[6.82-9.78]} &\textbf{3.47 \smat{[2.74-4.32]}} &4.00 \smat{[3.35-4.73]} \\
HR & 13.07 \smat{[12.90-13.23]} &15.18 \smat{[15.03-15.33]} &16.75 \smat{[16.60-16.90]} &14.46 \smat{[14.32-14.62]} &\textbf{10.92 \smat{[10.80-11.04]}} &12.38 \smat{[11.90-12.96]} \\
\bottomrule
\end{tabular}}
\end{table}

{From Table \ref{tab:demo_analysis}, we observe the following classification performance (Positive is better): ICU (+0.13), Mortality (+0.01), Smoker (-0.02), Pregnancy Stage (+0.22), Hypertension (0.00), SDB (no demographics), Mood Disturbance (-0.08), Valence (-0.01), Arousal (+0.04). Regression Tasks (Negative is better): AHI $>$ 3\% (-1.43), AHI $>$ 4\% (-1.61), gestation age (-1.54), SBP-VV (-0.31), DBP-VV (-0.46), SBP (-0.11) , DBP (-0.65), Avg. HR (-4.07), HR (-2.93). PaPaGei-S performs better for real-time sleep and cardiovascular outcomes such as sleep apnea, heart rate and blood pressure, respectively. In particular, we notice that outcomes such as heart rate benefit substantially from PPG rather than demographics. Demographics are useful in tasks without real-time dependence such as smoking, which is established to be associated with age and sex \citep{chung2020demographics}. Importantly, demographics do not add much to already homogeneous populations. For example, consider the NuMoM2B dataset which has women within a specific age range. Here, we observe that PaPaGei obtains much higher AUROC and MAE than the supervised baselines.}

{Furthermore, We observe that adding demographics to PaPaGei-S embeddings improves over PaPaGei in the following tasks: Mortality (+0.03), Hypertension (+0.03), AHI $>$ 3\% (-0.62), AHI $>$ 4\% (-0.09), gestation age (-0.03), SBP VV (-0.38), DBP VV (-0.03), SBP (0.40), DBP (0.10). Based on these results, PaPaGei-S + Demo is a stronger model in many cases. Importantly, these results indicate that PaPaGei-S embeddings learn features that are complementary to demographics are not simply proxies for age or sex. However, it is important to note that while demographic features can be valuable for personalization, they may not always be readily available, and in reality, we cannot use them in isolation to predict real-time outcomes such as blood pressure or heart rate. Therefore, our PaPaGei models are designed to function effectively with real-time sensor data alone, ensuring their applicability in situations where complete demographic information is not accessible.}

{These findings underscore an important point: \textbf{demographic features are not competing with PaPaGei but rather complement it}, as previously established in studies including demographics with sensor data \citep{spathis2022longitudinal}. This highlights the synergistic potential of combining PaPaGei's advanced feature extraction with demographic context for improved task performance.}

{\textbf{Predicting Demographics Targets.} Using the \model{} features, we predict downstream demographics such as age and sex (Table \ref{tab:demographic_targets}). \model{}-S achieves 7.78 MAE in age regression, 0.85 accuracy in age classification, and 0.79 accuracy in sex classification. Although our results trail larger closed studies \citep{abbaspourazad2023large} by 2.18, 0.05, and 0.13 for segment-level SSL, and by 5.59, 0.12, and 0.25 for patient-level SSL, they mark an advancement in open-source efforts. The superior performance of \cite{abbaspourazad2023large} can be attributed to two primary factors. First, the patient-level positive pair strategy achieves the best performance across all tasks. This approach encourages the model to form distinct clusters for each patient, effectively capturing demographic factors such as age and sex. In contrast, a segment-level approach pushes the model to cluster similar segments across individuals with varying demographics, potentially mixing demographic-specific information. Second, the single-device setup with a larger dataset is useful for effective model training (Table \ref{tab:studies}). Conversely, our evaluation, which spans three devices and utilizes smaller datasets, must handle greater data heterogeneity, thus making it more challenging.}

\begin{table}[h]
\centering
\caption{\textbf{Predicting personal characteristics with embeddings}. Downstream prediction on age regression, age classification, and sex classification in our pre-training datasets (VitalDB, MESA, MIMIC-III). The regression and classification tasks are reported using MAE and AUROC, respectively. Note: training and testing are conducted with completely different cohorts in the two studies, hence comparisons are difficult.}
\scalebox{0.9}{
\begin{tabular}{@{}llll@{}}
\toprule
Study & Age Regression ($\downarrow$) & Age Classification ($\uparrow$) & Sex Classification ($\uparrow$) \\ \midrule
\cite{abbaspourazad2023large}(Patient) &3.19 &0.97 &0.99 \\
\cite{abbaspourazad2023large}(Segment) &6.60                &0.90         &0.87 \\  
\model{}-S (Ours) & 8.78 \smat{[8.47-8.09]} &0.85 \smat{[0.83-0.87]} & 0.74 \smat{[0.72-0.76]} \\
 \bottomrule
\end{tabular}}
\label{tab:demographic_targets}
\end{table}

\section{Additional Prediction Plots}
\label{appendix:prediction_plots}
\begin{figure}
  \centering
  \subcaptionbox{AHI $>$ 4\% predictions}%
    {\includegraphics[width=0.20\textwidth]{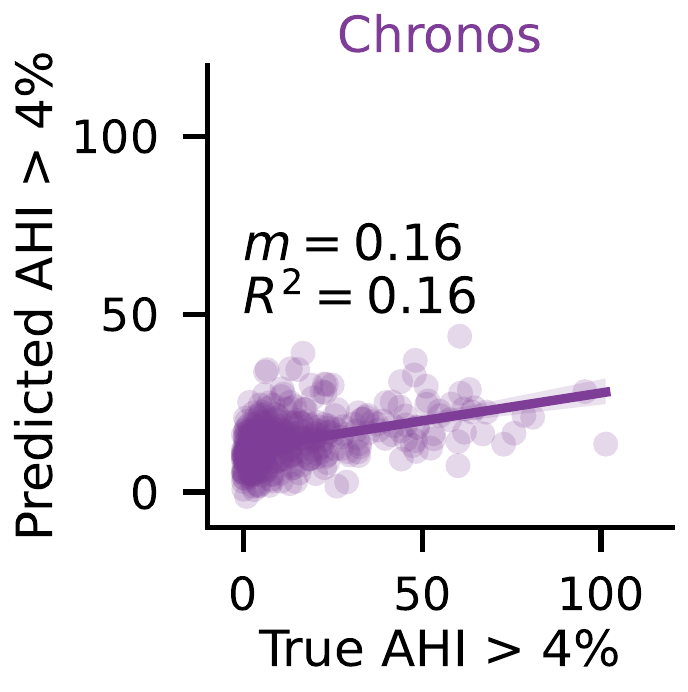}\includegraphics[width=0.18\textwidth]{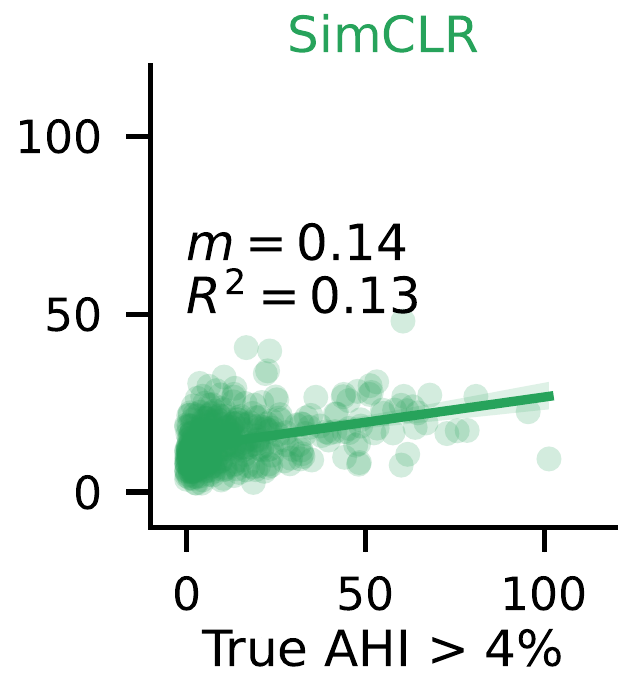} \includegraphics[width=0.18\textwidth]{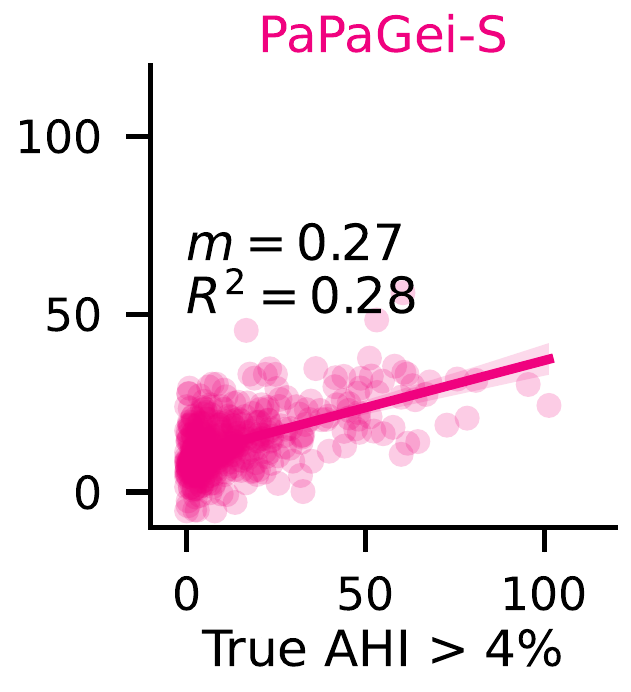}
    \includegraphics[width=0.40\textwidth]{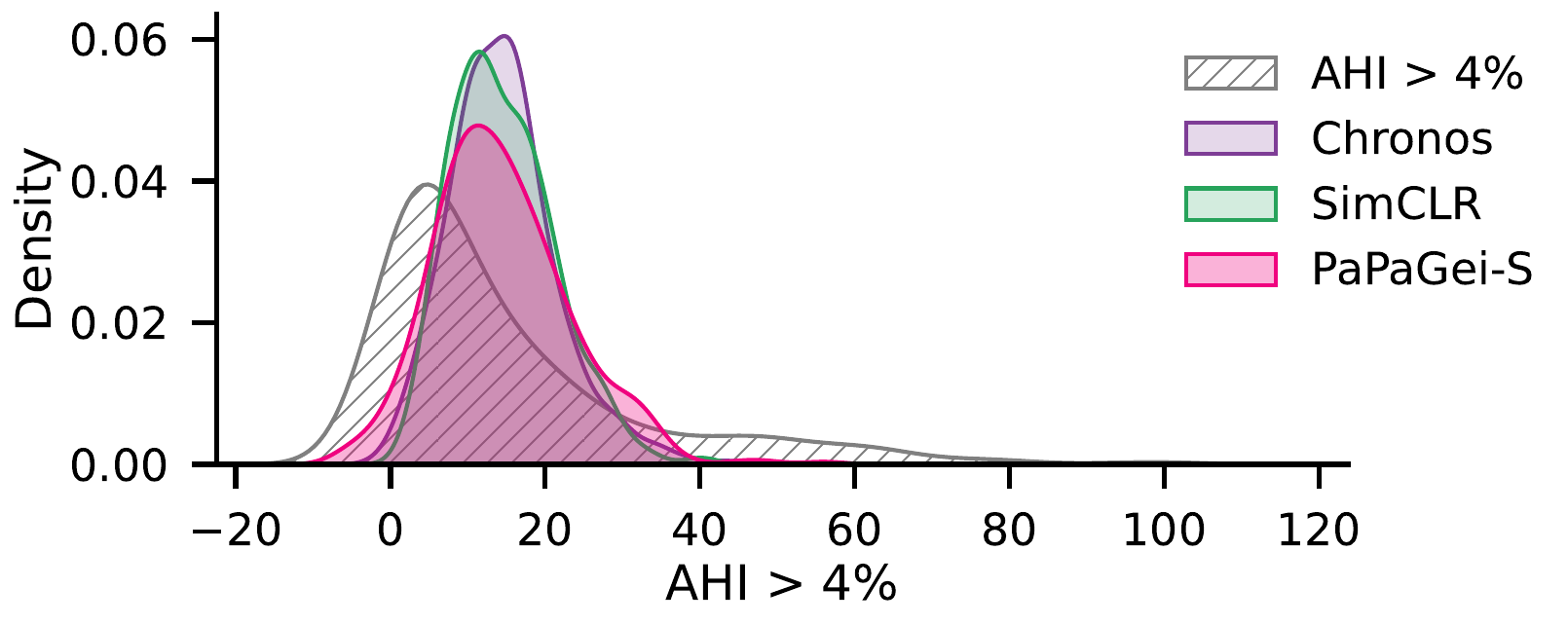}} \label{fig:predictions_ahi}%
  \subcaptionbox{Avg. HR predictions}%
    {\includegraphics[width=0.20\textwidth]{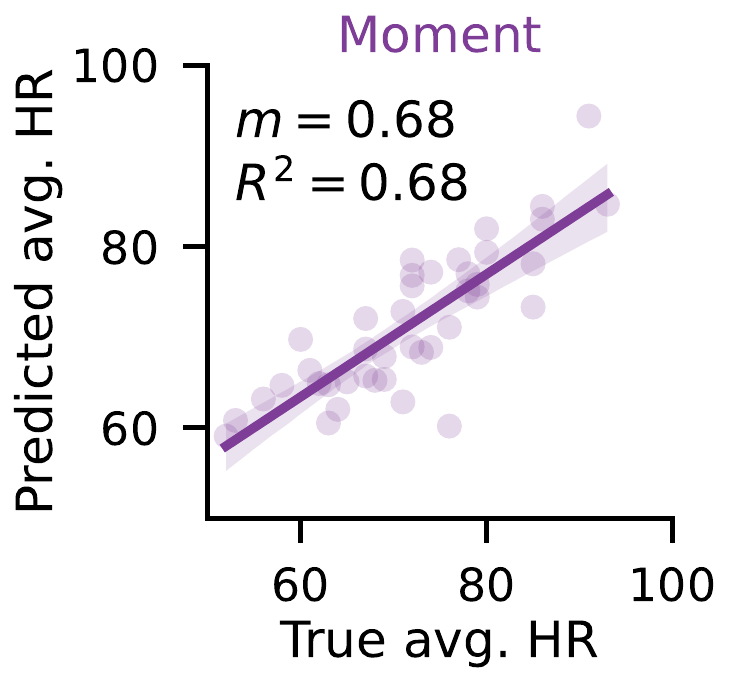}\includegraphics[width=0.18\textwidth]{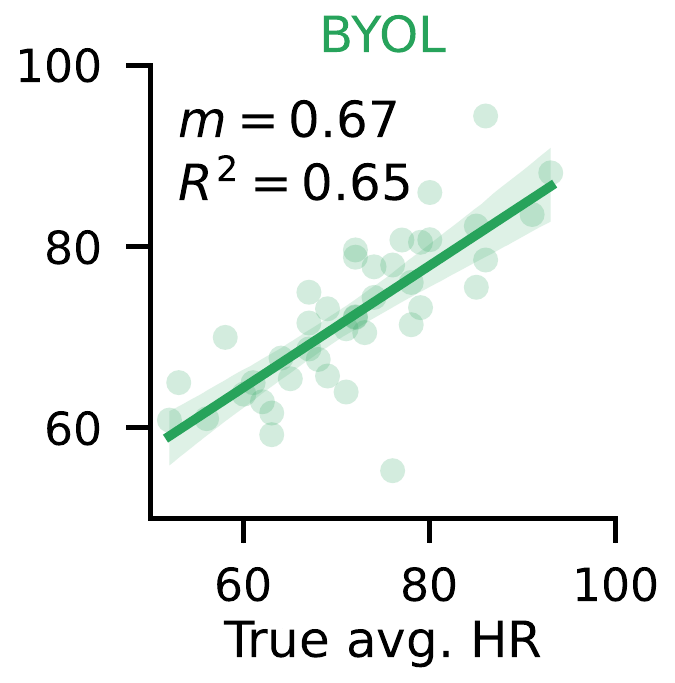}
    \includegraphics[width=0.18\textwidth]{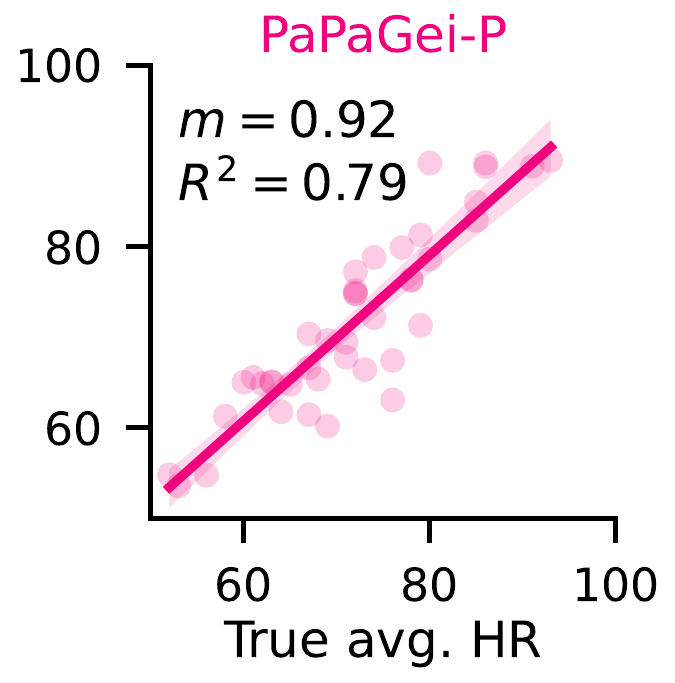}
    \includegraphics[width=0.40\textwidth]{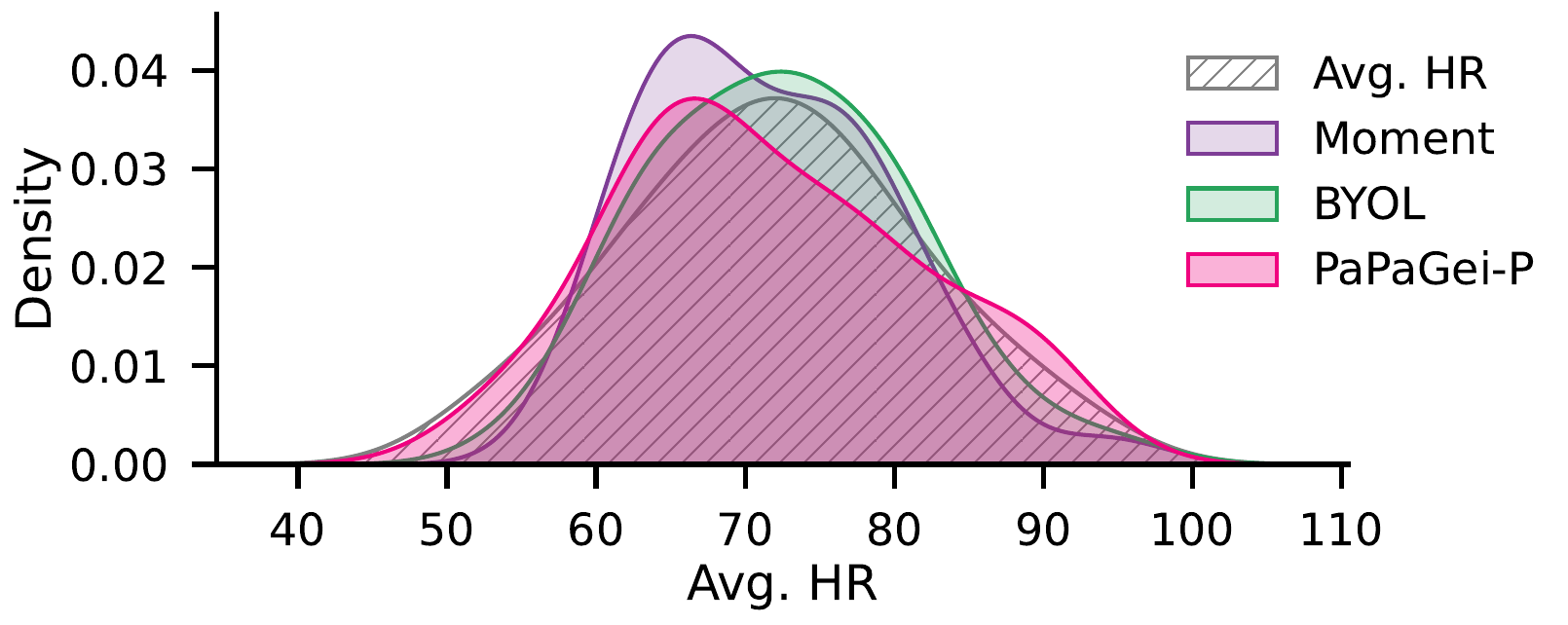}} \label{fig:predictions_hr}%
   \subcaptionbox{Gestation Age}%
    {\includegraphics[width=0.17\textwidth]{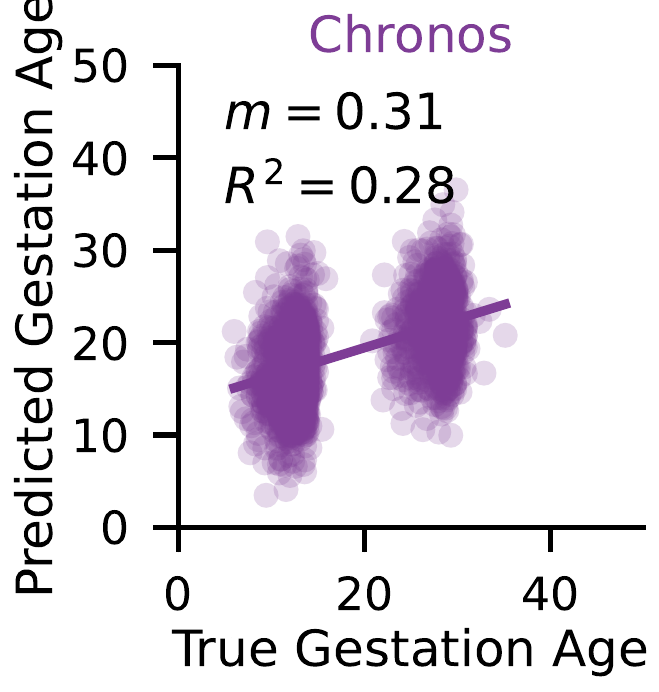}
    \includegraphics[width=0.17\textwidth]{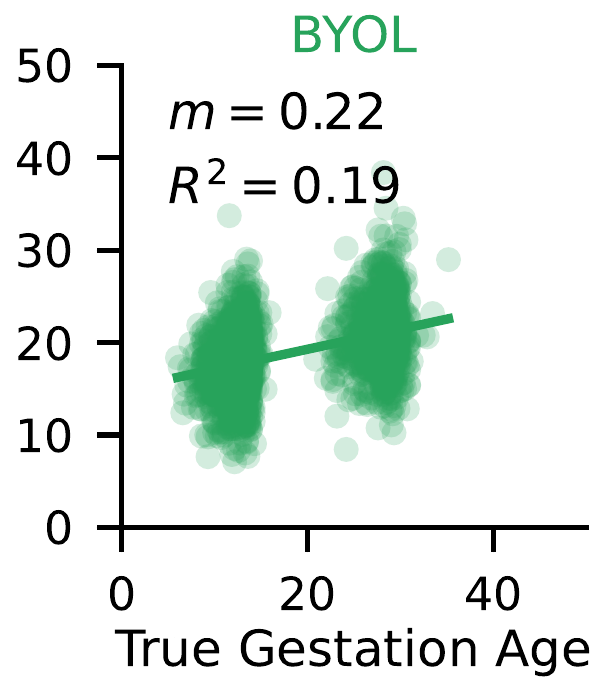}
    \includegraphics[width=0.17\textwidth]{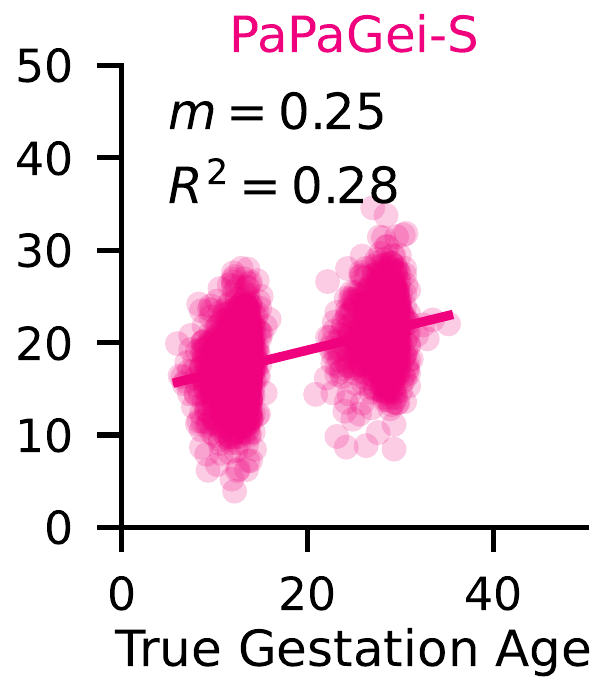}
    \includegraphics[width=0.48\textwidth]{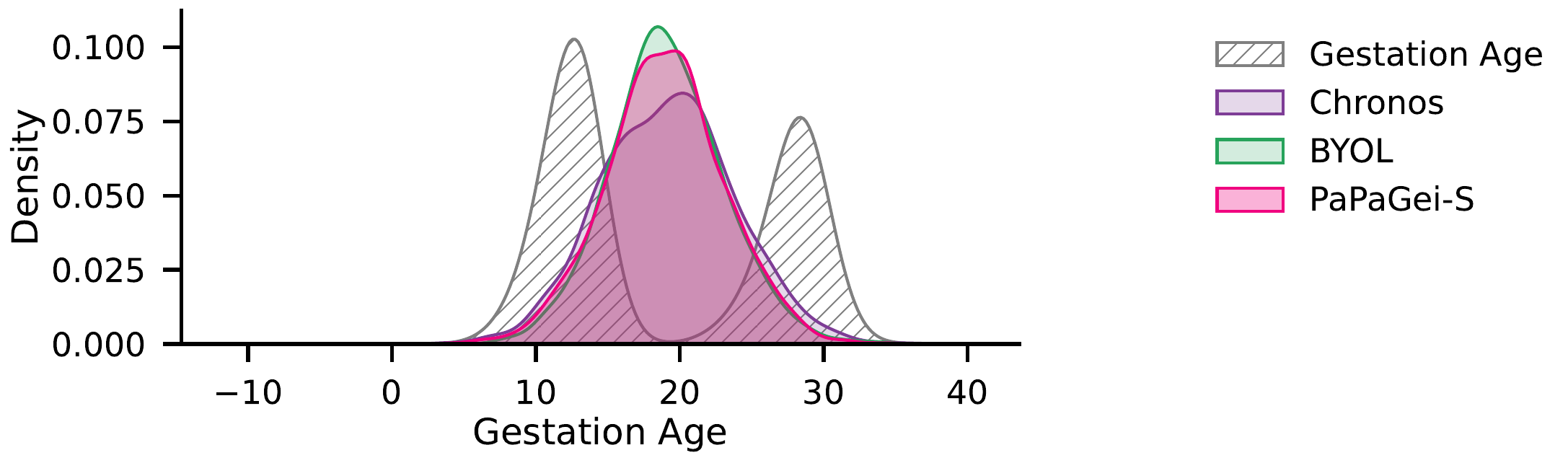}} \label{fig:predictions_ga}%
\subcaptionbox{Systolic BP (PPG-BP)}%
    {\includegraphics[width=0.20\textwidth]{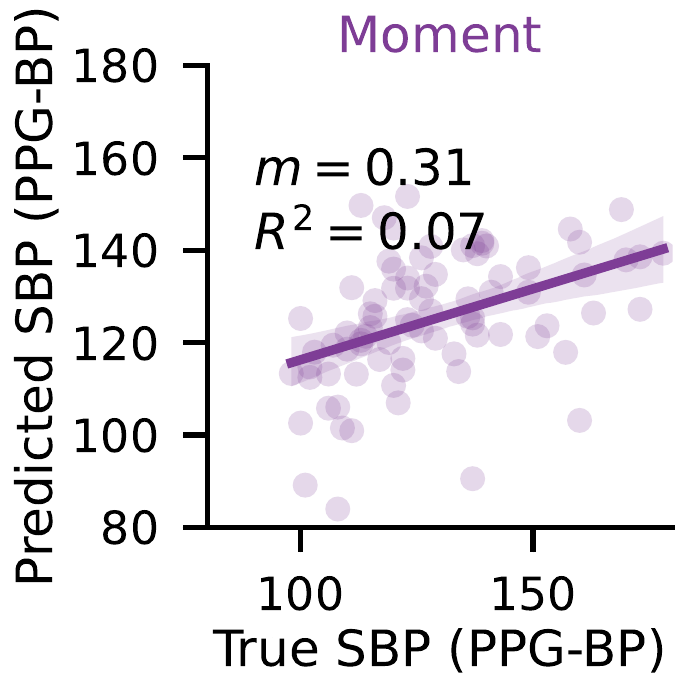}
    \includegraphics[width=0.18\textwidth]{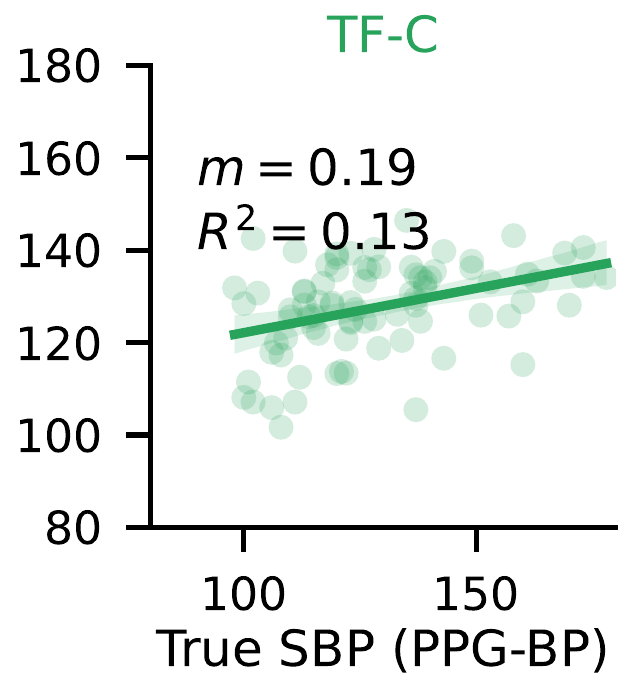}
    \includegraphics[width=0.18\textwidth]{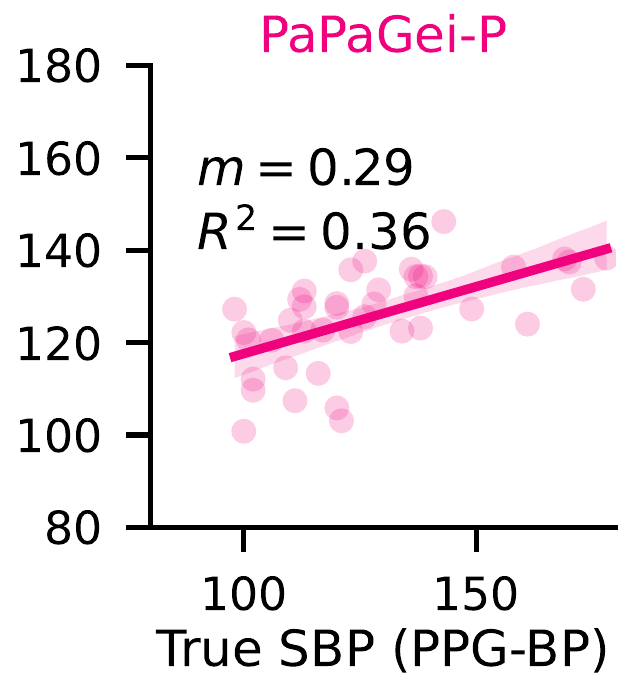}
    \includegraphics[width=0.40\textwidth]{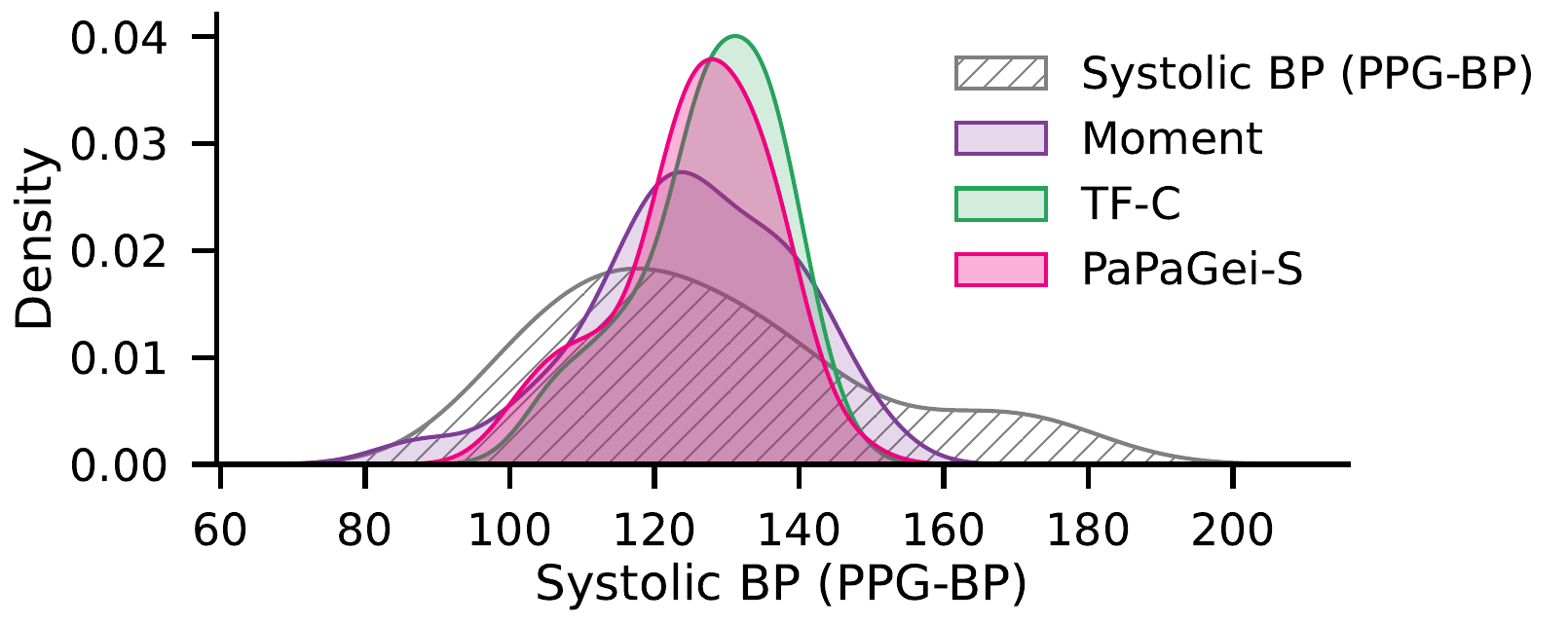}} \label{fig:predictions_sbp}%
\subcaptionbox{Diastolic BP (PPG-BP)}%
    {\includegraphics[width=0.20\textwidth]{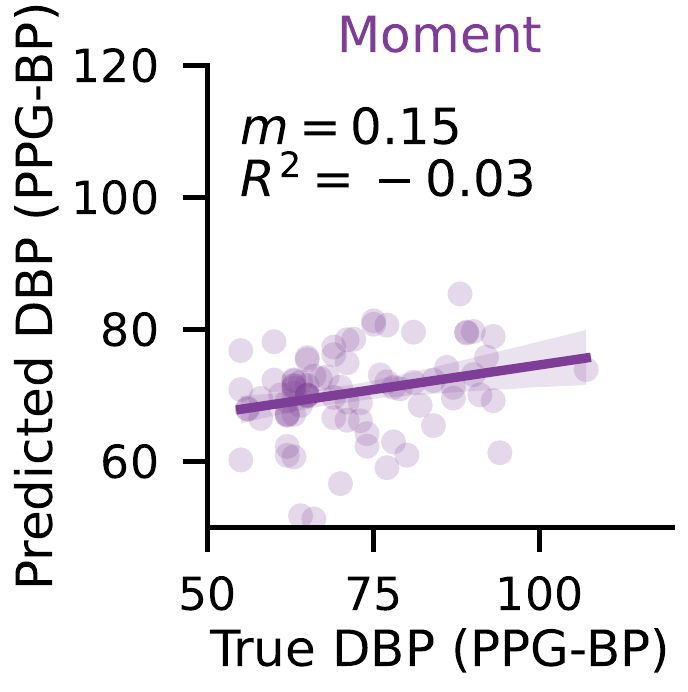}
    \includegraphics[width=0.18\textwidth]{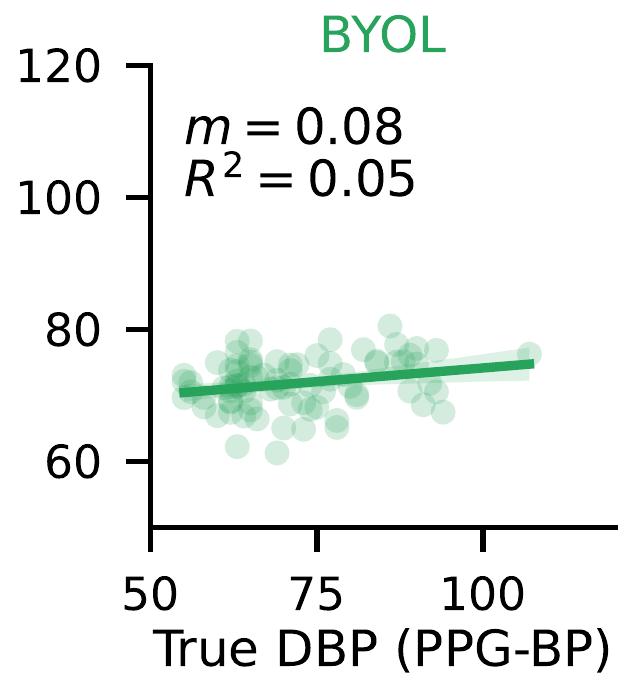}
    \includegraphics[width=0.18\textwidth]{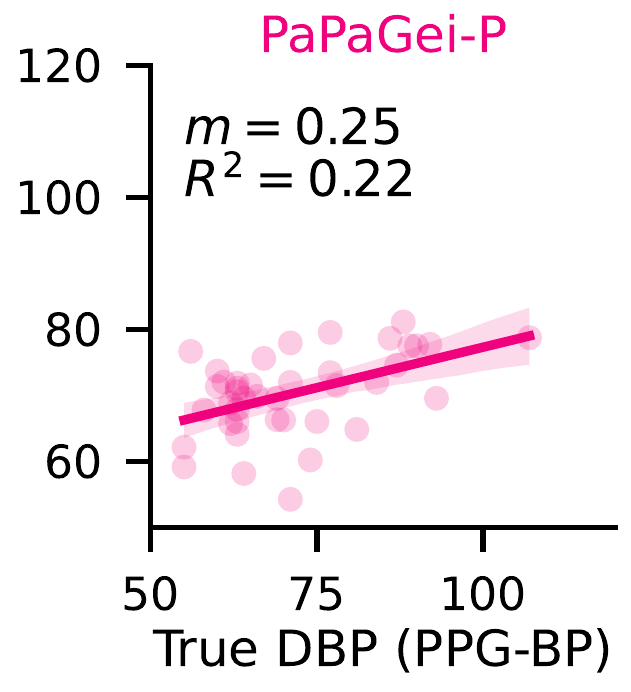}
    \includegraphics[width=0.40\textwidth]{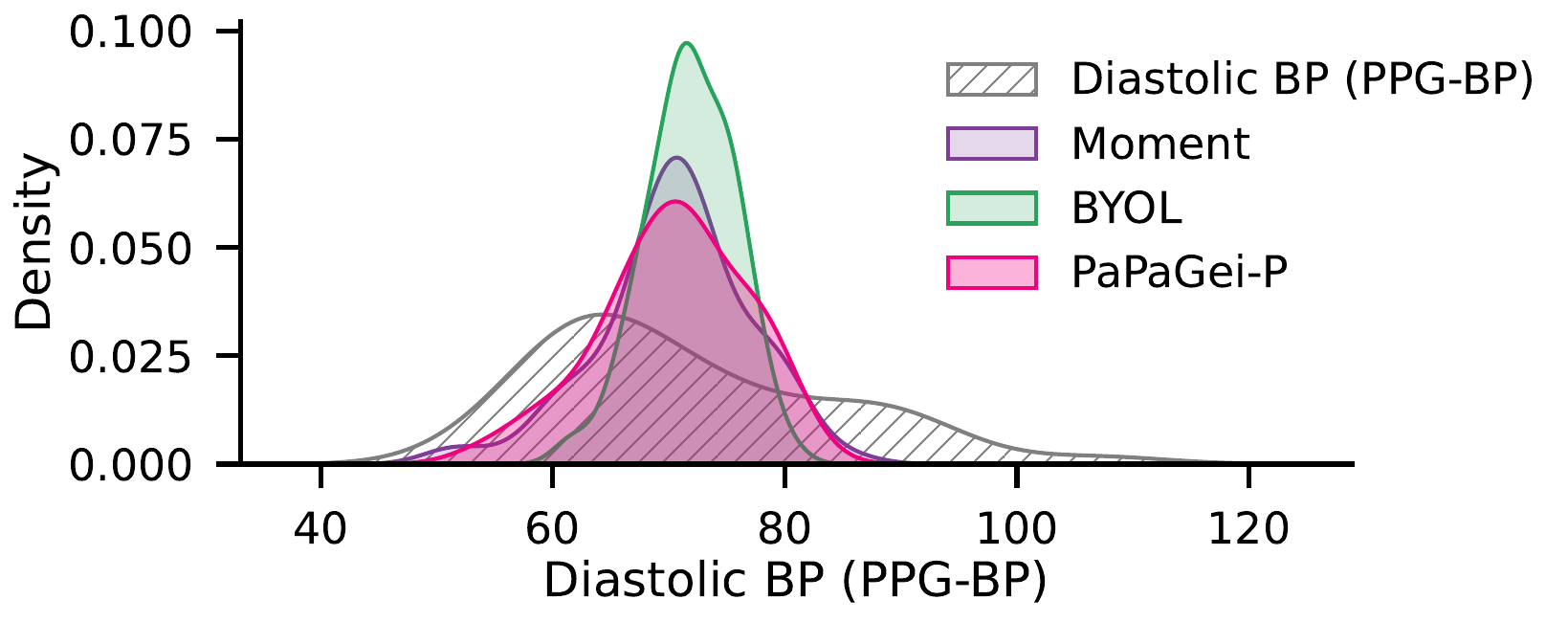}} \label{fig:predictions_dbp}%
  \caption{{Regression plots and prediction distribution of different models compared to ground truth for (a) Apnea/Hypopnea Index $>$ 4\%, (b) Average Heart Rate, (c) Gestation Age, (d) Systolic BP (PPG-BP), and (e) Diastolic BP (PPG-BP). $R^2$ is the coefficient of determination and $m$ is the correlation slope.}} 
  \label{fig:predictions_additional}
\end{figure}

{The regression plots to evaluate the agreement between true and predicted values in shown in Figure \ref{fig:predictions_additional}. From the Figure, we observe that \model{}'s predictions are more aligned to the true values for AHI $>$ 3\% ($R^2 = 0.28$), Avg. HR ($R^2 = 0.79$), gestation age ($R^2 = 0.28$), SBP ($R^2 = 0.36$), and DBP ($R^2 = 0.22$). Moreover, from the distribution plots in Figure \ref{fig:predictions_additional}, we notice that \model{} has stronger overlap for AHI $>$ 4\%, Avg. HR and DBP, indicating its ability to capture the tails for these tasks. Interestingly, we notice that all models are unable to capture the bi-modal nature of the gestation age measurements. Here, Chronos performs better than other methods to capture readings from the first visit.} 

\section{Effect of Skin Tone}
\begin{figure}[]
    \centering
    \begin{subfigure}[b]{1\textwidth}
         \includegraphics[width=\textwidth]{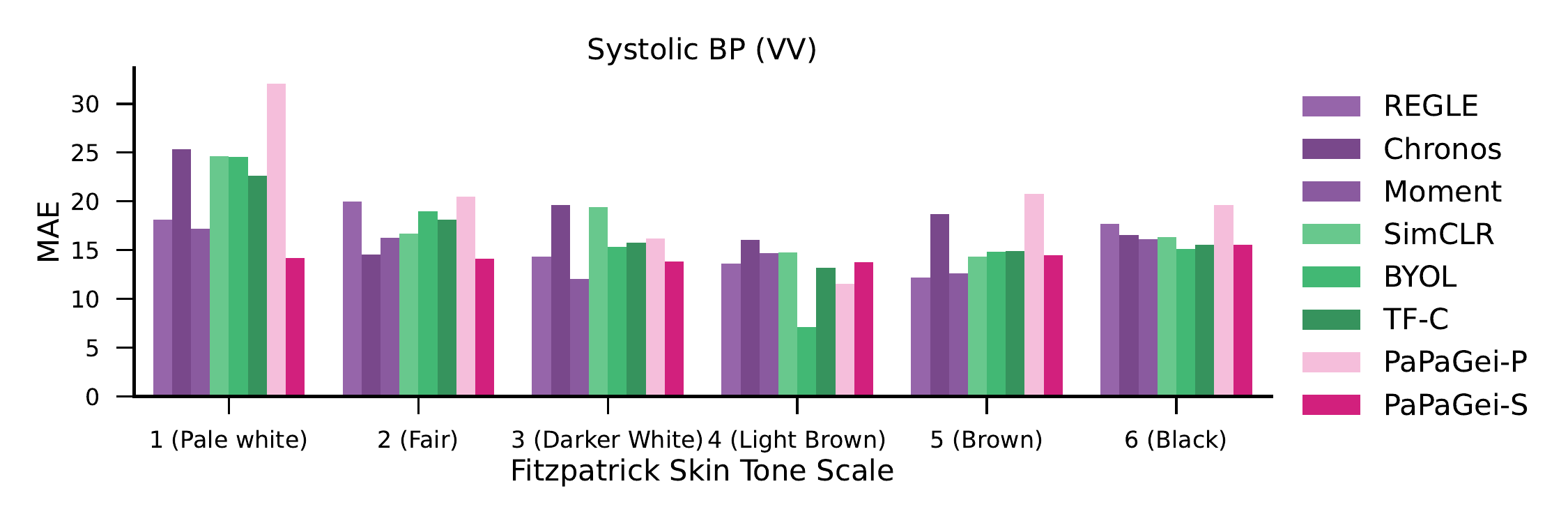}
     \end{subfigure}
     \begin{subfigure}[b]{1\textwidth}
         \includegraphics[width=\textwidth]{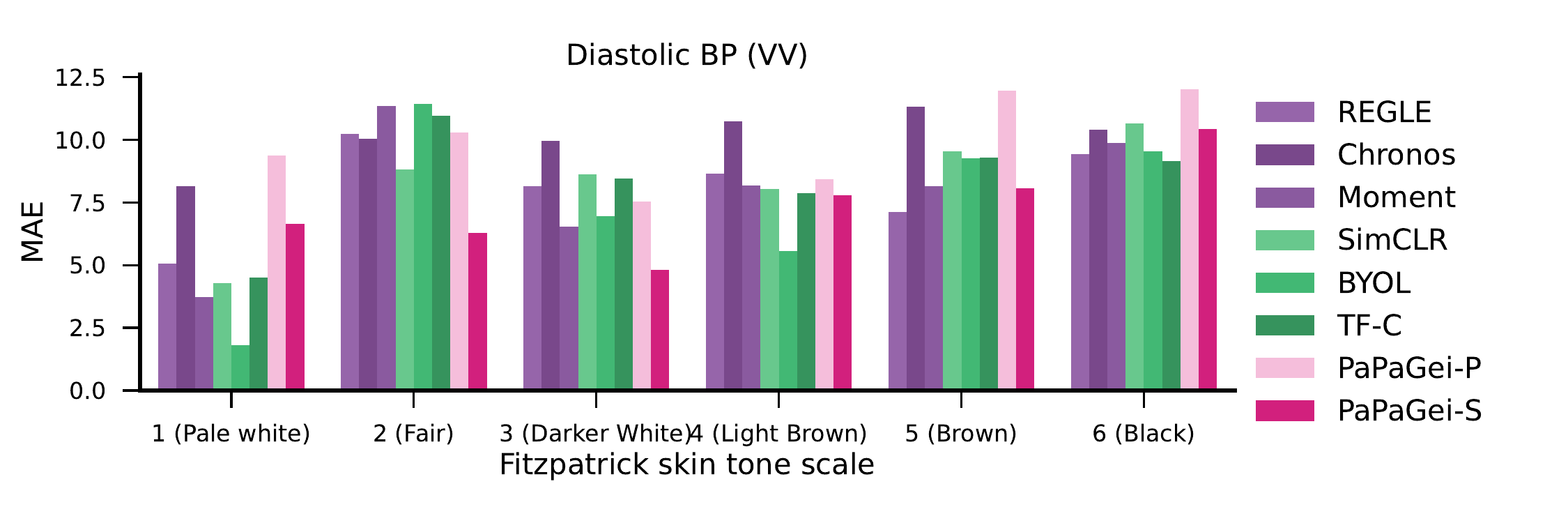}
     \end{subfigure}
    \caption{Detailed skin tone analysis for Blood Pressure estimation (VV dataset).}
    \label{fig:fitzpatrick_detailed}
    \vspace{-0.5cm}
\end{figure}

We present the skin tone analysis in a more granular way in Figure \ref{fig:fitzpatrick_detailed}. Here, \model{}-S clearly performs better than \model{}-P in most cases. Overall, we notice that \model{}-S is good for lighter skin tones in the 1-2 range for SBP and 2-3 range for DBP. While \model{}-S does not perform the best for darker skin tones, it's performance is comparable to other models for skin tone ratings of 4 and 5. Overall, these results indicate that \model{}-S is relatively robust to skin tone variations, and that additional future work is needed to make it better darker skin tones. 

\section{Extended related work}\label{appendix:extended_related_work}

Self-supervised learning (SSL) is the most prominent paradigm for learning general representations from large unlabeled datasets, including methods like SimCLR \citep{chen2020simple}, BYOL \citep{grill2020bootstrap}, and masked autoencoders (MAE) \citep{he2022masked}. Timeseries-specific objectives like TNC and TF-C have also shown promise \citep{tonekaboni2021unsupervised, zhang2022self}. SSL has gained traction in the domain of physiological signal analysis, with applications to health records \citep{chen2021forecasting, yeche2021neighborhood}, fitness and personalization \citep{spathis2021self}, as well as brain \citep{cheng2020subject} and heart signals \citep{kiyasseh2021clocs, sarkar2020self}. 

\begin{table}[]
\caption{\textbf{Comparison of large-scale PPG studies.} * indicates partial availability. The participants and hours indicate pre-training data.}
\centering
\scalebox{0.68}{
\begin{tabular}{@{}ccccccc@{}}
\toprule
Study & \#Participants (\#Hours) & \#Devices (Types) & Open Data & Open Weights & Open Code & \#Tasks (\#Datasets) \\ \midrule
\cite{abbaspourazad2023large} &141,207 (333K)  & \makecell{1 (Smartwatch)} &\xmark  &\xmark &\xmark  & $>$46 (1) \\
\cite{ding2024siamquality} &28,539 (300K) &\makecell{5-6 (ICU, Smartwatch)} &\xmarkpartial &\xmark &\checkmark & 4 (7)  \\
\cite{yun2024unsupervised} &170,714 (Varying)\tablefootnote{\url{https://biobank.ndph.ox.ac.uk/crystal/field.cgi?id=4205}} & \makecell{1 (Finger)} &\xmark &\checkmark &\checkmark* & 2 (4) \\ 
\midrule
\model{} (Ours) &13,517 (57K) &\makecell{7 (ICU, Smartwatch,\\ Finger, "Phone Ox.")} &\checkmark &\checkmark &\checkmark &20 (10)  \\ \bottomrule
\end{tabular}
}
\label{tab:studies}
\end{table}


However, despite the popularity of SSL, there are no widely used FMs for PPG data. While \citep{abbaspourazad2023large} showcased the potential of foundation models for physiological signals, it was based on a single proprietary dataset and device (Apple Watch) while the models were not released, limiting its practical use in the research community. Similarly, REGLE's work \citep{yun2024unsupervised} on the UK Biobank dataset showed that embedding PPG signals can improve genetic discovery and risk prediction outcomes. Although parts of that model and pipeline are public, the dataset is not openly accessible, and the primary goal was not to create a foundation model for PPG but rather to focus on genetics. Another work on the same data showed that PPG embeddings are promising for cardiovascular risk prediction \citep{weng2024predicting}. SiamQuality \citep{ding2024siamquality} also trained an unreleased model on 36 million PPG signals using proprietary data. Importantly, most of these works pre-trained on a single-device dataset and did not explore out-of-domain datasets or conduct transfer learning experiments, which are crucial for assessing the true generalizability of foundation models. These studies highlight the potential of PPG-based foundation models but also underscore the need for openly available, pre-trained models that can be widely used and adapted by the research community. 

On the other hand, generic time series foundation models have begun to gain popularity, mirroring the trend seen in Large Language Models (LLMs). These models are pre-trained on massive corpora of diverse time series data, aiming to learn universal representations that can be applied across various domains. For instance, Chronos \citep{ansari2024chronos} was trained on an impressive 84 billion observations (analogous to tokens in NLP) from 28 distinct datasets. However, it's notable that this diverse collection does not include physiological data. Similarly, Moment \citep{goswami2024moment} was trained on billions of observations from a wide-ranging dataset that includes weather, traffic, energy, and other domains. While Moment does incorporate a small amount of ECG data, it comprises only a tiny percentage of the overall data pool.


In contrast to these generic approaches, our work takes a domain-specific focus. We curate a large pre-training and evaluation benchmark dedicated exclusively to PPG data. While knowledge gained from generic time series foundation models may transfer to domain-specific tasks like PPG, we expect the performance to be limited compared to a model trained specifically on PPG data. Furthermore, foundation models for ECG \citep{mckeen2024ecg,song2024foundation} or EEG \citep{yuan2024brant} have shown promise but transferring from one domain-specific model (e.g., ECG) to another (PPG) is likely to be even more challenging, as the underlying signal characteristics can be quite different. {For instance, \citet{lai2023practical} trained a large-scale 12-lead ECG model for detecting 60 diagnostic terms, while \citet{mckeen2024ecg} developed an open-source ECG FM using 1.6 million 12-lead signals. In brain signal analysis, \citet{yuan2024brant} introduced Brant-2, an EEG and SEEG model supporting tasks like sleep staging and seizure detection. Building on this progress, we adopt a domain-specific approach focused on photoplethysmography (PPG) signals.} Building on the increasing interest in modality-specific foundation models, our specialized approach allows us to capture nuances and complexities specific to PPG signals. 

An increasingly popular approach involves feeding timeseries data and prompts directly to Large Language Models (LLMs) \citep{gruver2024large}. However, despite promising results, LLMs struggle with high-dimensional signals due to their text-based processing \citep{spathis2024first}. A modality-specific encoder like \model{} addresses this limitation by providing representations of raw signals \citep{belyaeva2023multimodal}, which can be combined with text and fed into more powerful multimodal foundation models, such as AnyMAL \citep{moon2023anymal}. This approach offers several advantages: computational efficiency through a fixed LLM, flexibility due to the modular design of encoder, adapter, and LLM components, and interoperability with other high-performing models (e.g., a state-of-the-art IMU encoder \citep{yuan2024self}). Crucially, this encoder-LLM approach does not require paired data with other modalities to train a single multimodal model. However, it may introduce complexity by limiting end-to-end gradient propagation and reduce interpretability in encoder-LLM communication compared to natural language prompts. Despite these trade-offs, \model{} serves dual purposes: as a generic feature extractor for various PPG signals and applications, and as a modality encoder in next-generation frontier models. This versatility positions it as a valuable tool for advancing multimodal sensory AI systems.

\section{Extended Discussion}
\label{appendix:extented_discussion}
In  \S\ref{sec:overall_performance}, we observed that \model{} outperforms baselines in at least 14 out of 20 tasks, with average classification and regression improvements of 4.7\%-6.3\% and 2.9\%-4.9\%, respectively. \model{}-S performed best for cardiovascular parameters like BP, Hypertension, and Avg. HR, which are closely linked to metrics such as sVRI and IPA \citep{liang2018hypertension}. Additionally, \model{}-P surpassed baselines FMs like Moment and is well-suited for tasks such as Smoking and Arousal.

By ablating different components of \model{}-S (Section \ref{sec:ablation_study}), we found that the full model performs best, with sVRI contributing the most. Adding IPA or SQI separately did not improve performance, suggesting that (a) IPA and SQI positively transfer in a multi-task setup, and (b) our design choice to include both to compensate for situations where IPA cannot be computed is effective (Section \ref{sec:segment_aware}). While combining \model{}-P and \model{}-S may seem intuitive, constraining positive pairs on both sVRI and participants leads to too many unique labels with limited samples. In our scalability analysis, we observed that the smallest model (5M parameters) outperformed others, aligning with other studies using CNNs with 3.3M parameters for biosignals \citep{abbaspourazad2023large}, likely due to the size of PPG datasets. Larger models like Chronos or Moment are impractical for wearables due to their size and privacy concerns with cloud-based inference for health data. Additionally, \model{}-S is more data-efficient for linear probing, showing greater performance gains with increased data availability, making it a promising backbone for small studies in future research.

Our studies in Section \ref{sec:case_studies} reveal that \model{}-S embeddings are more dispersed across participants, enhancing performance, while regression predictions more accurately reflect the true distribution. We attribute this to our positive pair selection, which chooses positive pairs across individuals based on sVRI. Moreover, our skin tone analysis shows that the method performs better on lighter skin tones, likely due to the model being trained predominantly on such data. For darker skin tones, performance was similar across models for diastolic BP, with REGLE and BYOL performing best, highlighting the need for future work creating more robust models for diverse skin tones.

{To provide future direction regarding the use of \model{}, we provide some suggestions. For instance, let’s consider the nuMoM2B dataset which consists of pregnant women. \model{}-S obtains an AUROC of 0.78 in pregnancy stage classification and 6.05 is gestation age classification. Compared to the pre-training population with diverse age and gender, the nuMoM2B consists of women generally aged between 20-35. Furthermore, the gestation age readings are collected approximately around the first and third trimester. Given these factors, the target nuMoM2B dataset has many variables contributing toward distribution shift. Therefore, PaPaGei-S can be fine-tuned to address the shift in the following ways: (1) We can align the pre-trained embeddings to the nuMoM2B embedding using unsupervised or semi-supervised domain adaptation. (2) Domain Generalization is also an option during the training phase to improve generalization robustness. (3) Newer methods such as LoRA can provide another way to quickly fine-tune. (4) Importantly, given that more women are present in the first visit compared to the third visit, we can optimize different metrics to improve accuracy under the imbalance. For example, AUPRC can be optimized instead of AUROC. Fairness of classification across genders can also be considered during training. Exploring these avenues to further enhance the performance and applicability of PaPaGei is a promising direction for future studies.} Moreover, future work may benefit from exploring PPG specific augmentations such as GAN-based approaches \citep{kiyasseh2020plethaugment}; and systematically evaluating different augmentations to provide insights into useful PPG augmentations.

\end{document}